\documentclass{article}

\PassOptionsToPackage{numbers, compress}{natbib}
\usepackage[final]{neurips_2022}

\usepackage[utf8]{inputenc} % allow utf-8 input
\usepackage[T1]{fontenc}    % use 8-bit T1 fonts
\usepackage[colorlinks=true,
            linkcolor=violet,
            citecolor=violet,
            anchorcolor=violet,
            urlcolor=violet]{hyperref}
\usepackage{url}            % simple URL typesetting
\usepackage{booktabs}       % professional-quality tables
\usepackage{amsfonts}       % blackboard math symbols
\usepackage{nicefrac}       % compact symbols for 1/2, etc.
\usepackage{microtype}      % microtypography
\usepackage{xcolor}         % colors
 
\usepackage{bm}             % added by Chen Wu
\usepackage{graphicx}       % added by Chen Wu
\usepackage{amsmath}        % added by Chen Wu
\usepackage{amssymb}        % added by Chen Wu
\usepackage{amsthm}         % added by Chen Wu
\usepackage{xspace}         % added by Chen Wu
\usepackage{adjustbox}      % added by Chen Wu
\usepackage{subfigure}      % added by Chen Wu
\usepackage{multirow}       % added by Chen Wu
\usepackage{pifont}         % added by Chen Wu
\usepackage[ruled,vlined]{algorithm2e}  % added by Chen Wu
\newcommand{\boldz}{\bm{z}}                                 % added by Chen Wu
\newcommand{\boldx}{\bm{x}}                                 % added by Chen Wu
\newcommand{\boldy}{\bm{y}}                                 % added by Chen Wu
\newcommand{\boldeps}{\bm{\epsilon}}                        % added by Chen Wu
\newcommand{\boldtheta}{\bm{\theta}}                        % added by Chen Wu
\newcommand{\boldmu}{\bm{\mu}}                              % added by Chen Wu
\newcommand{\boldgamma}{\bm{\gamma}}                        % added by Chen Wu
\newcommand{\gen}{G}                                        % added by Chen Wu
\newcommand{\inn}{f_{\boldtheta}}                           % added by Chen Wu
\newcommand{\pzsingle}{p_{\boldtheta}}                      % added by Chen Wu
\usepackage{eqparbox}                                       % added by Chen Wu
\newcommand{\down}{$\downarrow$}                            % added by Chen Wu
\newcommand{\promptgen}{PromptGen\xspace}                   % added by Chen Wu
\newcommand{\anonymoustext}[1]{}                          % added by Chen Wu
\newcommand{\opensource}{\url{https://github.com/ChenWu98/Generative-Visual-Prompt}}   % added by Chen Wu

\usepackage{inconsolata}

\title{
Generative Visual Prompt: Unifying Distributional \\ Control of Pre-Trained Generative Models
}

\author{%
  Chen Henry Wu, \ \  Saman Motamed, \ \ Shaunak Srivastava, \ \ Fernando De la Torre 
  \\
  Robotics Institute, Carnegie Mellon University,
  Pittsburgh, PA  \\
  \texttt{\{chenwu2,ftorre\}@cs.cmu.edu, \{saman.moatamed,shaunak1999\}@gmail.com} \\
}

\begin{document}

\maketitle

\begin{abstract}
Generative models (e.g., GANs, diffusion models) learn the underlying data distribution in an unsupervised manner. 
However, many applications of interest require sampling from a particular region of the output space or sampling evenly over a range of characteristics.
For efficient sampling in these scenarios, we propose \textbf{Generative Visual Prompt} (\promptgen), a framework for distributional control over pre-trained generative models by incorporating knowledge of other off-the-shelf models.
\promptgen defines control as energy-based models (EBMs) and samples images in a feed-forward manner by approximating the EBM with invertible neural networks, avoiding optimization at inference. Our experiments demonstrate how \promptgen can efficiently sample from several unconditional generative models (e.g., StyleGAN2, StyleNeRF, diffusion autoencoder, NVAE) in a controlled or/and de-biased manner using various off-the-shelf models: (1) with the CLIP model as control, \promptgen can sample images guided by text, 
(2) with image classifiers as control, \promptgen can de-bias generative models across a set of attributes or attribute combinations, and (3) with inverse graphics models as control, \promptgen can sample images of the same identity in different poses. (4) Finally, \promptgen reveals that the CLIP model shows a ``reporting bias'' when used as control, and \promptgen can further de-bias this controlled distribution in an iterative manner.\footnote{The code is available at \opensource.
}

\end{abstract}

\section{Introduction}

\begin{figure}[!ht]
\centering
  \includegraphics[width=\linewidth]{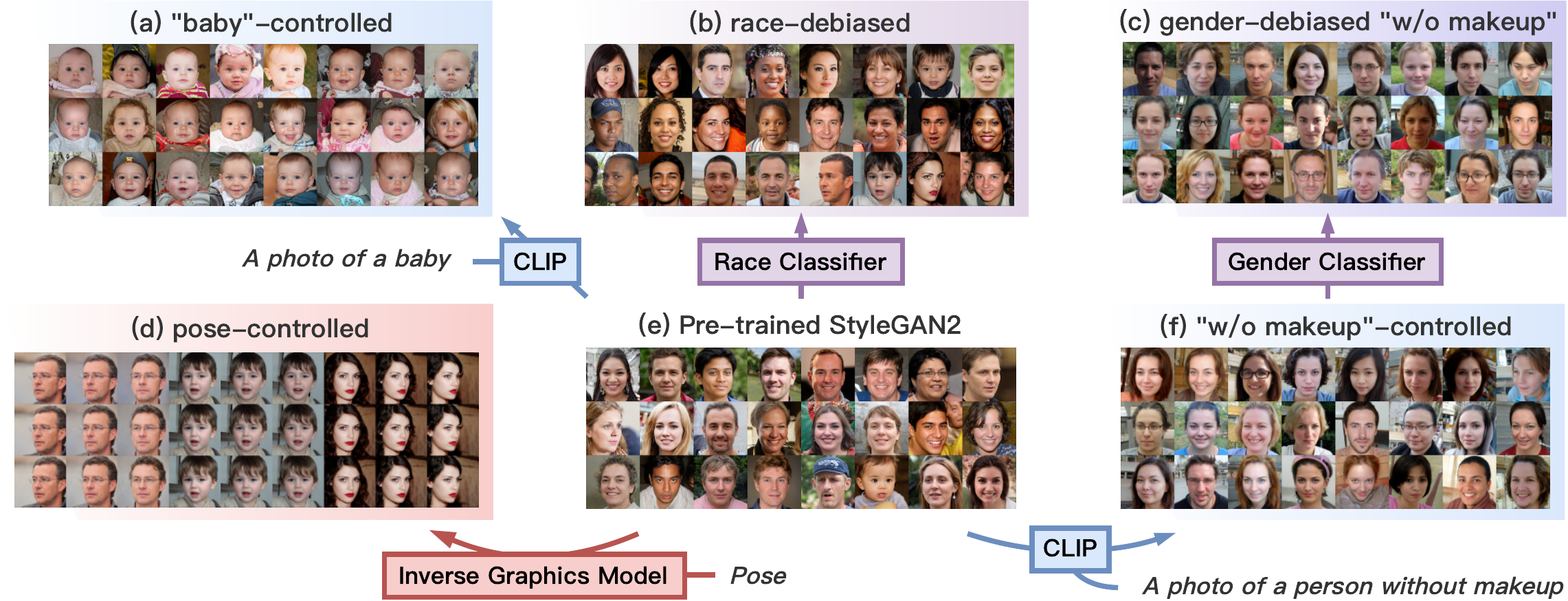}
\caption{\label{fig:first_fig} 
\promptgen uses different off-the-shelf models (e.g., CLIP~\cite{Radford2021LearningTV}, inverse graphics models, and classifiers) to control the output distribution of a pre-trained and fixed generative model (e.g., StyleGAN2). Colored boxes (e.g., the blue CLIP) indicate the control. See the text for details.
}
\end{figure}

% GAN's problem
Generative models learn the underlying high-dimensional data distribution and
have achieved promising performance on image synthesis \cite{brock2018large,VahdatK20,Karras2020AnalyzingAI,gu2022stylenerf}. Though being well praised, they still face two main limitations. First, since generative models are typically trained in an unsupervised way, they lack controllability, meaning that it is unclear how to sample from a specific region of the space. Second, generative models are prone to inherit the imbalance and bias of training data~\cite{ramaswamy2020debiasing,Karakas2022FairStyleDS}. 
For instance, StyleGAN2 is more likely to produce images of white individuals, see Figure~\ref{fig:first_fig}(e). Previous works have studied these challenges separately, and typical methods include editing of ``style'' codes \cite{HarkonenHLP20,Shen2022InterFaceGANIT,Karakas2022FairStyleDS} and explicit conditions~\cite{LiQLT19,Shoshan2021GANControlEC}. However, these methods are either model-dependent (i.e., requiring a well-structured style space) or label-intensive (i.e., requiring all training samples to be labeled for explicit conditions), limiting their generality and practical use. 

%%%%%%%%%%%%%%%%%%%%%%%%%%%%%%%%%%%%%%%%%%%%%%%%%%%%%%%%%%%%%%%%%%%%%%%%%%%%%%
%%%%%%%%%%%%%%%%%%%%%%%%%%%%%%%%%%%%%%%%%%%%%%%%%%%%%%%%%%%%%%%%%%%%%%%%%%%%%%

To address the above challenges, this paper advocates a unified formulation, \textit{distributional control of generative models}, which enables controllability by incorporating the knowledge of off-the-shelf models (e.g., CLIP \cite{Radford2021LearningTV}, classifiers, or inverse graphics models \cite{Feng2021LearningAA}). 
Based on this unified view, we propose to learn
%conditional distribution (condition to the control) 
distributions in the latent space of a pre-trained generative model while keeping the pre-trained weights fixed. 
%Latent codes sampled from our learned distribution, when mapped by the pretrained GAN to the image space, are encouraged to satisfy the specified distributional control. 
Given its conceptual similarity to prompting \cite{Lester2021ThePO,Zhou2021LearningTP,Zhou2022ConditionalPL,Jia2022VisualPT,Gu2022ADL}, we term our framework as Generative Visual Prompt (\promptgen).
\promptgen requires \textbf{no training data}, and the only supervision comes from off-the-shelf models that help define the control. 
Specifically, \promptgen allows the user to define controls using an energy-based model (EBM) approximated by an invertible neural network (INN). 
%The invertibility allows us to sample latent codes efficiently while evaluating the density, enabling us to minimize the KL divergence from the derived EBM. 
Unlike methods that require optimization at inference in EBM sampling \cite{Nguyen2017PlugP,Du2019ImplicitGA,liu2021learning,nie2021controllable}, \promptgen samples images in a \textbf{feed-forward} manner, which is highly efficient. Moreover, \promptgen \textbf{stands alone} at inference, meaning that we can discard the off-the-shelf models, which define the control, after training.

We illustrate the benefits of \promptgen with several experiments.
Figure~\ref{fig:first_fig} demonstrates our main findings with StyleGAN2 trained on FFHQ \cite{Karras2019ASG} without any labels, while we also show results for StyleNeRF \cite{gu2022stylenerf}, diffusion autoencoder \cite{Preechakul2021DiffusionAT}, and NVAE \cite{VahdatK20} in Section~\ref{sec:experiments}.
Figure~\ref{fig:first_fig}(a) illustrates how we can sample from StyleGAN2 based on text descriptions such as ``a photo of a baby''. 
Figure~\ref{fig:first_fig}(b) shows that \promptgen can leverage a race classifier (potentially trained on a different dataset) to sample uniformly across all races, de-biasing the pre-trained StyleGAN2. 
Figure~\ref{fig:first_fig}(d) illustrates that \promptgen can generate images of the same identity in different poses, guided by a pose regressor. 

Finally, it is worth pointing out that
\promptgen not only offers \textbf{generality} for algorithmic design and \textbf{modularity} for control composition, but also enables \textbf{iterative} controls when some controls are contingent on others. 
For instance, one may train a text-controlled distribution and then de-bias this distribution. 
To achieve this, we view the composition of the INN and the generative model as a new ``generative model'' to be controlled. 
Figure~\ref{fig:first_fig}(f) illustrates that \promptgen reveals ``reporting bias'' of the CLIP model~\cite{Radford2021LearningTV}, where ``without makeup'' is -- perhaps surprisingly -- positively correlated with female, and Figure~\ref{fig:first_fig}(c) shows that \promptgen can further mitigate this bias with iterative control.

%%%%%%%%%%%%%%%%%%%%%%%%%%%%%%%%%%%%%%%%%%%%%%%%%%%%%%%%%%%%%%%%%%%%%%%%%%%%%%
\begin{table}[!th]
%\small
    \caption{Comparison between methods. }
    \label{tab:related-methods}
    \centering
    \begin{adjustbox}{width=\linewidth}
    \begin{tabular}{@{}lcccc@{}}
        \toprule
        & \multirow{2}*{StyleFlow \cite{Abdal2021StyleFlowAE}} &\ \ PPGM \cite{Nguyen2017PlugP} / & Guided & \multirow{2}*{\promptgen} \\
        &      & LACE \cite{nie2021controllable}   &  DDPM \cite{dhariwal2021diffusion}          & \\
        \midrule
        Arbitrary control (e.g., CLIP)          & \ding{55}     & \checkmark            & \checkmark    & \checkmark \\
        Low-dimensional latent space                            & \checkmark    & \checkmark            & \ding{55}     & \checkmark \\
        Stands alone at inference                               & \checkmark            & \ding{55}             & \ding{55}     & \checkmark \\
        Feed-forward (i.e., no inference-time optim.)     & \checkmark    & \ding{55}             & \ding{55}     & \checkmark \\
        Iterative distributional control                        & \ding{55}     & \ding{55}             & \ding{55}     & \checkmark \\
        \bottomrule
    \end{tabular}
    \end{adjustbox}
\end{table}
%%%%%%%%%%%%%%%%%%%%%%%%%%%%%%%%%%%%%%%%%%%%%%%%%%%%%%%%%%%%%%%%%%%%%%%%%%%%%%

\section{Related Work}

Over the past few years, generative models have gained the ability to generate images with high visual quality. A few of the most widely used methods include generative adversarial networks (GANs) \cite{Goodfellow2014GenerativeAN}, VAE \cite{KingmaW13}, invertible neural networks \cite{Dinh2014NICE,DinhSB17}, and diffusion models \cite{HoJA20,SongE19}. In particular, generative models trained on large amounts of unlabeled data, e.g., BigGAN \cite{brock2018large}, StyleGANs \cite{Karras2019ASG,Karras2020AnalyzingAI,karras2021aliasfree,Sauer2022StyleGANXLSS}, Glow \cite{Kingma2018GlowGF}, and diffusion models \cite{NicholD21,dhariwal2021diffusion}, achieve promising image synthesis results.

Despite their success, controllability and de-biasing are still two fundamental challenges generative models face.
For controllability, existing methods include explicit conditioning at training \cite{LiQLT19,Shoshan2021GANControlEC} or local editing of the learned representation, e.g, ``style'' codes \cite{Abdal2021StyleFlowAE,Patashnik2021StyleCLIPTM,Liu2021FuseDreamTT,Shen2022InterFaceGANIT,Xia2021GANIA}. 
For de-biasing, existing methods include local editing of ``style'' codes \cite{ramaswamy2020debiasing,Karakas2022FairStyleDS} and importance sampling for either training \cite{Grover2020FairGM} or inference \cite{humayun2022magnet}. Existing works study these problems separately, each requiring a specific design for the task studied. On the contrary, \promptgen is a unified framework for arbitrary controls defined by off-the-shelf models. The benefits of unifying tasks have been shown by a recent trend of works in multiple research areas across vision and language \cite{T5,Lu202012in1MV,Saharia2021PaletteID,Xie2022UnifiedSKGUA,Reed2022AGA}. Moreover, one can fine-tune generative models \cite{Gal2021StyleGANNADACD} for domain adaptation, which is orthogonal to \promptgen: since \promptgen maps a generative model to another generative model (Algorithm~\ref{alg:promptgan-framework}), fine-tuning can be applied before or after \promptgen training. We leave this exploration to future studies.

Previous methods usually sample from EBM \cite{LeCun2006ATO} with Markov Chain Monte Carlo (MCMC) \cite{Tieleman2008TrainingRB,Welling2011BayesianLV,Du2019ImplicitGA,Grathwohl2020Your,Du2020CompositionalVG,liu2021learning}. 
Among them, plug-and-play generative models (PPGMs) \cite{Nguyen2017PlugP} and LACE \cite{nie2021controllable} define latent-space EBMs. 
However, MCMC requires inference-time optimization, which is inefficient and requires the off-the-shelf models to be available at inference; this is also the criticism for diffusion models, regardless of being gradient-guided \cite{NicholD21,Sehwag2022GeneratingHF} or not \cite{HoJA20,ho2021classifierfree,Nichol2021GLIDETP}. 
In contrast, \promptgen achieves efficient, feed-forward sampling. Table~\ref{tab:related-methods} shows a comparison with previous methods.
%INNs have been explored in the latent space of VAEs \cite{JimenezRezende2015VariationalIW} and autoencoders \cite{Bhm2020ProbabilisticA}. 
Our INN training is similar to that proposed by \cite{No2019BoltzmannGS} to sample from physical systems, which is later used by \cite{Whang2021ComposingNF} to solve inverse problems with two composed INNs. Notably, \cite{No2019BoltzmannGS} and \cite{Whang2021ComposingNF} do not leverage a low-dimensional latent space, and \cite{Whang2021ComposingNF} requires training a separate model for each image sample. In this paper, we use INN to model arbitrary EBMs for various generative models.

%%%%%%%%%%%%%%%%%%%%%%%%%%%%%%%%%%%%%%%%%%%%%%%%%%%%%%%%%%%%%%%%%%%%%%%%%%%%%%
\begin{figure}[!th]
\centering
    \includegraphics[width=\linewidth]{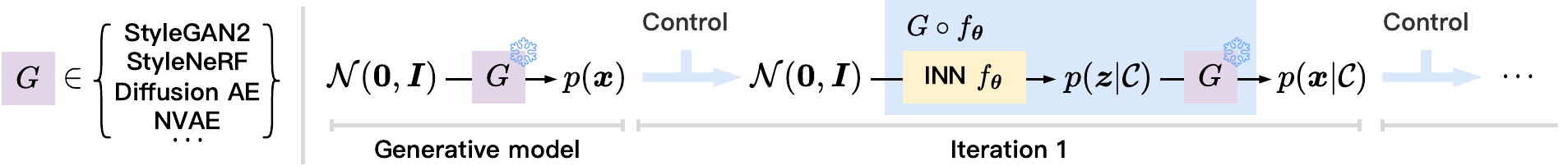}
\caption{\label{fig:overview} 
Overview of Generative Visual Prompt (\promptgen). 
Given a pre-trained generative model $\gen$ and a control $\mathcal{C}$, we learn a distribution $p(\boldz|\mathcal{C})$ in $\gen$'s latent space, while keeping $\gen$ fixed.
Each control $\mathcal{C}$ can have multiple components, e.g., $\mathcal{C} = \{\text{text = ``a photo of a baby'', gender = \texttt{male}}\}$.
\promptgen views the composition $\gen \circ \inn$ as a new ``generative model'' for iterative control. 
}
\end{figure}
%%%%%%%%%%%%%%%%%%%%%%%%%%%%%%%%%%%%%%%%%%%%%%%%%%%%%%%%%%%%%%%%%%%%%%%%%%%%%%

\section{Method}
\label{sec:method}

Figure~\ref{fig:overview} illustrates our \promptgen framework. 
To begin, the user selects a pre-trained generative model $\gen$. \promptgen then lets the user specify a control $\mathcal{C}$ as an energy-based model (EBM) $p(\boldz|\mathcal{C})$. We train an INN $\inn$ to approximate $p(\boldz|\mathcal{C})$. \promptgen views the functional composition $\gen \circ \inn$ as a new generative model and can perform iterative control. 
Algorithm~\ref{alg:promptgan-framework} describes the overall procedure.

%%%%%%%%%%%%%%%%%%%%%%%%%%%%%%%%%%%%%%%%%%%%%%%%%%%%%%%%%%%%%%%%%%%%%%%%%%%%%%
\begin{algorithm}[!th]
\DontPrintSemicolon
\textbf{Input:} Generative model $\gen: \mathbb{R}^{d} \to \mathcal{X}$\;
	\Repeat{\ user stops the iteration}{
		\smallskip
		1. \textbf{Input:} control $\mathcal{C}$ of the current iteration\;
		2. Define an EBM $p(\boldz|\mathcal{C})$ for $\mathcal{C}$ \ \ (Section~\ref{subsec:derive-ebm})\;
		3. Train an INN $\inn: \mathbb{R}^{d} \to \mathbb{R}^{d}$ to approximate $p(\boldz|\mathcal{C})$ \ \ (Section~\ref{subsec:approx-ebm})\; 
		4. $\gen \leftarrow \gen \circ \inn$\;
	}
	\Return{$G: \mathbb{R}^{d} \to \mathcal{X}$}, which is a feed-forward network\;
	\caption{Generative Visual Prompt (\promptgen)}
	\label{alg:promptgan-framework}
\end{algorithm}
%%%%%%%%%%%%%%%%%%%%%%%%%%%%%%%%%%%%%%%%%%%%%%%%%%%%%%%%%%%%%%%%%%%%%%%%%%%%%%

\subsection{Latent-Space EBM for Distributional Control}
\label{subsec:derive-ebm}
The plug-and-play generative model \cite{Nguyen2017PlugP} was first proposed to use a latent-space EBM for controllable image synthesis. If a fixed generative model is used, then theoretically, any image-space EBM can be viewed as a latent-space EBM, as shown by \cite{Grathwohl2020Your} and \cite{nie2021controllable}. We define the latent-space EBM following similar formulation as~\cite{Nguyen2017PlugP,Grathwohl2020Your,nie2021controllable}, with some new energy functions. 
We then extend the formulation to incorporate the moment constraint \cite{Csiszr2004InformationTA}, which was adopted for language modeling \cite{khalifa2021a}, but unlike \cite{khalifa2021a}, we define the moment constraint in the latent space to accommodate generative vision models. 

We define a control $\mathcal{C}$ as $M$ independent properties $\{\boldy_1, \ldots, \boldy_M\}$, e.g., $\boldy_1$ can be a text description and $\boldy_2$ can be an attribute. 
The controllability can be defined as the EBM (detailed in Appendix~\ref{subapp:image-ebm})
\begin{equation}
\label{eq:image-energy}
    p(\boldx|\mathcal{C}) = \frac{p_{\boldx}(\boldx)e^{-E_{\mathcal{C}}(\boldx)}}{Z_X}, \quad E_{\mathcal{C}}(\boldx) = \sum_{i=1}^{M} \lambda_i E_i(\boldx, \boldy_i), Z_X = \int_{\boldx'} p_{\boldx}(\boldx')e^{-E_{\mathcal{C}}(\boldx')} d\boldx',
\end{equation}
which reweights the image prior $p_{\boldx}(\boldx)$ with energy $E_{\mathcal{C}}(\boldx)$, where images with smaller energy are preferred.  
Using a pre-trained generative model $\gen: \mathbb{R}^{d} \to \mathcal{X}$ that maps a latent code $\boldz$ to an image $\boldx$, the image prior $p_{\boldx}(\boldx)$ is defined by sampling a latent code $\boldz$ from $p_{\boldz} = \mathcal{N}(\bm{0}, \bm{I})$ and mapping it to $\boldx = \gen(\boldz)$. Appendix~\ref{subapp:latent-space-ebm} shows that this EBM is equivalent to the latent-space EBM
\begin{align}
\label{eq:latent-energy}
\begin{split}
    &\quad\quad\quad\quad\quad\quad\quad\quad p(\boldz|\mathcal{C}) = \frac{p_{\boldz}(\boldz)e^{-E_{\mathcal{C}}(\gen(\boldz))}}{Z}, \\
    &E_{\mathcal{C}}(\gen(\boldz)) = \sum_{i=1}^{M} \lambda_i E_i(\gen(\boldz), \boldy_i), \quad Z = \int_{\boldz'} p_{\boldz}(\boldz')e^{-E_{\mathcal{C}}(\gen(\boldz'))} d\boldz'.
\end{split}
\end{align}
Latent-space EBM allows us to use any off-the-shelf model to specify the control.
The following are some examples that are discussed in this paper (explained in Appendix~\ref{subapp:image-ebm}): 

\textbf{Classifier energy: } Given a classifier $P(\cdot|\boldx)$ and the target class $a$ that we want to sample images from, we define the classifier energy as $E_{\text{classifier}}(\boldx, a) = - \log P(a|\boldx)$.

\textbf{CLIP energy:} Using the CLIP model~\cite{Radford2021LearningTV}, we define the CLIP energy as the cosine distance between the embeddings of the image and the text $\bm{t}$, averaged over $L$ differentiable augmentations \cite{ZhaoLLZ020,Liu2021FuseDreamTT}:
\begin{equation}
\label{eq:clip-energy}
    E_{\text{CLIP}}(\boldx, \bm{t}) = \frac{1}{L} \sum_{l=1}^{L}\bigg(1 - \cos\Big\langle\text{CLIP}_{\text{img}}\big(\text{DiffAug}_{l}(\boldx)\big), \text{CLIP}_{\text{text}}(\bm{t})\Big\rangle\bigg).
\end{equation}

\textbf{Inverse graphics energy:} Given an inverse graphics model, $f_{\mathcal{X} \rightarrow \mathcal{P}}$, which infers image parameters (e.g., pose and expression), and the target parameters $\bm{\rho}$, we define the inverse graphics energy as
\begin{equation}
\label{eq:inv-graphics-energy}
    E_{\text{inv-graphics}}(\boldx, \bm{\rho}) = d\big\langle f_{\mathcal{X} \rightarrow \mathcal{P}}(\boldx), \bm{\rho}\big\rangle^2,
\end{equation}
where $d\langle\cdot, \cdot\rangle$ is the geodesic distance between the inferred parameters and the target parameters. 

\textbf{Moment constraint: \ }
Some controls cannot be \textit{directly} defined by off-the-shelf models, and the moment constraint \cite{Csiszr2004InformationTA,khalifa2021a} is one of them. Given a mapping $\boldgamma: \mathcal{X} \to \mathbb{R}^{K}$ (e.g., $\boldgamma$ can be a classifier that outputs the probability simplex), the moment constraint defines the target distribution $p(\boldx|\mathcal{C})$ as 
\begin{equation}
\label{eq:moment-constraint-1}
    p(\boldx|\mathcal{C}) = \underbrace{\mathop{\arg\min}_{p(\boldx|\mathcal{C})} \mathbb{D}_{\mathsf{KL}}\Big(p(\boldx|\mathcal{C})\|p_{\boldx}(\boldx)\Big)}_{\text{Deviation from the pre-trained distribution}}, \quad\text{s.t. } \underbrace{\mathbb{E}_{\boldx \sim p(\boldx|\mathcal{C})}\big[\boldgamma(\boldx)\big] = \boldmu}_{\text{Moment constraint}},
\end{equation}
where $\boldmu$ is the user-specified vector. 
For example, if we want to generate images that are uniformly distributed across races, we may use a race classifier as $\boldgamma$ and define $\boldmu = \big(|\mathcal{A}|^{-1}, \ldots, |\mathcal{A}|^{-1}\big)$ where $\mathcal{A}$ is the set of races. 
It means that we would like to find a distribution that 1) stays close to the original, pre-trained generative model $\gen$ and 2) mitigates $\gen$'s bias.
In this paper, we generalize the moment constraint to the latent space, and approximate the above objective as (detailed in Appendix~\ref{subapp:distributional-derivation}):
\begin{align}
\label{eq:moment-constraint-2}
    &p(\boldz|\mathcal{C}) = \frac{p_{\boldz}(\boldz) \exp\Big(\hat{\bm{\beta}}^\top \bm{\gamma}\big(\gen(\boldz)\big)\Big)}{Z}, \quad Z = \int_{\boldz'} p_{\boldz}(\boldz')\exp\Big(\hat{\bm{\beta}}^\top \bm{\gamma}\big(\gen(\boldz')\big)\Big) d\boldz', \\
\label{eq:moment-constraint-3}
    &\hat{\bm{\beta}} = \mathop{\arg\min}_{\bm{\beta}} \mathbb{E}_{\boldz^{(1)}, \ldots, \boldz^{(N)} \overset{\text{i.i.d.}}{\sim}p_{\boldz}(\boldz), \boldx^{(j)} = \gen(\boldz^{(j)}) }  \bigg\|\frac{\sum_{j=1}^{N}\exp\big(\bm{\beta}^\top \bm{\gamma}(\boldx^{(j)})\big)\boldgamma(\boldx^{(j)})}{\sum_{j'=1}^{N}\exp\big(\bm{\beta}^\top \bm{\gamma}(\boldx^{(j')})\big)} - \boldmu \bigg\|_2^2. 
\end{align}
Notably, to optimize $\hat{\bm{\beta}}$ in Eq.~(\ref{eq:moment-constraint-3}), we leverage existing gradient-based optimization tools such as stochastic gradient descent (SGD) and Adam \cite{Kingma2015AdamAM}. 

\subsection{Approximating EBM with Invertible Neural Network}
\label{subsec:approx-ebm}

Given the functional form of the EBM $p(\boldz|\mathcal{C})$, our next step is to approximate it with an efficient sampling network. To achieve this, 
we train a distribution $\pzsingle(\boldz)$ that minimizes the KL divergence $\mathbb{D}_{\mathsf{KL}}(\pzsingle(\boldz)\|p(\boldz|\mathcal{C}))$. 
Estimating $\mathbb{D}_{\mathsf{KL}}(\pzsingle(\boldz)\|p(\boldz|\mathcal{C}))$ requires easily sampling $\boldz \sim \pzsingle$ and tractably computing $\pzsingle(\boldz)$. 
Inspired by \cite{No2019BoltzmannGS}, we model $\pzsingle$ with an INN $\inn$ that defines a bijection $\boldz = \inn(\boldeps)$, which has two merits besides the invertibility: (1) one can easily sample $\boldz$ by sampling $\boldeps \sim \mathcal{N}(\bm{0}, \bm{I})$ and mapping it to $\boldz = \inn(\boldeps)$, and (2) $\pzsingle(\boldz)$ has a closed-form solution:
\begin{equation}
\label{eq:inn}
    \log \pzsingle(\boldz) = \log \mathcal{N}(\boldeps|\bm{0}, \bm{I}) - \log |\det(\frac{\partial \inn}{\partial \boldeps})|, \quad \boldz = \inn(\boldeps).
\end{equation}
Based on these properties of INN, we can rewrite our KL divergence objective $\mathbb{D}_{\mathsf{KL}}(\pzsingle(\boldz)\|p(\boldz|\mathcal{C}))$ as (full derivations in Appendix~\ref{subapp:approx-ebm}) the following form:
\begin{equation}
\label{eq:kl-single-image}
\begin{split}
    &\mathbb{D}_{\mathsf{KL}}(\pzsingle(\boldz)\|p(\boldz|\mathcal{C})) = \mathbb{E}_{\boldz \sim \pzsingle(\boldz), \boldx = \gen(\boldz)}\Big[\log \frac{\pzsingle(\boldz)}{p_{\boldz}(\boldz)e^{-E_{\mathcal{C}}(\boldx)}/Z}\Big] \\
    %&\quad\quad\quad= \mathbb{E}_{\boldeps \sim \mathcal{N}(\bm{0}, \bm{I}), \boldz = \inn(\boldeps), \boldx = \gen(\boldz)}\Big[\log \mathcal{N}(\boldeps|\bm{0}, \bm{I}) - \log |\det(\frac{\partial \inn}{\partial \boldeps})| + E_{\mathcal{C}}(\boldx) + \log Z\Big] \\
    &= \mathbb{E}_{\boldeps \sim \mathcal{N}(\bm{0}, \bm{I}), \boldz = \inn(\boldeps), \boldx = \gen(\boldz)}\Big[- \log |\det(\frac{\partial \inn}{\partial \boldeps})| - \log p_{\boldz}(\boldz) + E_{\mathcal{C}}(\boldx)\Big] - \mathbb{H}_{\mathcal{N}(\bm{0}, \bm{I})} + \log Z.
\end{split}
\end{equation}
Since $\mathbb{H}_{\mathcal{N}(\bm{0}, \bm{I})}$ and $\log Z$ are independent of $\boldtheta$, our training objective becomes
\begin{equation}
\label{eq:objective-single-image}
    \mathop{\arg\min}_{\boldtheta} \mathbb{E}_{\boldeps \sim \mathcal{N}(\bm{0}, \bm{I}), \boldz = \inn(\boldeps), \boldx = \gen(\boldz)}\Big[\underbrace{- \log |\det(\partial \inn/\partial \boldeps)|}_{\mathcal{L}_{\text{jac}}} \  \underbrace{- \ \log p_{\boldz}(\boldz)}_{\mathcal{L}_{\text{latent-prior}}} \  \underbrace{+ \  E_{\mathcal{C}}(\boldx)}_{\mathcal{L}_{\text{energy}}}\Big]. 
\end{equation}

\begin{figure}[t]
\centering
    \includegraphics[width=\linewidth]{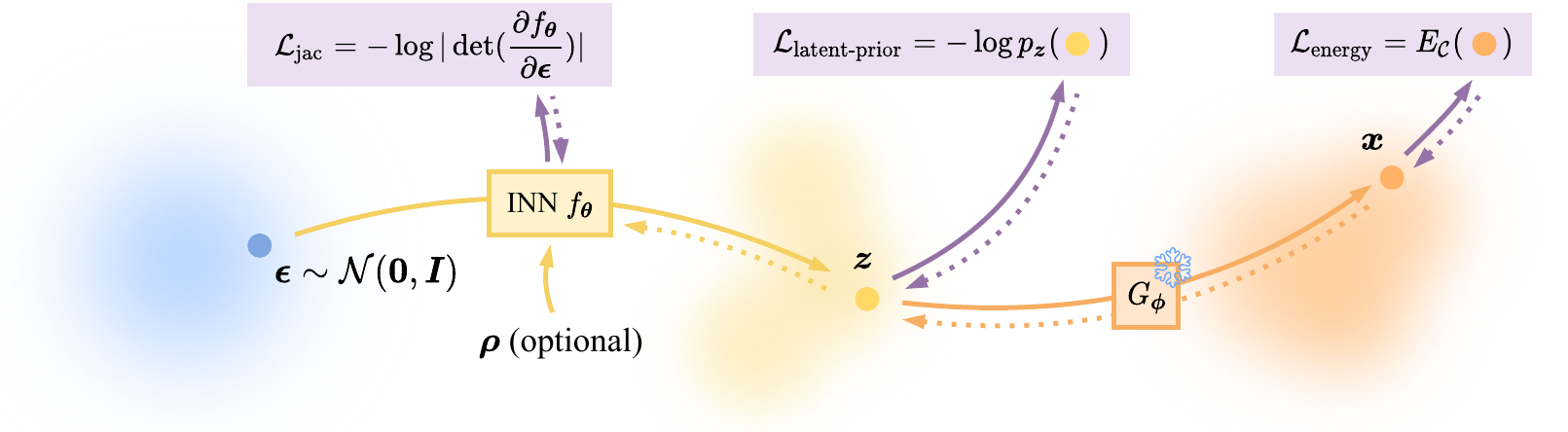}
\caption{\label{fig:approximate-ebm} Illustration for Algorithm~\ref{alg:approximate-ebm}. In the forward pass (solid curves), we sample $\boldeps \sim \mathcal{N}(\bm{0}, \bm{I})$, map $\boldeps$ to latent code $\boldz = \inn(\boldeps)$ with an INN $\inn$, and map $\boldz$ to an image $\boldx = \gen(\boldz)$ with a fixed generative model $\gen$. Dashed curves show the gradients. Details are provided in Section~\ref{subsec:approx-ebm}. }
\end{figure}

Figure~\ref{fig:approximate-ebm} gives an illustration of our process, and  Algorithm~\ref{alg:approximate-ebm} describes the algorithmic details.
%%%%%%%%%%%%%%%%%%%%%%%%%%%%%%%%%%%%%%%%%%%%%%%%%%%%%%%%%%%%%%%%%%%%%%%%%%%%%%
%%%%%%%%%%%%%%%%%%%%%%%%%%%%%%%%%%%%%%%%%%%%%%%%%%%%%%%%%%%%%%%%%%%%%%%%%%%%%%

%%%%%%%%%%%%%%%%%%%%%%%%%%%%%%%%%%%%%%%%%%%%%%%%%%%%%%%%%%%%%%%%%%%%%%%%%%%%%%
%\SetCustomAlgoRuledWidth{0.92\linewidth}
\begin{algorithm}[t]
\DontPrintSemicolon
	\While{not converged}{		
		\smallskip
		1. Sample $\boldeps \sim \mathcal{N}(\bm{0}, \bm{I})$\;
		2. Map $\boldeps$ to latent code $\boldz = \inn(\boldeps)$\;
		3. Map $\boldz$ to an image $\boldx = \gen(\boldz)$\;
		4. Optimize $\boldtheta$ with gradient $\displaystyle \nabla_{\boldtheta} \Big(- \log |\det(\frac{\partial \inn}{\partial \boldeps})| - \log p_{\boldz}(\boldz) + E_\mathcal{C}(\boldx)\Big)$\;
	}
	\caption{Approximating Latent-Space EBM with INN }
	\label{alg:approximate-ebm}
\end{algorithm}
%%%%%%%%%%%%%%%%%%%%%%%%%%%%%%%%%%%%%%%%%%%%%%%%%%%%%%%%%%%%%%%%%%%%%%%%%%%%%%

\begin{figure}[!ht]
\centering
    \subfigure[\label{subfig:stylenerf} w/ StyleNeRF on FFHQ (baby)]{
        \includegraphics[width=0.48\linewidth]{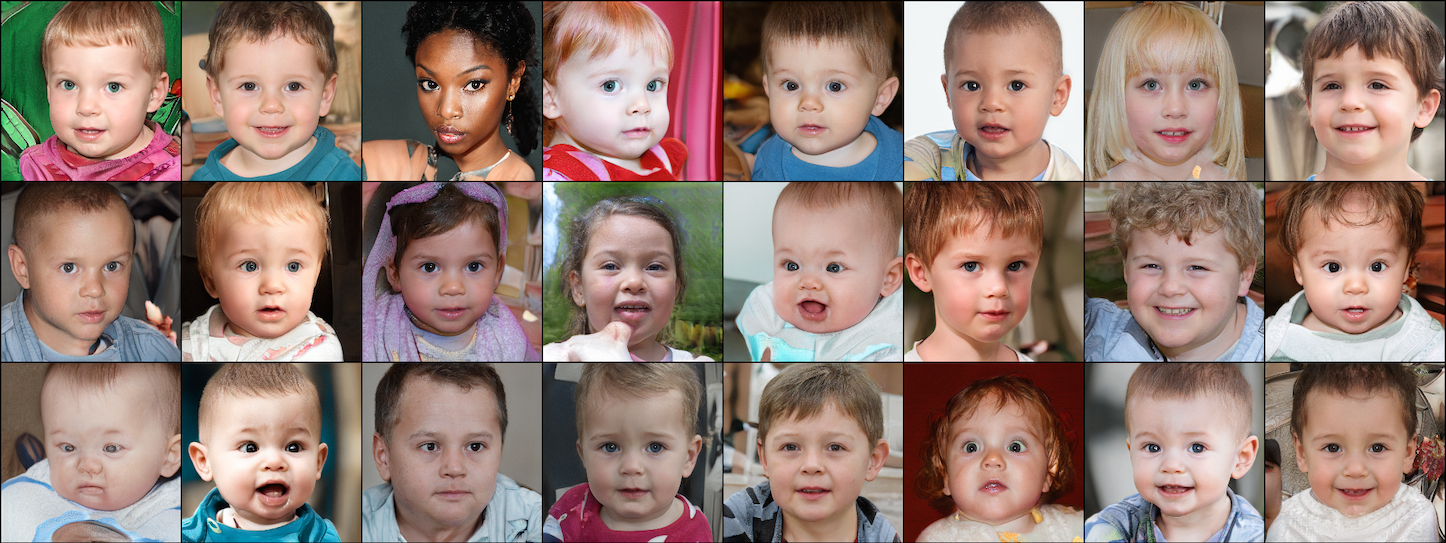}
    }
    \subfigure[\label{subfig:nvae} w/ NVAE on FFHQ (baby)]{
        \includegraphics[width=0.48\linewidth]{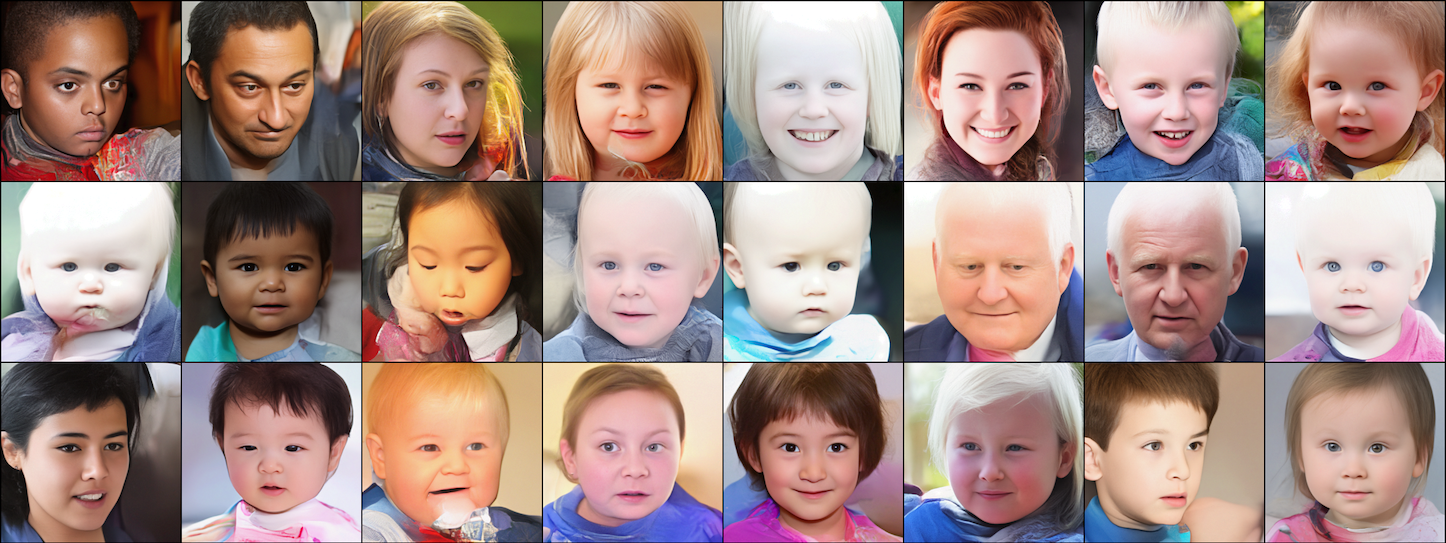}
    }
    \subfigure[\label{subfig:diffae} w/ DiffAE on FFHQ (baby)]{
        \includegraphics[width=0.48\linewidth]{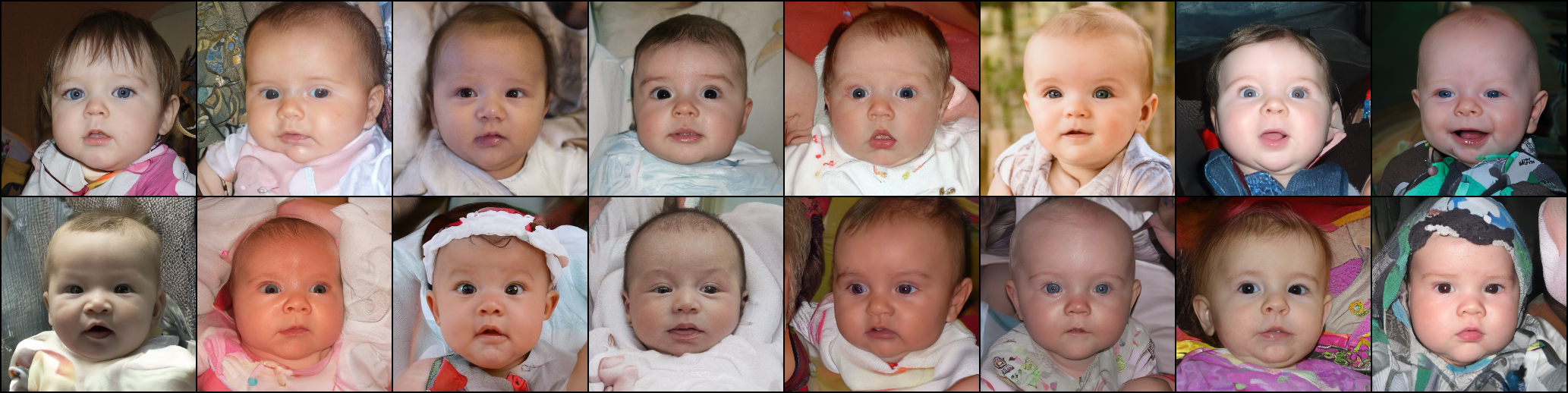}
    }
    \subfigure[\label{subfig:british-short-hair} w/ StyleGAN2 on AFHQ-Cat (British shorthair)]{
        \includegraphics[width=0.48\linewidth]{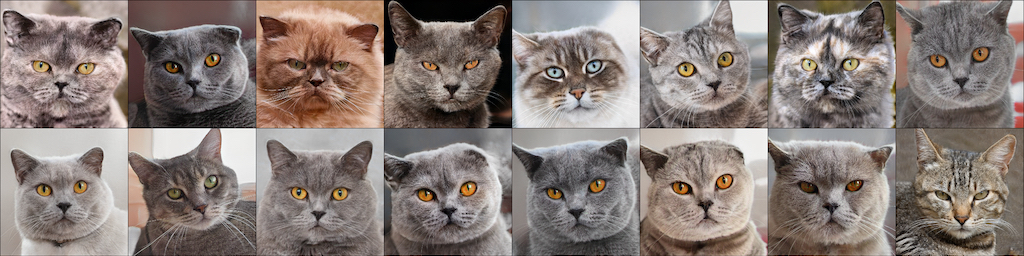}
    }
    \subfigure[\label{subfig:autumn} w/ StyleGAN2 on Landscape (autumn)]{
        \includegraphics[width=0.48\linewidth]{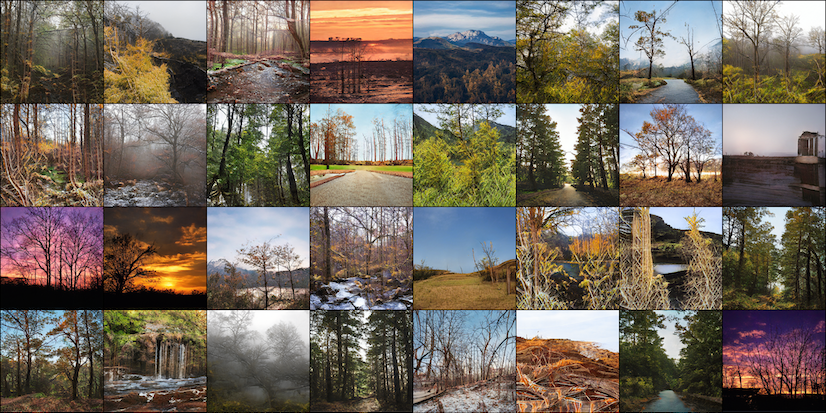}
    }
    \subfigure[\label{subfig:winter} w/ StyleGAN2 on Landscape (winter)]{
        \includegraphics[width=0.48\linewidth]{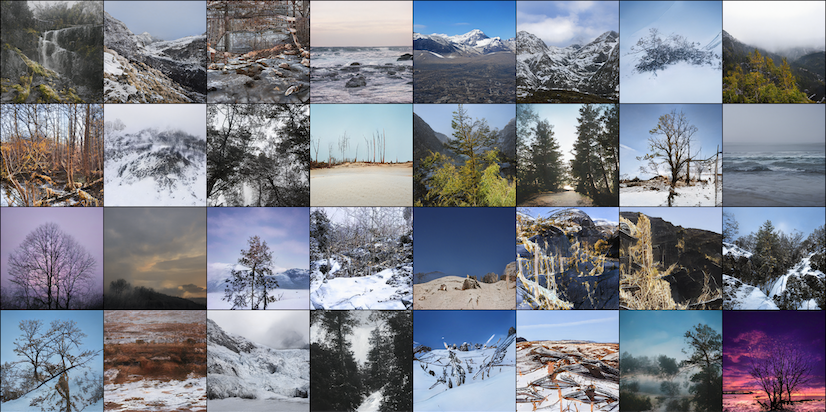}
    }
    \subfigure[\label{subfig:biggan} w/ BigGAN trained on ImageNet 512$^2$]{
        \includegraphics[width=0.98\linewidth]{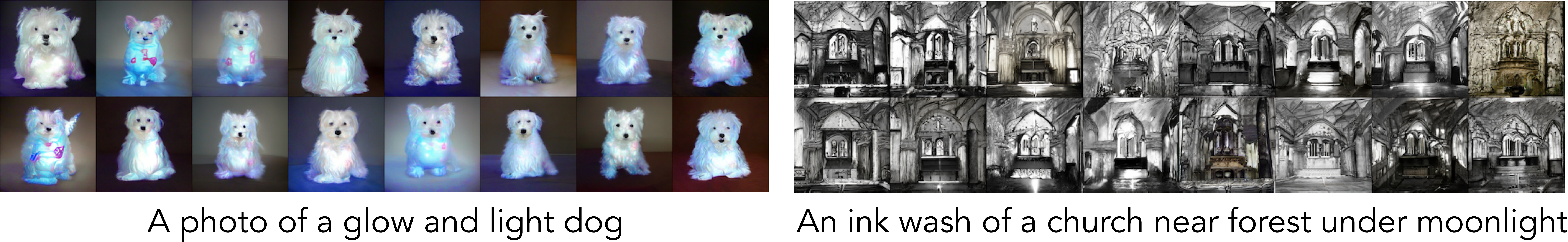}
    }
\caption{\label{fig:clip-results-more} \promptgen is applicable to different generative models in various domains. Figure~\ref{subfig:stylenerf}, Figure~\ref{subfig:nvae}, and Figure~\ref{subfig:diffae} are \promptgen applied to StyleNeRF \cite{gu2022stylenerf}, NVAE \cite{VahdatK20}, and diffusion autoencoder \cite{Preechakul2021DiffusionAT}, with text description \underline{a }p\underline{hoto of a bab}y. Figure~\ref{subfig:british-short-hair} shows \promptgen applied to StyleGAN2 \cite{Karras2020AnalyzingAI} trained on AFHQ-Cats \cite{choi2020starganv2}, with text description \underline{a }p\underline{hoto of a British shorthair cat} Figure~\ref{subfig:autumn} and Figure~\ref{subfig:winter} are \promptgen applied to StyleGAN2 trained on Landscape-HQ \cite{Skorokhodov2021AligningLA}, with text descriptions \underline{a }p\underline{hoto of }\{\underline{autumn},\underline{ winter}\}\underline{ scene}. Figure~\ref{subfig:biggan} is the extension to the embedding space of BigGAN \cite{brock2018large} on ImageNet \cite{Russakovsky2015ImageNetLS}. All images are resized for visualization. See Appendix~\ref{app:clip-samples} for results on more datasets and text descriptions.}
\end{figure}

\noindent\textbf{\promptgen in a class-embedding space} (Figure~\ref{subfig:biggan}) \ \ \ 
Previous works \cite{brock2018large,Sauer2022StyleGANXLSS} have shown that \textit{class conditioning} boosts generative models' performances on ImageNet \cite{Russakovsky2015ImageNetLS}. Specifically, class-conditioned generative models map a latent code $\boldz$ and a class embedding $\boldy$ to $\boldx = \gen(\boldz, \boldy)$. 
To extend \promptgen to these models, we train an INN $h_{\bm{\theta}}$ to map $\bm{\xi} \sim \mathcal{N}(\bm{\mu}, \bm{\sigma}^2\bm{I})$ to $\boldy = h_{\bm{\theta}}(\bm{\xi})$, where $\bm{\mu}$ and $\bm{\sigma}$ are the mean and standard deviation of $\gen$'s class embeddings. 
The motivation for defining the distribution of $\bm{\xi}$ as $\mathcal{N}(\bm{\mu}, \bm{\sigma}^2\bm{I})$ but not $\mathcal{N}(\bm{0}, \bm{I})$ is that we want the learned INN $h_{\bm{\theta}}$ to be volume-preserving, which is easier to train. 
The training objective is to minimize $\mathbb{D}_{\mathsf{KL}}(\pzsingle(\boldz, \boldy)\|p(\boldz, \boldy|\mathcal{C}))$, which is equivalent to (full derivations in Appendix~\ref{subapp:extension-to-cgan}):
\begin{equation}
\label{eq:objective-single-image-cgan}
    \begin{split}
        &\mathop{\arg\min}_{\boldtheta} \mathbb{E}_{\boldeps \sim \mathcal{N}(\bm{0}, \bm{I}), \bm{\xi} \sim \mathcal{N}(\bm{\mu}, \bm{\sigma}^2\bm{I}), \boldz = \inn(\boldeps), \boldy = h_{\bm{\theta}}(\bm{\xi}), \boldx = \gen(\boldz, \boldy)}\Big[- \log |\det(\frac{\partial \inn}{\partial \boldeps})| - \log p_{\boldz}(\boldz) \\
        &\quad\quad\quad\quad\quad\quad\quad\quad\quad\quad\quad\quad\quad\quad\quad\quad - \log |\det(\frac{\partial h_{\bm{\theta}}}{\partial \bm{\xi}})| - \log p_{\boldy}(\boldy) + E_{\mathcal{C}}\big(\boldx\big)\Big],
    \end{split}
\end{equation}
where $p_{\boldy}(\boldy) = \mathcal{N}(\boldy|\bm{\mu}, \bm{\sigma}^2\bm{I})$ is the estimated class-embedding distribution. 

\noindent\textbf{\promptgen with conditional INN \ \ } 
To generalize the control to continuous values, e.g., the scene parameters $\bm{\rho}$ in Eq.~(\ref{eq:inv-graphics-energy}), we condition the INN on $\bm{\rho}$.
The condition is modeled by replacing $\boldz = \inn(\boldeps)$ with $\boldz = \inn(\boldeps, \bm{\rho})$. Space limited, we provide details and derivations in Appendix~\ref{subapp:extension-conditional-inn}. 
This extension results in a similar architecture to StyleFlow \cite{Abdal2021StyleFlowAE}, but StyleFlow \cite{Abdal2021StyleFlowAE} uses MLE training on image-label pairs and is only applicable to explicit condition, which is not capable of modeling EBMs. 
When the condition is modeled by the equivariant operations proposed by \cite{Worrall2017InterpretableTW}, \promptgen also satisfies the homomorphism property \cite{Worrall2017InterpretableTW} in the latent space. 
%which is not guaranteed by the encoder-decoder architecture in \cite{Worrall2017InterpretableTW}. 

%%%%%%%%%%%%%%%%%%%%%%%%%%%%%%%%%%%%%%%%%%%%%%%%%%%%%%%%%%%%%%%%%%%%%%%%%%%%%%
%%%%%%%%%%%%%%%%%%%%%%%%%%%%%%%%%%%%%%%%%%%%%%%%%%%%%%%%%%%%%%%%%%%%%%%%%%%%%%

%%%%%%%%%%%%%%%%%%%%%%%%%%%%%%%%%%%%%%%%%%%%%%%%%%%%%%%%%%%%%%%%%%%%%%%%%%%%%%
\begin{figure}[!th]
\centering
    \subfigure[\label{subfig:styleclip}StyleCLIP~\cite{Patashnik2021StyleCLIPTM}]{
        \includegraphics[width=0.316\linewidth]{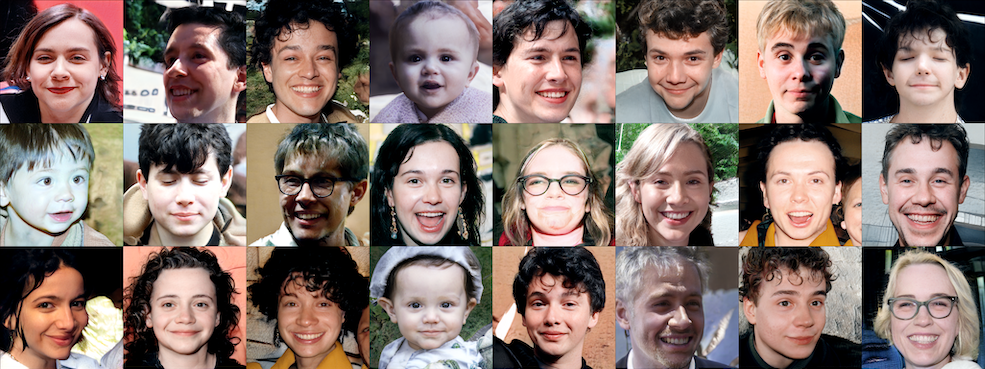}
    }
    \subfigure[\label{subfig:stylegan-nada}StyleGAN2-NADA~\cite{Gal2021StyleGANNADACD}]{
        \includegraphics[width=0.316\linewidth]{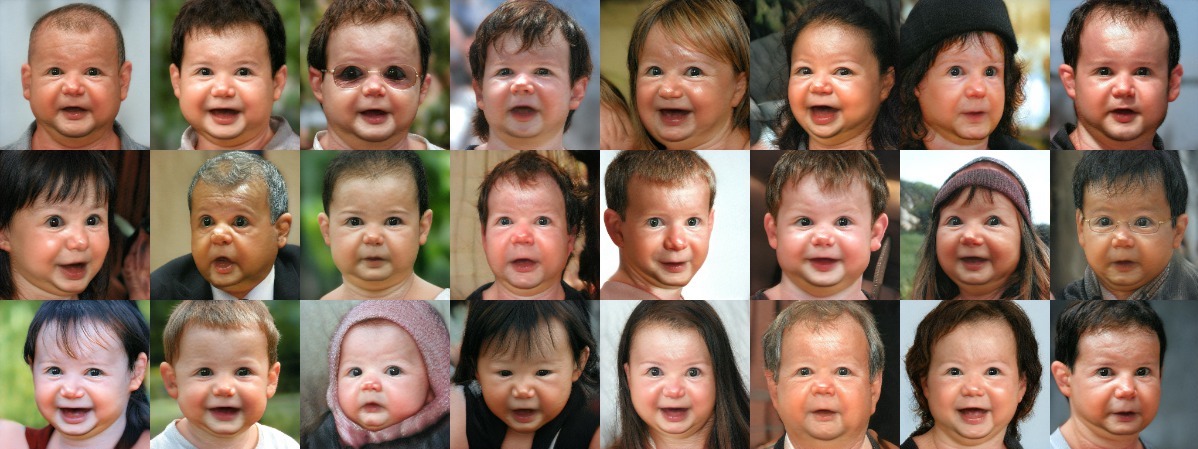}
    }
    \subfigure[\label{subfig:promptstyleganbaby}\promptgen (ours) w/ StyleGAN2]{
        \includegraphics[width=0.316\linewidth]{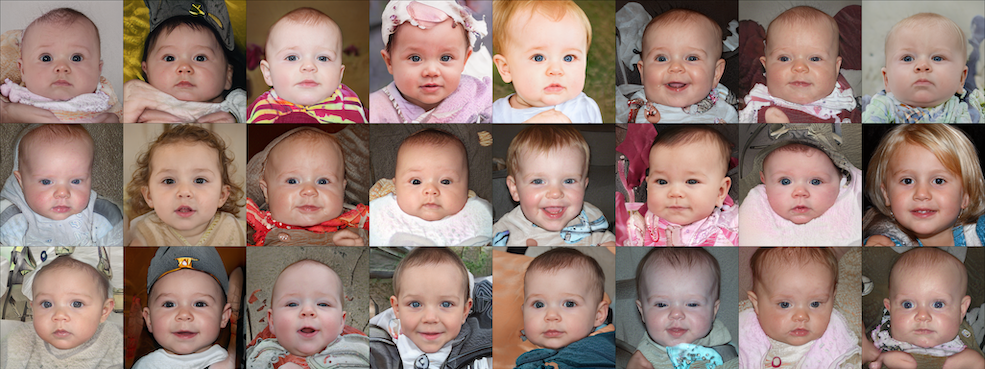}
    }
\caption{\label{fig:clip-results-baseline} Image synthesis based on text description, guided by the CLIP model. As the pre-trained generative model, we use StyleGAN2 trained on FFHQ 1024$^2$ \cite{Karras2019ASG} with truncation $\psi=0.7$. The text description used in this experiment is \underline{a }p\underline{hoto of a bab}y. StyleCLIP requires optimization at inference, while StyleGAN2-NADA and \promptgen do not. All images are resized for visualization.}
\end{figure}
%%%%%%%%%%%%%%%%%%%%%%%%%%%%%%%%%%%%%%%%%%%%%%%%%%%%%%%%%%%%%%%%%%%%%%%%%%%%%%

\section{Experiments}
\label{sec:experiments}
This section describes the experimental validation of \promptgen. See Appendix~\ref{app:synthetic} for experiments on synthetic data and Appendix~\ref{app:clip-samples}, \ref{app:debias-samples}, \ref{app:mesh}, and \ref{app:error-decomposition} for additional experiments on images and 3D meshes.
Additional experimental details are provided in Appendix~\ref{subapp:experiment-details}. 

\subsection{Image Synthesis based on Text Description}
\label{subsec:clip-experiment}
This experiment illustrates the capability of \promptgen to sample images from a generative model driven by a text description $\bm{t}$ using the pre-trained CLIP model~\cite{Radford2021LearningTV} (the ViT-B/32 version). 
We used the CLIP energy from Eq.~(\ref{eq:clip-energy}), with text descriptions such as \underline{a }p\underline{hoto of a bab}y. 
%Instead of descriptions that favors local edits (e.g., \underline{a }p\underline{hoto of a }p\underline{erson with }p\underline{ur}p\underline{le hair}), we use descriptions such as \underline{a }p\underline{hoto of a bab}y.

Figure~\ref{fig:clip-results-baseline} shows a comparison between \promptgen and two previous CLIP-guided image generation methods, StyleCLIP \cite{Patashnik2021StyleCLIPTM} and StyleGAN2-NADA \cite{Gal2021StyleGANNADACD}. 
We observe that \promptgen generates diverse and high-quality images of babies, while StyleCLIP struggles in controllability and image quality, and StyleGAN2-NADA generates baby-like adults.\footnote{Since we care about distributions, \textit{\textbf{none} of the images in this paper are cherry- or lemon-picked}.} These results show that (1) locally editing the latent code (i.e., StyleCLIP) is not always an effective method for controlling generative models (e.g., not all images' latent code can be locally edited into a baby); (2) domain adaptation (i.e., StyleGAN-NADA) is not effective in seeking modes in a generative model.
Figure~\ref{subfig:stylenerf}, Figure~\ref{subfig:nvae}, and Figure~\ref{subfig:diffae} show how \promptgen applies to StyleNeRF \cite{gu2022stylenerf}, NVAE \cite{VahdatK20}, and diffusion autoencoder (DiffAE; \cite{Preechakul2021DiffusionAT}). 
Although these generative models are all trained on FFHQ, baby images sampled from them have distinct characteristics. 
Figures~\ref{subfig:british-short-hair}-\ref{subfig:winter} show \promptgen applied to cats and landscape generation. 
Figure~\ref{subfig:biggan} shows the extension (Section~\ref{subsec:approx-ebm}) of our \promptgen to the class-embedding space of BigGAN \cite{brock2018large}, a class-conditional GAN trained on ImageNet \cite{Russakovsky2015ImageNetLS}. 
We observe that \promptgen helps BigGAN generate images with complex text descriptions, which are out of BigGAN's training image distribution; however, the diversity seems to be limited in this extension.

%%%%%%%%%%%%%%%%%%%%%%%%%%%%%%%%%%%%%%%%%%%%%%%%%%%%%%%%%%%%%%%%%%%%%%%%%%%%%%
\begin{figure}[!ht]
\centering
    \subfigure[Race dist. (FFHQ) \label{subfig:debias-results-distribution}]{
        \includegraphics[width=0.31\linewidth]{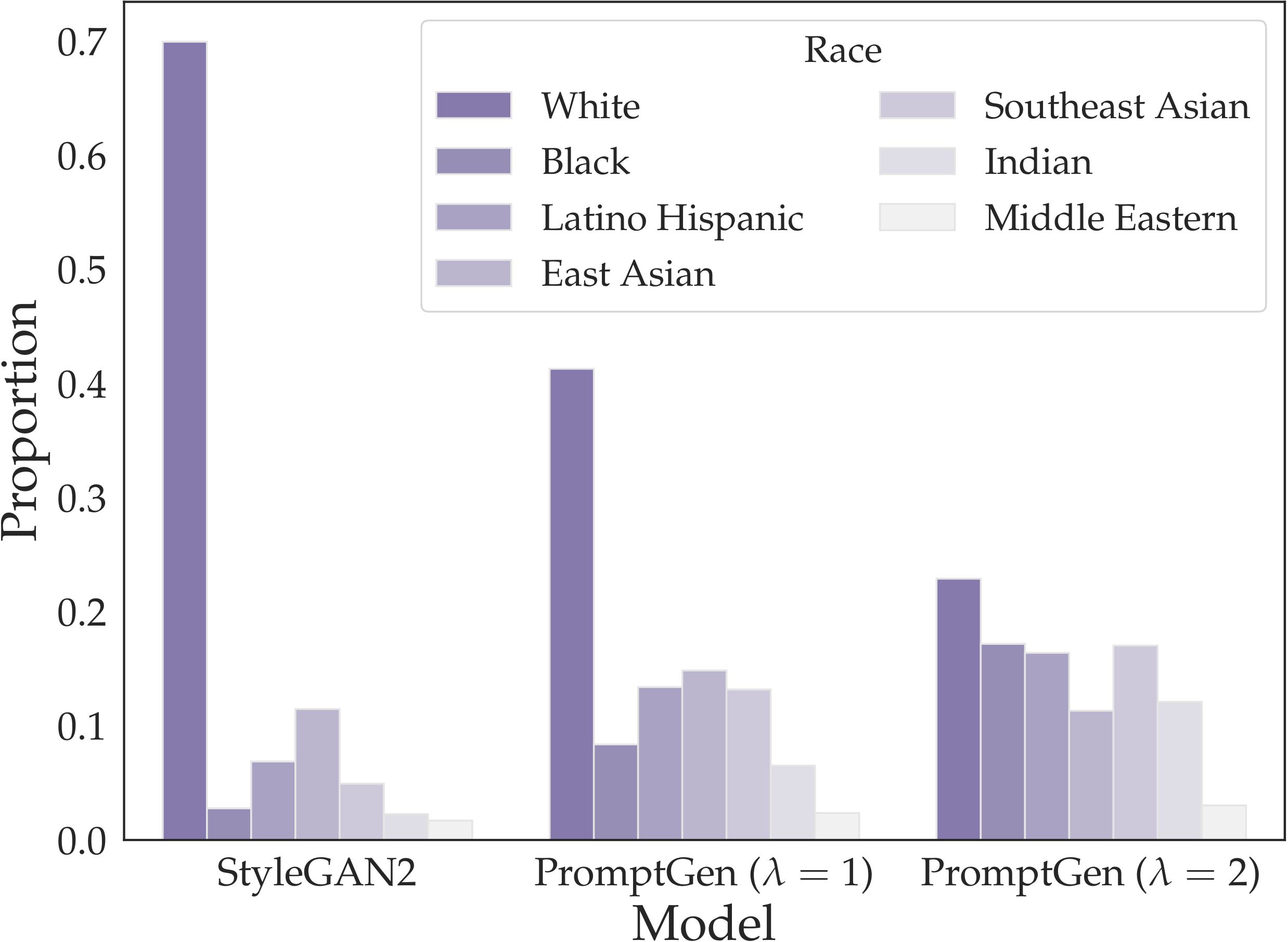}
    }
    \subfigure[Age dist. (FFHQ)]{
        \includegraphics[width=0.31\linewidth]{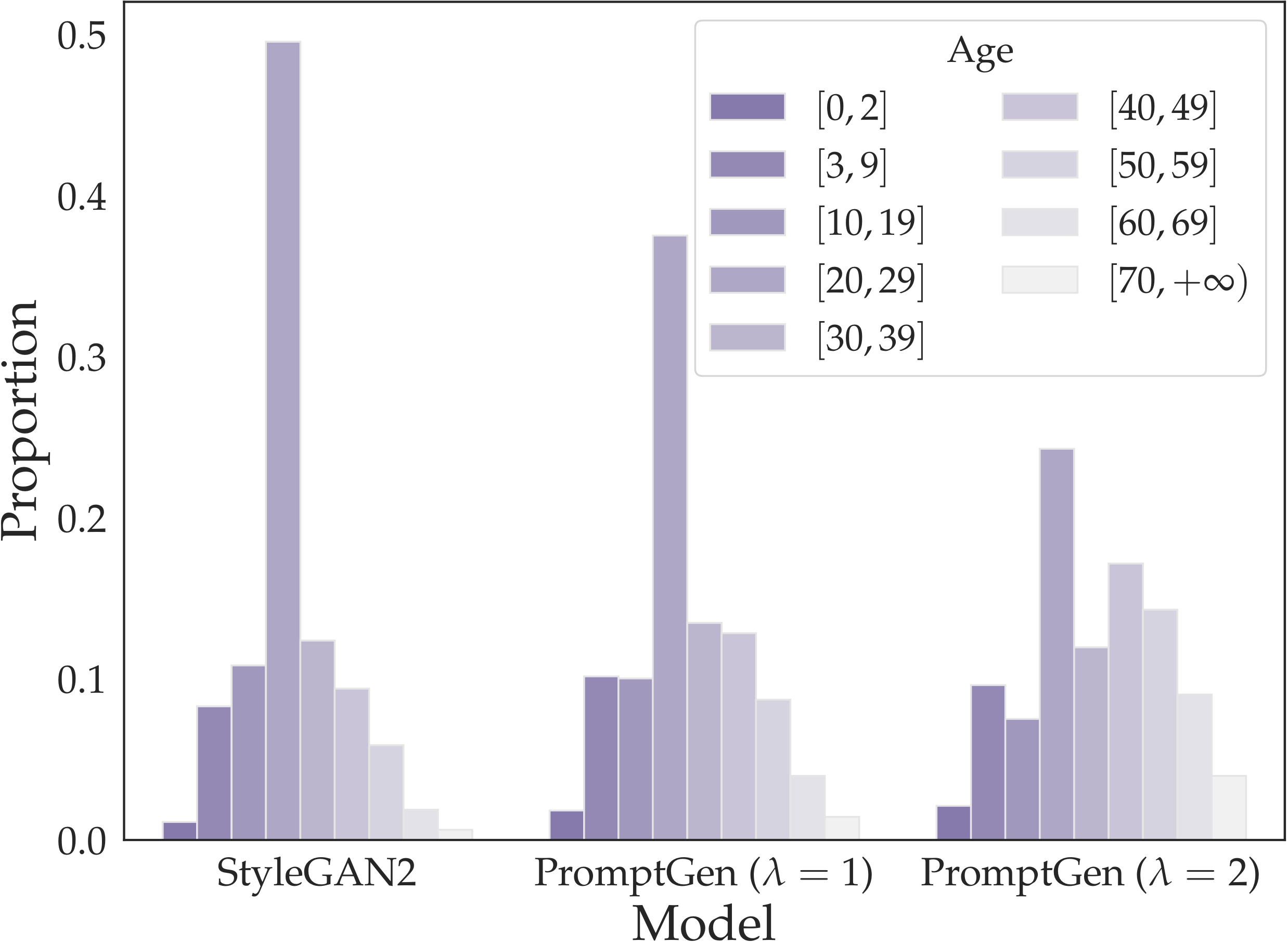}
    }
    \subfigure[Race dist. (MetFaces)]{
        \includegraphics[width=0.31\linewidth]{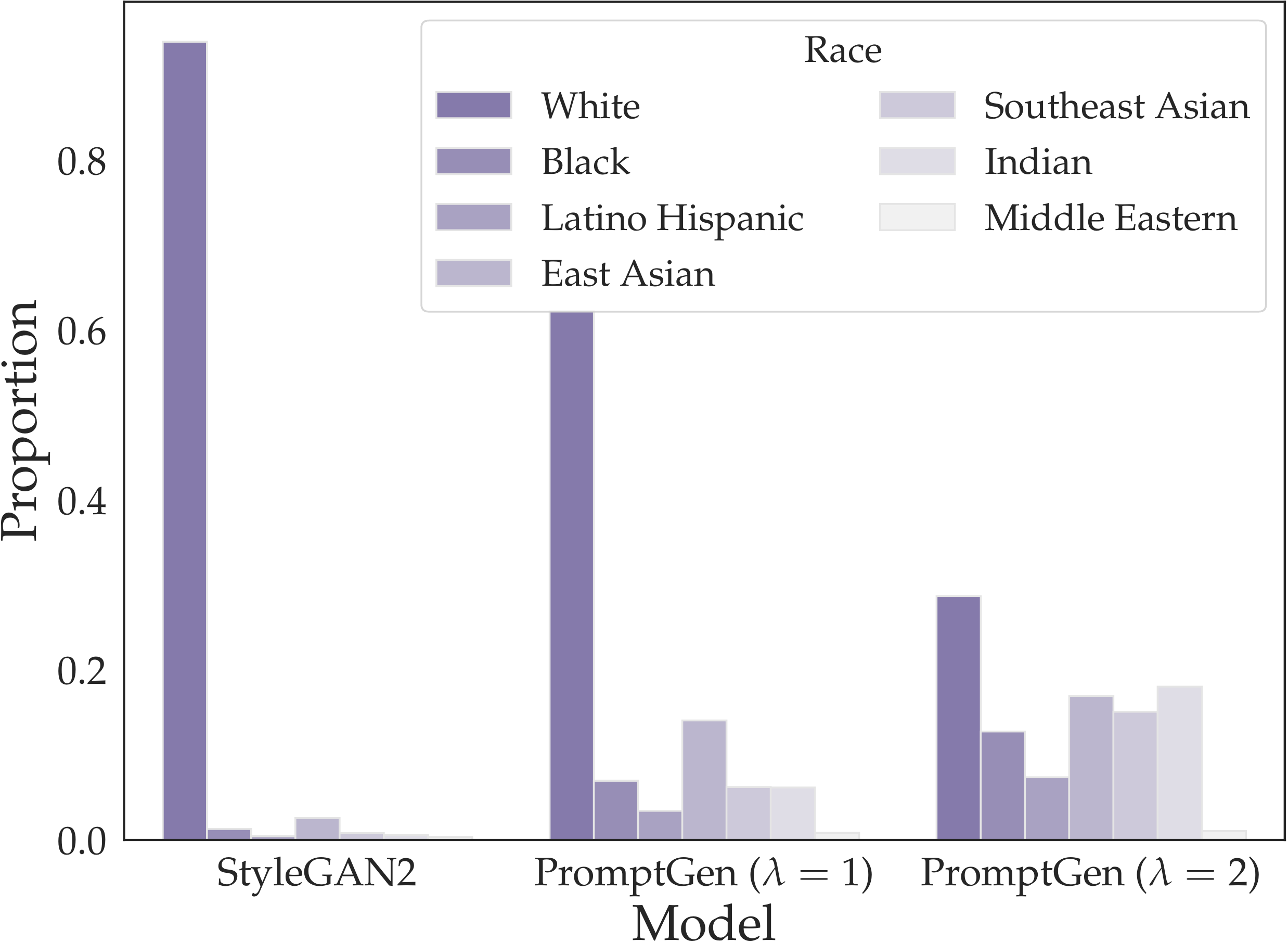}
    }
\caption{\label{fig:debias-results} With the moment constraint, \promptgen de-biases StyleGAN2 (on FFHQ and MetFaces 1024$^2$, truncation $\psi=0.7$). See Appendix~\ref{app:debias-samples} for image samples and Table~\ref{tab:debias-results} for quantitative results.}
\end{figure}
%%%%%%%%%%%%%%%%%%%%%%%%%%%%%%%%%%%%%%%%%%%%%%%%%%%%%%%%%%%%%%%%%%%%%%%%%%%%%%

\subsection{De-Biasing Pre-Trained Generative Models}
\label{subsec:debias-experiment}

An important problem in generative models is to generate fair distributions w.r.t a set of attributes of interest. 
For instance, Figure~\ref{fig:debias-results}(a) shows that StyleGAN2 generates images with bias across races and ages. \promptgen
de-biases StyleGAN2 models trained on FFHQ 1024$^2$ and MetFaces 1024$^2$ \cite{KarrasAHLLA20} in terms of \textit{categorical} attributes, using the moment constraint defined in Eq.~(\ref{eq:moment-constraint-1}) and Eq.~(\ref{eq:moment-constraint-2}). As control, we used a classifier trained on FairFace 224$^2$ \cite{Krkkinen2021FairFaceFA} as $\boldgamma$. 
We defined $\boldmu = \big(|\mathcal{A}|^{-1}, \ldots, |\mathcal{A}|^{-1}\big)$, where $\mathcal{A}$ is the set of races.
Similar to the energy weights $\lambda_i$ defined in Eq.~(\ref{eq:latent-energy}), we propose to rescale the trained $\hat{\bm{\beta}}$ as $\lambda \hat{\bm{\beta}}$. 
Figure~\ref{fig:debias-results} shows that \promptgen de-biases the race and age effectively. 
% Notably, results on MetFaces show the transferability (from real faces to art) of \promptgen for de-biasing. 

Existing de-biasing baselines consider \textit{binary} attributes \cite{Karakas2022FairStyleDS}. 
For a fair comparison with them, we adopted their setting to use binary classifiers trained on CelebA \cite{Liu2015DeepLF} for de-biasing and evaluation. Since classifiers trained on CelebA also suffer from the spurious correlation between attributes, we did \textit{not} use the moment constraint for this experiment. Instead, since \promptgen allows conditional image generation with the classifier energy, we generated the same number of samples conditioned on each attribute or attribute combination. 
Table~\ref{tab:debias-results-binary} shows that \promptgen has competitive performance on de-biasing for attributes and attribute combinations. 

%%%%%%%%%%%%%%%%%%%%%%%%%%%%%%%%%%%%%%%%%%%%%%%%%%%%%%%%%%%%%%%%%%%%%%%%%%%%%%
\begin{table}[!ht]
%\small
    \caption{Comparison with baselines for de-biasing binary attributes and their correlations. Following \cite{Karakas2022FairStyleDS}, we use classifiers on CelebA \cite{Liu2015DeepLF}. Baseline performances are copied from \cite{Karakas2022FairStyleDS}. \promptgen has competitive performance, even in the cases where FairStyle achieves nearly perfect performance. }
    \label{tab:debias-results-binary}
    \centering
    \begin{adjustbox}{width=\linewidth}
    \begin{tabular}{@{}lcccccc@{}}
        \toprule
        & \multicolumn{6}{c}{FFHQ (binary attributes)} \\
        \cmidrule(l){2-7}
        & $\mathbb{D}_\text{KL}^{\text{gender}}$\down & $\mathbb{D}_\text{KL}^{\text{eyeglasses}}$\down & $\mathbb{D}_\text{KL}^{\text{blond hair}}$\down & $\mathbb{D}_\text{KL}^{\text{age}+\text{gender}}$\down & $\mathbb{D}_\text{KL}^{\text{age}+\text{eyeglasses}}$\down & $\mathbb{D}_\text{KL}^{\text{gender}+\text{eyeglasses}}$\down \\
        \midrule
        FFHQ (real data)                        & 0.015                     & 0.186                 & --        & \ \ \quad 0.246       & \ \ \quad 0.355   & \ \ \quad 0.242 \\
        StyleGAN2 \cite{Karras2020AnalyzingAI}  & 0.018                     & 0.180                 & --        & \ \ \quad 0.279       & \ \ \quad 0.384   & \ \ \quad 0.250 \\
        StyleFlow \cite{Abdal2021StyleFlowAE}   & 0.023                     & 0.061                 & --        & \ \ \quad 0.214       & \ \ \quad 0.162   & \ \ \quad 0.121 \\
        FairGen \cite{Tan2020ImprovingTF}       & 4.21 $\times 10^{-4}$     & 7.07 $\times 10^{-4}$ & --        & \quad 0.0373          & \quad 0.0330      & \ \ 0.00185 \\
        FairStyle \cite{Karakas2022FairStyleDS} & \bf 3.20 $\times 10^{-7}$ & \bf 0                 & --        & \quad 0.0257          & \quad 0.0157      & \bf 0.000241 \\
        \midrule
        \promptgen (ours)                       & 1.71 $\times 10^{-5}$     & 1.72 $\times 10^{-5}$ & 0.0008    & \bf 0.000558          & \bf 0.000415      & 0.000628 \\
        \bottomrule
    \end{tabular}
    \end{adjustbox}
\end{table}
%%%%%%%%%%%%%%%%%%%%%%%%%%%%%%%%%%%%%%%%%%%%%%%%%%%%%%%%%%%%%%%%%%%%%%%%%%%%%%

%%%%%%%%%%%%%%%%%%%%%%%%%%%%%%%%%%%%%%%%%%%%%%%%%%%%%%%%%%%%%%%%%%%%%%%%%%%%%%
\begin{figure}[!ht]
\centering
    \includegraphics[width=0.99\linewidth]{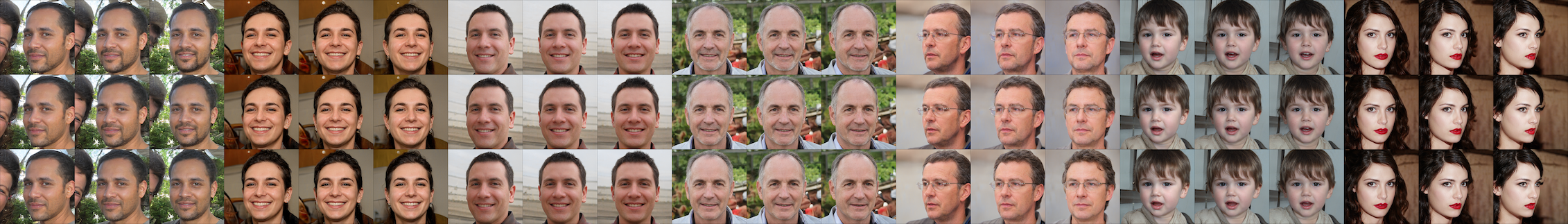}
\caption{\label{fig:inverse-graphics-results} Using inverse graphics model DECA \cite{Feng2021LearningAA} and scene parameters $\bm{\rho}_{\text{pose}} \in \text{SO}(3)$, \promptgen controls the pose of StyleGAN2 while preserving the identity. 
All images are resized for visualization.
}
\end{figure}
%%%%%%%%%%%%%%%%%%%%%%%%%%%%%%%%%%%%%%%%%%%%%%%%%%%%%%%%%%%%%%%%%%%%%%%%%%%%%%

%%%%%%%%%%%%%%%%%%%%%%%%%%%%%%%%%%%%%%%%%%%%%%%%%%%%%%%%%%%%%%%%%%%%%%%%%%%%%%
\begin{table}[!ht]
%\small
    \caption{Pose-controlled face generation. In this experiment, \promptgen uses StyleGAN2 as the pre-trained generative model. Results for GRAF \cite{SchwarzLN020}, pi-GAN \cite{Chan2021piGANPI}, GIRAFFE \cite{Niemeyer2021GIRAFFERS}, and StyleNeRF \cite{gu2022stylenerf} are from \cite{gu2022stylenerf}; results for DiscoFaceGAN (DFG) \cite{Deng2020DisentangledAC} and GAN-Control \cite{Shoshan2021GANControlEC} are from \cite{Shoshan2021GANControlEC}.}
    \label{tab:inverse-graphics}
    \centering
    \begin{adjustbox}{width=0.99\linewidth}
    \begin{tabular}{@{}l@{}cccccccc@{}}
        \toprule
                            & StyleGAN2 & GRAF   & pi-GAN    & GIRAFFE   & StyleNeRF & DFG     & GAN-Control &  \promptgen \\
        \midrule
        Resolution          & 1024$^2$  & 256$^2$           & 256$^2$           & 256$^2$           & 1024$^2$          & 256$^2$           & 512$^2$             & 1024$^2$  \\
        Pose                & \ding{55} & \checkmark    & \checkmark    & \checkmark    & \checkmark    & \checkmark    & \checkmark    & \checkmark \\
        FID\down            & 3         & 71            & 85            & 35            & 8             & 13            & 6             & 4 \\
        Dist. w/ same ID\down        & --        & -- & -- & -- & -- & 0.83 & 0.68 & 0.45 \\
        Dist. w/ diff. ID        & --        & -- & -- & -- & -- & 1.73 & 1.90 & 1.37 \\
        \bottomrule
    \end{tabular}
    \end{adjustbox}
\end{table}
%%%%%%%%%%%%%%%%%%%%%%%%%%%%%%%%%%%%%%%%%%%%%%%%%%%%%%%%%%%%%%%%%%%%%%%%%%%%%%

\subsection{Pose-Guided Face Synthesis} 
\label{subsec:expression-pose-experiments}

With an inverse graphics model, \promptgen can control the pose of faces generated by StyleGAN2.
We used the DECA model \cite{Feng2021LearningAA}, 
which infers the parameters of FLAME \cite{Li2017LearningAM}, a parametric facial graphics model.
We set $\bm{\rho} \in \text{SO}(3)$ as FLAME's three neck poses and used the conditional INN extension introduced in Section~\ref{subsec:approx-ebm}. 
To enable generating different poses of the same identity (ID), we propose ID energy using the IR-SE50 model \cite{Deng2019ArcFaceAA}. 
Specifically, given a canonical pose $\bm{\rho}_0$, we define $\boldz_0 = \inn(\boldeps, \bm{\rho}_0)$, and the ID energy is defined as (detailed in Appendix~\ref{subapp:inverse-graphics-detail})
\begin{equation}
\label{eq:identity-energy}
    E_{\text{ID}}(\boldx_0, \boldx) = 1 - \cos\big\langle R(\boldx_0), R(\boldx)\big\rangle, \quad \boldx_0 = \gen(\boldz_0), \boldx = \gen(\boldz),
\end{equation}
where $R$ is the IR-SE50 model \cite{Deng2019ArcFaceAA} that computes face embeddings. Figure~\ref{fig:inverse-graphics-results} shows that \promptgen generates faces of the same ID in different poses, even without being explicitly trained with poses as conditions. 
We computed the FID score \cite{Heusel2017GANsTB} of each model using Clean-FID \cite{Parmar2021OnAR}; following \cite{Karras2020AnalyzingAI}, we did not use the truncation trick when computing the FID score. 
Table~\ref{tab:inverse-graphics} shows that \promptgen outperforms existing models in terms of the FID score.
Following \cite{Shoshan2021GANControlEC}, we then reported the average IR-SE50 \cite{Deng2019ArcFaceAA} embedding distances for images with the same ID and with different IDs. 
Results show that \promptgen achieves the best ID preservation, with a slight sacrifice of ID diversity.

\subsection{Iterative Distributional Control via Functional Composition}
\label{subsec:iterative-control}

\begin{figure}[!ht]
\centering
    \subfigure[\label{subfig:curious-clip} \promptgen (iteration 1) with text \underline{a }p\underline{hoto of a }p\underline{erson without makeu}p. Gender distribution: female: 81.6\%; male: 18.4\%.]{
        \includegraphics[width=0.65\linewidth]{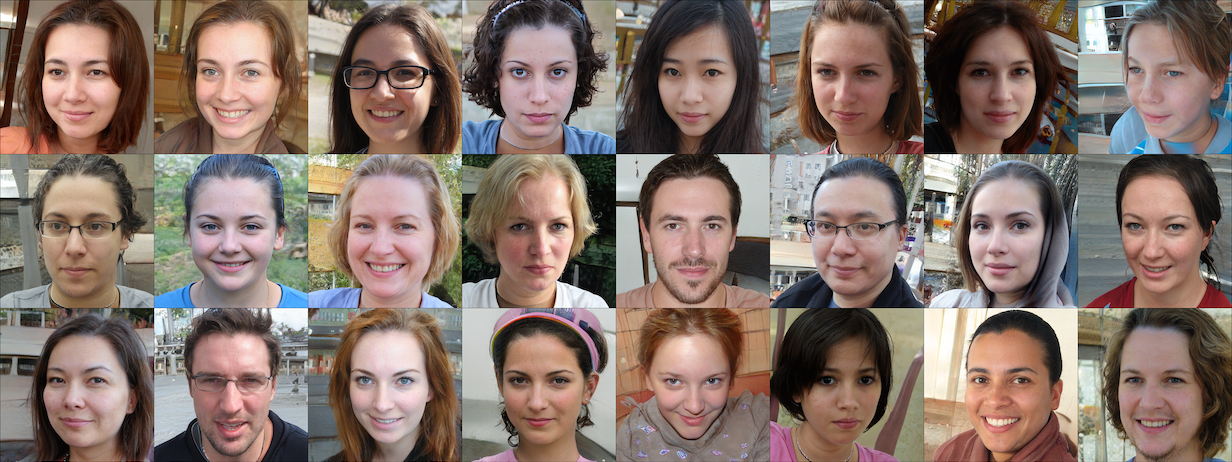}
    }
    \subfigure[\label{subfig:curious-clip-debias} \promptgen (iteration 2). Gender de-biasing ($\lambda = 2$) for the distribution learned in iteration 1. Gender distribution: female: 49.3\%; male: 50.7\%.]{
        \includegraphics[width=0.65\linewidth]{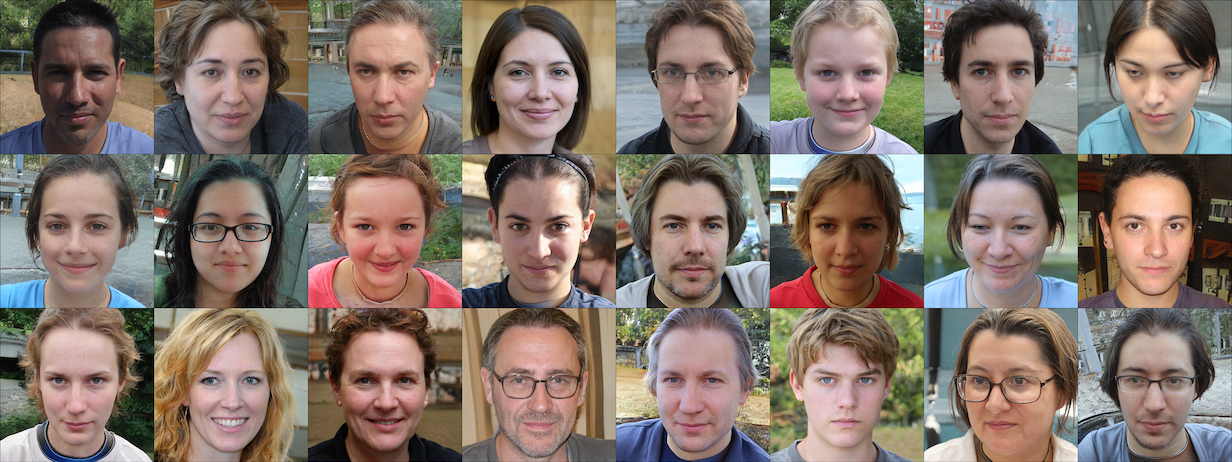}
    }
\caption{\label{fig:iterative-results} A curious case of the CLIP model: with description \underline{a }p\underline{hoto of a }p\underline{erson without makeu}p, \promptgen generates more female images than male, showing that CLIP learns a ``reporting bias''. Besides revealing this ``reporting bias'', \promptgen can also mitigate this bias via an iterative control, allowing us to de-bias the text-controlled distribution. All images are resized for visualization.
}
\end{figure}

This section discusses an interesting bias that \promptgen reveals about the CLIP model. 
Figure~\ref{subfig:curious-clip} shows images generated by 
\promptgen (with StyleGAN2) with the CLIP model and the text description \underline{a }p\underline{hoto of a }p\underline{erson without makeu}p, where more females are generated than males. This bias should not be attributed to \textit{image pre-training data} since images contain a bias in the opposite direction, i.e., men are less likely to have makeup. 
We argue that this bias should be explained by CLIP having learned a ``reporting bias'' in \textit{vision-language pre-training data}: people are more likely to \textit{say} ``a person without makeup'' when the person is a female (detailed analysis in Appendix~\ref{app:error-decomposition}). 

Besides revealing the above ``reporting bias'', \promptgen can also mitigate this bias via an iterative control, enabled by the functional composition in Algorithm~\ref{alg:promptgan-framework}. Specifically, in the second iteration, we de-biased the gender distribution of $\gen \circ \inn$ instead of $\gen$, where $\inn$ is the INN learned for the text control. For de-biasing, we used the moment constraint with $\hat{\bm{\beta}}$ trained for $\gen \circ \inn$. Figure~\ref{subfig:curious-clip-debias} shows that females and males are uniformly distributed after the moment constraint in the second iteration.

%%%%%%%%%%%%%%%%%%%%%%%%%%%%%%%%%%%%%%%%%%%%%%%%%%%%%%%%%%%%%%%%%%%%%%%%%%%%%%
%%%%%%%%%%%%%%%%%%%%%%%%%%%%%%%%%%%%%%%%%%%%%%%%%%%%%%%%%%%%%%%%%%%%%%%%%%%%%%

\subsection{Inference Latency}
\label{subsec:analysis}
We compared our \promptgen and the plug-and-play generative model (PPGM)~\cite{Nguyen2017PlugP} in terms of the inference latency. PPGM uses Langevin dynamics~\cite{Welling2011BayesianLV} to optimize over the latent space at inference, while \promptgen samples images in a feed-forward manner. In this experiment, we used PPGM and our \promptgen to approximate the same EBM with CLIP energy. 
Inference times were estimated on an NVIDIA RTX A4000 GPU.
Table~\ref{tab:clip-results} shows that our \promptgen has the highest performance and efficiency. 
Compared to our \promptgen, PPGM requires 100$\times$ inference time to achieve comparable results. 
Moreover, learning $\inn$ in \promptgen requires training-time optimization, and this amortized optimization is useful when one wants to reuse a controlled distribution many times.

%%%%%%%%%%%%%%%%%%%%%%%%%%%%%%%%%%%%%%%%%%%%%%%%%%%%%%%%%%%%%%%%%%%%%%%%%%%%%%
\begin{table}[!tbh]
    \caption{Comparison between \promptgen and PPGM, when approximating the \textit{same} EBM with CLIP energy. $n$ is the number of inference-time optimization steps used by PPGM. All models used StyleGAN2 as the pre-trained generative model. The text descriptions are \underline{a }p\underline{hoto of a }\{\underline{bab}y,\underline{ bo}y,\underline{ }g\underline{irl}\}. }
    \label{tab:clip-results}
    \centering
    \begin{adjustbox}{width=0.99\linewidth}
    \begin{tabular}{@{}lccc@{}}
        \toprule
                                & PPGM ($n=10$) & PPGM ($n=50$)     & \promptgen (ours) \\
        \midrule
        CLIP energy (baby)\down                         & 0.7327    & 0.7134        & \bf 0.7038 \\
        CLIP energy (girl)\down                         & 0.7257    & \bf 0.7184    & 0.7199 \\
        CLIP energy (boy)\down                          & 0.7263    & 0.7114        & \bf 0.7081 \\
        \midrule
        Inference time per sample (sec.)\down           & 4.4       & 21.5          & \bf 0.2 \\
        Back-propagation through CLIP at inference      & 10        & 50            & 0 \\
        When is it equal to the EBM?                    & $n \to \infty$ & $n \to \infty$ & $\mathbb{D}_{\mathsf{KL}}(\pzsingle(\boldz)\|p(\boldz|\mathcal{C})) = 0$ \\
        \bottomrule
    \end{tabular}
    \end{adjustbox}
\end{table}
%%%%%%%%%%%%%%%%%%%%%%%%%%%%%%%%%%%%%%%%%%%%%%%%%%%%%%%%%%%%%%%%%%%%%%%%%%%%%%

\section{Conclusions and Future Work}
\label{sec:conclusion}

This paper proposes \promptgen, a unified framework to learn latent distributions for distributional control of pre-trained generative models, by leveraging the knowledge of other off-the-shelf models. Unlike previous approaches, which require optimization for generating each sample at inference, \promptgen trains an invertible neural network to approximate the latent distribution. Thus, \promptgen can be viewed as an amortization of the inference-time optimization into a feed-forward neural network, allowing for efficient, feed-forward sampling at inference. Further, this amortization also makes \promptgen not require the availability of those off-the-shelf models at inference. In practice, \promptgen offers generality for algorithmic design and modularity for control composition, and it also enables iterative controls via functional composition. Experiments validate that \promptgen applies to various generative models (StyleGAN2, StyleNeRF, diffusion autoencoder, and NVAE), control types (continuous, discrete, and moment constraint), off-the-shelf models (CLIP, classifiers, and inverse graphics models), and data domains (faces, churches, animals, ImageNet, and landscapes). 

\noindent\textbf{Limitation and future work: \ \ } We provide an error analysis in Appendix~\ref{app:error-decomposition} and a discussion on societal impact in Section~\ref{app:societal-impact}. \promptgen is restricted by the pre-trained generative model's coverage \cite{Bau2019SeeingWA,HuhZZPH20}. Notably, since \promptgen focuses on mode-seeking and reweighting instead of domain adaptation, the support set of the output distribution will not change after \promptgen learning. We find it interesting to further explore combining \promptgen and domain adaptation \cite{Gal2021StyleGANNADACD} to achieve better generalization and adaptability.
\promptgen depends on the off-the-shelf models that provide knowledge about the control (Appendix~\ref{app:error-decomposition});
it can be beneficial to explore how to learn energy functions \cite{EngelHR18,Hu2018DeepGM} (besides our moment constraint) for fine-grained control with less bias. 

\section{Societal Impact}
\label{app:societal-impact}

Like any technology, \promptgen has both benefits and drawbacks for society. On the positive side, we have demonstrated that \promptgen can uncover biases learned by text-image models (like CLIP) and to de-bias text-image models and pre-trained generative models, suggesting that \promptgen may be a helpful tool for fair AI if used properly. The efficient inference of \promptgen also helps reduce the computational expense, giving a positive impact on the environment. However, improved controllability makes it easier to synthesize targeted images; for instance, the creation of deceptive media such as DeepFakes \cite{westerlund2019emergence,Vaccari2020DeepfakesAD} and privacy leakage. To battle these cases, we expect researchers and practitioners to use technologies that can detect fake media and mitigate privacy leakage. We also encourage practitioners to consider these risks when using \promptgen to develop systems.

\begin{ack}
The authors would like to thank the anonymous reviewers, Zoltán Ádám Milacski, Jianchun Chen, and Shubhra Aich for their valuable feedback on drafts of this paper. 
\end{ack}

\bibliography{reference}
\bibliographystyle{plain}

\medskip

\clearpage

%%%%%%%%%%%%%%%%%%%%%%%%%%%%%%%%%%%%%%%%%%%%%%%%%%%%%%%%%%%%
\section*{Checklist}

\begin{enumerate}

\item For all authors...
\begin{enumerate}
  \item Do the main claims made in the abstract and introduction accurately reflect the paper's contributions and scope?
    \answerYes{}
  \item Did you describe the limitations of your work?
    \answerYes{\\ In Section~\ref{sec:conclusion} and Appendix~\ref{app:error-decomposition}.}
  \item Did you discuss any potential negative societal impacts of your work?
    \answerYes{\\ In Section~\ref{sec:conclusion}.}
  \item Have you read the ethics review guidelines and ensured that your paper conforms to them?
    \answerYes{}
\end{enumerate}

\item If you are including theoretical results...
\begin{enumerate}
  \item Did you state the full set of assumptions of all theoretical results?
    \answerYes{\\ In Section~\ref{sec:method} and Appendix~\ref{app:method-details-and-derivations}.}
	\item Did you include complete proofs of all theoretical results?
    \answerYes{\\ In Section~\ref{sec:method} and Appendix~\ref{app:method-details-and-derivations}.}
\end{enumerate}

\item If you ran experiments...
\begin{enumerate}
  \item Did you include the code, data, and instructions needed to reproduce the main experimental results (either in the supplemental material or as a URL)?
    \answerNA{\\ We will make our code publicly available on GitHub, with instruction for each experiment. We made great efforts to ensure that most experiments will be run in one line of command. }
  \item Did you specify all the training details (e.g., data splits, hyperparameters, how they were chosen)?
    \answerYes{\\ In Appendix~\ref{subapp:experiment-details}; other training details will be included in the code release.}
	\item Did you report error bars (e.g., with respect to the random seed after running experiments multiple times)?
    \answerNo{} 
	\item Did you include the total amount of compute and the type of resources used (e.g., type of GPUs, internal cluster, or cloud provider)?
    \answerYes{\\ In Appendix~\ref{subapp:experiment-details}.}
\end{enumerate}

\item If you are using existing assets (e.g., code, data, models) or curating/releasing new assets...
\begin{enumerate}
  \item If your work uses existing assets, did you cite the creators?
    \answerYes{\\ We have cited the original papers.}
  \item Did you mention the license of the assets?
    \answerYes{\\ License will be included in the code.}
  \item Did you include any new assets either in the supplemental material or as a URL?
    \answerNA{We do not have new assets.}
  \item Did you discuss whether and how consent was obtained from people whose data you're using/curating?
    \answerYes{\\ All data we use are published datasets. }
  \item Did you discuss whether the data you are using/curating contains personally identifiable information or offensive content?
    \answerNA{\\ All data we use are published datasets. }
\end{enumerate}

\item If you used crowdsourcing or conducted research with human subjects...
\begin{enumerate}
  \item Did you include the full text of instructions given to participants and screenshots, if applicable?
    \answerNA{\\ We do not have user study.}
  \item Did you describe any potential participant risks, with links to Institutional Review Board (IRB) approvals, if applicable?
    \answerNA{\\ We do not have user study.}
  \item Did you include the estimated hourly wage paid to participants and the total amount spent on participant compensation?
    \answerNA{\\ We do not have user study.}
\end{enumerate}

\end{enumerate}

%%%%%%%%%%%%%%%%%%%%%%%%%%%%%%%%%%%%%%%%%%%%%%%%%%%%%%%%%%%%

\clearpage

\appendix

\section{\promptgen on Synthetic Data}
\label{app:synthetic}

%%%%%%%%%%%%%%%%%%%%%%%%%%%%%%%%%%%%%%%%%%%%%%%%%%%%%%%%%%%%%%%%%%%%%%%%%%%%%%
\begin{figure}[t]
\centering
    \includegraphics[width=0.97\linewidth]{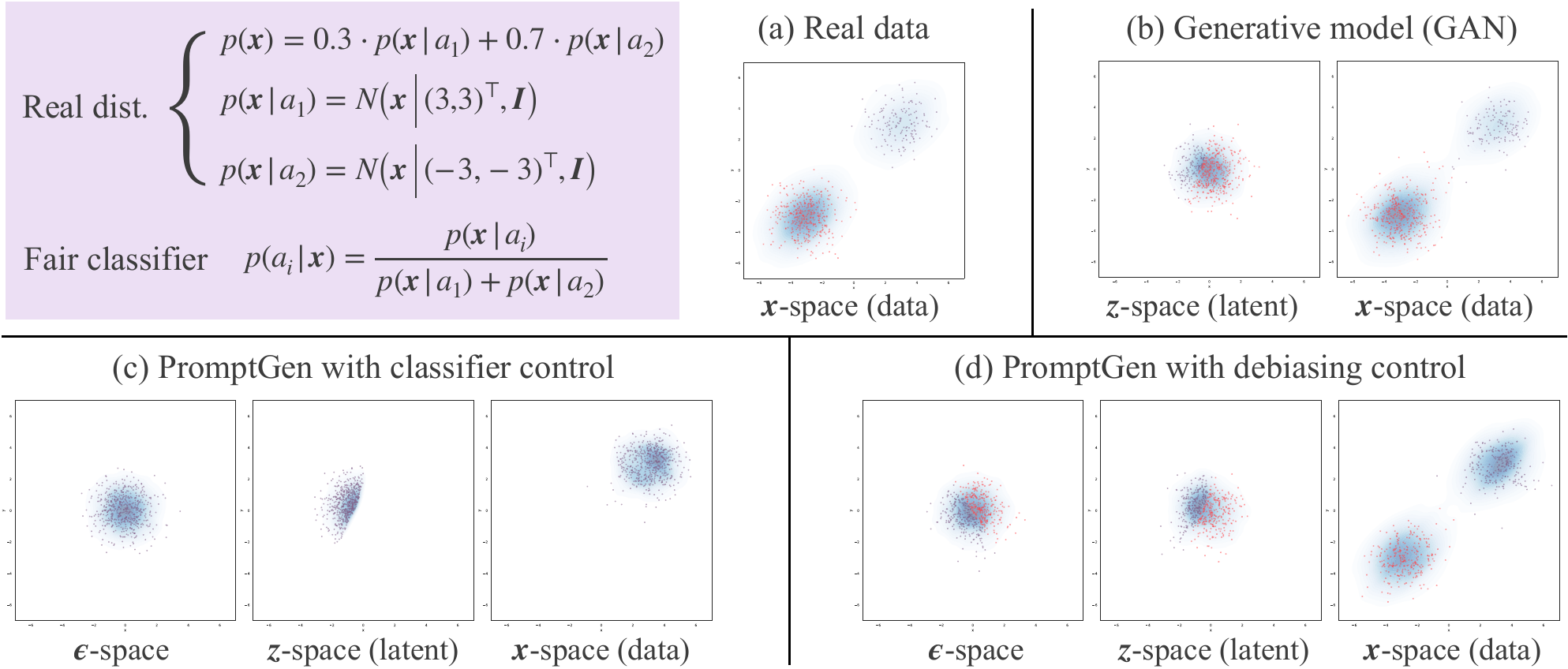}
\caption{\label{fig:toy-example} \promptgen on synthetic data. (a) The real distribution is a Gaussian mixture distribution biased towards one mode. We derive a closed-form fair classifier. (b) A GAN trained on this real distribution. (c) With the fair classifier as control, \promptgen learns a distribution concentrated in a specific region. (d) We derive the moment constraint following Section~\ref{subsec:derive-ebm} and Section~\ref{subsec:debias-experiment}, and \promptgen learns to up-weight the under-represented regions. }
\end{figure}
%%%%%%%%%%%%%%%%%%%%%%%%%%%%%%%%%%%%%%%%%%%%%%%%%%%%%%%%%%%%%%%%%%%%%%%%%%%%%%

We demonstrate the behavior of our \promptgen with two-dimensional synthetic data, using GAN as the generative model. 
This synthetic experiment is illustrated in Figure~\ref{fig:toy-example}. 
Specifically, we created a ``real'' distribution that is a Gaussian mixture distribution biased towards one mode. 
We derive a closed-form fair classifier based on this distribution, detailed in the purple block in Figure~\ref{fig:toy-example}.
When a GAN is trained to approximate this distribution, its outputs are biased towards the over-represented mode, as shown by Figure~\ref{fig:toy-example}(b).
Figure~\ref{fig:toy-example}(c) illustrates the controllability experiment: using the fair classifier as control, \promptgen learns a distribution concentrated in a specific region of the output space. 
Figure~\ref{fig:toy-example}(d) illustrates the de-biasing experiment: using the moment constraint (Section~\ref{subsec:derive-ebm} and Section~\ref{subsec:debias-experiment}), \promptgen upweights the under-represented regions of the output space. 

\section{Method Details and Derivations}
\label{app:method-details-and-derivations}

\subsection{Controllability as EBM}
\label{subapp:image-ebm}
In Section~\ref{sec:method}, we defined a control $\mathcal{C}$ as $M$ independent properties $\{\boldy_1, \ldots, \boldy_M\}$. For example, $\boldy_1$ can be a text description, and $\boldy_2$ can be an attribute. 
In this part, we first elaborate on why controllability can be formed as EBMs in Eq.~(\ref{eq:image-energy}). We then provide concrete examples of the distributions derived from different energy functions defined in Section~\ref{subsec:derive-ebm}. 

We denote the image prior as $p_{\boldx}(\boldx)$, the only distribution that can be estimated from data when labels for the control are not provided during generative model pre-training. Given the control $\mathcal{C} = \{\boldy_1, \ldots, \boldy_M\}$, we resort to Bayes' theorem to rewrite the conditional distribution $p(\boldx|\mathcal{C})$ as
\begin{align}
    p(\boldx|\mathcal{C}) &\propto p_{\boldx}(\boldx)p(\mathcal{C}|\boldx) \\
    &= p_{\boldx}(\boldx) p(\boldy_1|\boldx) \prod_{i=2}^{M} p(\boldy_i|\boldx, \boldy_{<i}) \\
\label{eq:image-condition}
    &= p_{\boldx}(\boldx) \prod_{i=1}^{M} p(\boldy_i|\boldx) \quad \text{(independence assumption)}.
\end{align}
For Eq.~(\ref{eq:image-condition}) to be well-defined, we need to define $p(\boldy_i|\boldx)$ for each $\boldy_i$. In order to incorporate the knowledge of arbitrary off-the-self models (besides image classifiers), we define each $p(\boldy_i|\boldx)$ as 
\begin{equation}
\label{eq:off-the-shelf}
    p(\boldy_i|\boldx) = \frac{\exp(- \lambda_i E_i(\boldx, \boldy_i))}{Z_i}, \quad Z_i = \int_{\boldy_i'} \exp(- \lambda_i E_i(\boldx, \boldy_i')) d\boldy_i'.
\end{equation}
Eq.~(\ref{eq:off-the-shelf}) says that $p(\boldy_i|\boldx)$ is proportional to the exponential of an energy function $E_i(\boldx, \boldy_i)$, where $\boldy_i$ with lower energy has higher density or mass. Note that Eq.~(\ref{eq:off-the-shelf}) defines a distribution over all possible values of $\boldy_i$ instead of all possible values of $\boldx$. 
Combining Eq.~(\ref{eq:image-condition}) and Eq.~(\ref{eq:off-the-shelf}), we have
\begin{equation}
    p(\boldx|\mathcal{C}) = \frac{p_{\boldx}(\boldx)e^{-E_{\mathcal{C}}(\boldx)}}{Z_X}, \quad E_{\mathcal{C}}(\boldx) = \sum_{i=1}^{M} \lambda_i E_i(\boldx, \boldy_i), Z_X = \int_{\boldx'} p_{\boldx}(\boldx')e^{-E_{\mathcal{C}}(\boldx')} d\boldx',
\end{equation}
which is the same equation as Eq.~(\ref{eq:image-energy}). 

In the following, we use Eq.~(\ref{eq:off-the-shelf}) to derive the distributions from the classifier energy, CLIP energy, and inverse graphics energy defined in Section~\ref{subsec:derive-ebm}. 

\textbf{Classifier energy: } Given a classifier $P(\cdot|\boldx)$ and the target class $a$, we define the classifier energy as $E_{\text{classifier}}(\boldx, a) = - \log P(a|\boldx)$. Using Eq.~(\ref{eq:off-the-shelf}), we arrive at:
\begin{equation}
    p_{\text{classifier}}(a|\boldx) = \frac{\exp(\lambda_{\text{classifier}} \log P(a|\boldx))}{\sum_{a'}\exp(\lambda_{\text{classifier}} \log P(a'|\boldx))} = \frac{P(a|\boldx)^{\lambda_{\text{classifier}}}}{\sum_{a'} P(a'|\boldx)^{\lambda_{\text{classifier}}}},
\end{equation}
which is equivalent to a temperature-adjusted distribution of the original classifier.  

\textbf{CLIP energy:} Using Eq.~(\ref{eq:off-the-shelf}), the CLIP energy in Eq.~(\ref{eq:clip-energy}) is equivalent to
\begin{equation}
    p_{\text{CLIP}}(\bm{t}|\boldx) \propto \exp\Bigg(-\frac{\lambda_{\text{CLIP}}}{L} \sum_{l=1}^{L}\bigg(1 - \cos\Big\langle\text{CLIP}_{\text{img}}\big(\text{DiffAug}_{l}(\boldx)\big), \text{CLIP}_{\text{text}}(\bm{t})\Big\rangle\bigg) \Bigg).
\end{equation}

\textbf{Inverse graphics energy:} Using Eq.~(\ref{eq:off-the-shelf}), the inverse graphics energy in Eq.~(\ref{eq:inv-graphics-energy}) is equivalent to
\begin{equation}
    p_{\text{inv-graphics}}(\bm{\rho}|\boldx) \propto \exp(- \lambda_{\text{inv-graphics}} d\big\langle f_{\mathcal{X} \rightarrow \mathcal{P}}(\boldx), \bm{\rho}\big\rangle^2).
\end{equation}
If the geodesic distance $d\big\langle \cdot, \cdot \big\rangle$ is the Euclidean distance, then $p_{\text{inv-graphics}}(\bm{\rho}|\boldx)$ is a Gaussian distribution whose mean is $\bm{\rho}$ and whose variance depends on the hyperparameter $\lambda_{\text{inv-graphics}}$; if the geodesic distance $d\big\langle \cdot, \cdot \big\rangle$ is the spherical distance (e.g., the distance between pose parameters defined on a unit sphere), then $p_{\text{inv-graphics}}(\bm{\rho}|\boldx)$ is a vMF distribution. 

\subsection{Equivalence between Image-Space EBM and Latent-Space EBM}
\label{subapp:latent-space-ebm}

\newtheorem{prop}{Proposition}
\begin{prop}
\label{prep:latent-space-ebm}
    \ \ Define $\boldx \sim p_{\boldx}(\boldx)$ as \underline{$\boldz \sim p_{\boldz}(\boldz), \boldx = \gen(\boldz)$} and $p(\boldx|\mathcal{C})$ as \underline{$\boldz \sim p(\boldz|\mathcal{C}), \boldx = \gen(\boldz)$}, where $p(\boldz|\mathcal{C})$ is defined as the following EBM:
    \begin{align}
    \label{eq:latent-energy-copy}
        p(\boldz|\mathcal{C}) = \frac{p_{\boldz}(\boldz)e^{-E_{\mathcal{C}}(\gen(\boldz))}}{Z}, \quad E_{\mathcal{C}}(\boldx) &= \sum_{i=1}^{M} \lambda_i E_i(\boldx, \boldy_i), Z = \int_{\boldz'} p_{\boldz}(\boldz')e^{-E_{\mathcal{C}}(\gen(\boldz'))} d\boldz'.
    \end{align}
    We have
    \begin{equation}
        p(\boldx|\mathcal{C}) = \frac{p_{\boldx}(\boldx)e^{-E_{\mathcal{C}}(\boldx)}}{Z_X}, \quad Z_X = \int_{\boldx'} p_{\boldx}(\boldx')e^{-E_{\mathcal{C}}(\boldx')} d\boldx'.
    \end{equation}
\end{prop}

In spirit, our proof follows the proof in \cite{nie2021controllable}, which follows \cite{Grathwohl2020Your}. The difference between our proof and that in \cite{nie2021controllable} is that we derive $p(\boldx|\mathcal{C})$ from $p(\boldz|\mathcal{C})$ while they derived $p(\boldz|\mathcal{C})$ from $p(\boldx|\mathcal{C})$.

\begin{proof}
    Based on Lemma 1 in \cite{Grathwohl2020Your} and Lemma 1 in \cite{nie2021controllable}, $p(\boldz|\mathcal{C})$ is equivalent to rejection sampling with proposal distribution $p_{\boldz}(\boldz)$ and acceptance probability
    \begin{align}
        r(\boldz) = \frac{e^{-E_{\mathcal{C}}(\gen(\boldz))}}{M_{\mathcal{C}}\cdot Z}, \quad \text{where } \forall \boldz, M_{\mathcal{C}} > \frac{e^{-E_{\mathcal{C}}(\gen(\boldz))}}{Z}.
    \end{align}
    Since $p(\boldx|\mathcal{C})$ is defined as \underline{$\boldz \sim p_{\boldz}(\boldz|\mathcal{C}), \boldx = \gen(\boldz)$}, $p(\boldx|\mathcal{C})$ is equivalent to rejection sampling with proposal distribution \underline{$p_{\boldz}(\boldz), \boldx = \gen(\boldz)$} and acceptance probability
    \begin{align}
        r(\boldx) = \frac{e^{-E_{\mathcal{C}}(\boldx)}}{M_{\mathcal{C}}\cdot Z}.
    \end{align}
    Note that the above proposal distribution \underline{$\boldz \sim p_{\boldz}(\boldz), \boldx = \gen(\boldz)$} is the same as $p_{\boldx}(\boldx)$ by definition. Based on Lemma 1 in \cite{Grathwohl2020Your} and Lemma 1 in \cite{nie2021controllable}, we arrive at:
    \begin{align}
        p(\boldx|\mathcal{C}) &= \frac{p_{\boldx}(\boldx)r(\boldx)}{\mathbb{E}_{\boldx' \sim p_{\boldx}(\boldx')}[r(\boldx')]} \\
        &= \frac{p_{\boldx}(\boldx)e^{-E_{\mathcal{C}}(\boldx)} / (M_{\mathcal{C}}\cdot Z)}{\mathbb{E}_{\boldx' \sim p_{\boldx}(\boldx')}[e^{-E_{\mathcal{C}}(\boldx')} / (M_{\mathcal{C}}\cdot Z)]} \\
        &= \frac{p_{\boldx}(\boldx)e^{-E_{\mathcal{C}}(\boldx)}}{\mathbb{E}_{\boldx' \sim p_{\boldx}(\boldx')}[e^{-E_{\mathcal{C}}(\boldx')}]} \\
        &= \frac{p_{\boldx}(\boldx)e^{-E_{\mathcal{C}}(\boldx)}}{Z_X}, \quad Z_X = \int_{\boldx'} p_{\boldx}(\boldx')e^{-E_{\mathcal{C}}(\boldx')} d\boldx'.
    \end{align}
\end{proof}

\subsection{Derivation of Eq.~(\ref{eq:kl-single-image}): Approximating EBM with INN}
\label{subapp:approx-ebm}
The full derivation of Eq.~(\ref{eq:kl-single-image}) is given by:
\begin{equation}
\label{eq:kl-single-image-proof}
\begin{split}
    &\quad\ \mathbb{D}_{\mathsf{KL}}(\pzsingle(\boldz)\|p(\boldz|\mathcal{C})) \\
    &= \mathbb{E}_{\boldz \sim \pzsingle(\boldz)}\Big[\log \frac{\pzsingle(\boldz)}{p(\boldz|\mathcal{C})}\Big] \\
    &= \mathbb{E}_{\boldz \sim \pzsingle(\boldz), \boldx = \gen(\boldz)}\Big[\log \frac{\pzsingle(\boldz)}{p_{\boldz}(\boldz)e^{-E_{\mathcal{C}}(\boldx)}/Z}\Big] \\
    &= \mathbb{E}_{\boldeps \sim \mathcal{N}(\bm{0}, \bm{I}), \boldz = \inn(\boldeps), \boldx = \gen(\boldz)}\Big[\log \mathcal{N}(\boldeps|\bm{0}, \bm{I}) - \log |\det(\frac{\partial \inn}{\partial \boldeps})| \\
    &\quad\quad\quad\quad\quad\quad\quad\quad\quad\quad\quad\quad\quad\quad\quad\quad\quad\quad\quad- \log p_{\boldz}(\boldz) + E_{\mathcal{C}}(\boldx) + \log Z\Big] \\
    &= \mathbb{E}_{\boldeps \sim \mathcal{N}(\bm{0}, \bm{I}), \boldz = \inn(\boldeps), \boldx = \gen(\boldz)}\Big[- \log |\det(\frac{\partial \inn}{\partial \boldeps})| - \log p_{\boldz}(\boldz) \\
    &\quad\quad\quad\quad\quad\quad\quad\quad\quad\quad\quad\quad\quad\quad\quad\quad\quad\quad\quad + E_{\mathcal{C}}(\boldx)\Big] - \mathbb{H}_{\mathcal{N}(\bm{0}, \bm{I})} + \log Z.
\end{split}
\end{equation}

\subsection{Derivation of Eq.~(\ref{eq:objective-single-image-cgan}): Extension to Generative Models with a Class-Embedding Space}
\label{subapp:extension-to-cgan}

The full derivation of  Eq.~(\ref{eq:objective-single-image-cgan}) is given by:
\begin{equation}
\label{eq:kl-single-image-cgan-proof}
\begin{split}
    &\quad\ \mathbb{D}_{\mathsf{KL}}(\pzsingle(\boldz, \boldy)\|p(\boldz, \boldy|\mathcal{C})) \\
    &= \mathbb{E}_{(\boldz, \boldy) \sim \pzsingle(\boldz, \boldy)}\Big[\log \frac{\pzsingle(\boldz, \boldy)}{p(\boldz, \boldy|\mathcal{C})}\Big] \\
    &= \mathbb{E}_{(\boldz, \boldy) \sim \pzsingle(\boldz, \boldy), \boldx = \gen(\boldz, \boldy)}\Big[\log \frac{\pzsingle(\boldz, \boldy)}{p_{\boldz, \boldy}(\boldz, \boldy)e^{-E_{\mathcal{C}}(\boldx)}/Z}\Big] \\
    &= \mathbb{E}_{(\boldz, \boldy) \sim \pzsingle(\boldz, \boldy), \boldx = \gen(\boldz, \boldy)}\Big[\log \frac{\pzsingle(\boldz, \boldy)}{p_{\boldz}(\boldz)p_{\boldy}(\boldy)e^{-E_{\mathcal{C}}(\boldx)}/Z}\Big] \quad \text{($\boldz$ and $\boldy$ are independent)} \\
    &= \mathbb{E}_{\boldeps \sim \mathcal{N}(\bm{0}, \bm{I}), \bm{\xi} \sim \mathcal{N}(\bm{\mu}, \bm{\sigma}^2\bm{I}), \boldz = \inn(\boldeps), \boldy = h_{\bm{\theta}}(\bm{\xi}), \boldx = \gen(\boldz, \boldy)}\Big[\log \mathcal{N}(\boldeps|\bm{0}, \bm{I}) - \log |\det(\frac{\partial \inn}{\partial \boldeps})| \\
    &\quad\quad\quad + \log \mathcal{N}(\bm{\xi}|\bm{\mu}, \bm{\sigma}^2\bm{I}) - \log |\det(\frac{\partial h_{\bm{\theta}}}{\partial \bm{\xi}})| - \log p_{\boldz}(\boldz) - \log p_{\boldy}(\boldy) + E_{\mathcal{C}}(\boldx) + \log Z\Big] \\
    &= \mathbb{E}_{\boldeps \sim \mathcal{N}(\bm{0}, \bm{I}), \bm{\xi} \sim \mathcal{N}(\bm{\mu}, \bm{\sigma}^2\bm{I}), \boldz = \inn(\boldeps), \boldy = h_{\bm{\theta}}(\bm{\xi}), \boldx = \gen(\boldz, \boldy)}\Big[- \log |\det(\frac{\partial \inn}{\partial \boldeps})| - \log |\det(\frac{\partial h_{\bm{\theta}}}{\partial \bm{\xi}})| \\
    &\quad\quad\quad\quad\quad\quad\quad\quad - \log p_{\boldz}(\boldz) - \log p_{\boldy}(\boldy) + E_{\mathcal{C}}(\boldx) \Big] - \mathbb{H}_{\mathcal{N}(\bm{0}, \bm{I})} - \mathbb{H}_{\mathcal{N}(\bm{\mu}, \bm{\sigma}^2\bm{I})} + \log Z.
\end{split}
\end{equation}

\subsection{Extension to Conditional INN}
\label{subapp:extension-conditional-inn}
Algorithm~\ref{alg:approximate-ebm-conditional} illustrates how we extend the training algorithm (Algorithm~\ref{alg:approximate-ebm}) to conditional INN, which helps generalize to controls specified by continuous values $\bm{\rho}$.
The training objective becomes 
\begin{equation}
\label{eq:objective-kl-single-image-conditional}
    \mathop{\arg\min}_{\bm{\theta}} \mathbb{E}_{\bm{\rho} \sim p_{\bm{\rho}}(\bm{\rho})}\big[\mathbb{D}_{\mathsf{KL}}\big(\pzsingle(\boldz|\bm{\rho})\|p(\boldz|{\mathcal{C}_{\bm{\rho}}})\big)\big],
\end{equation}
where $\mathcal{C}_{\bm{\rho}}$ means that the control is specified by value $\bm{\rho}$. The derivation of Algorithm~\ref{alg:approximate-ebm-conditional} is given by:
\begin{equation}
\label{eq:kl-single-image-conditional}
\begin{split}
    &\quad\ \mathbb{E}_{\bm{\rho} \sim p_{\bm{\rho}}(\bm{\rho})}\big[\mathbb{D}_{\mathsf{KL}}\big(\pzsingle(\boldz|\bm{\rho})\|p(\boldz|{\mathcal{C}_{\bm{\rho}}})\big)\big] \\
    &= \mathbb{E}_{\bm{\rho} \sim p_{\bm{\rho}}(\bm{\rho}), \boldz \sim \pzsingle(\boldz|\bm{\rho})}\Big[\log \frac{\pzsingle(\boldz|\bm{\rho})}{p(\boldz|{\mathcal{C}_{\bm{\rho}}})}\Big] \\
    &= \mathbb{E}_{\bm{\rho} \sim p_{\bm{\rho}}(\bm{\rho}), \boldz \sim \pzsingle(\boldz|\bm{\rho}), \boldx = \gen(\boldz)}\Big[\log \frac{\pzsingle(\boldz|\bm{\rho})}{p_{\boldz}(\boldz)e^{-E_{\mathcal{C}_{\bm{\rho}}}(\boldx)}/Z}\Big] \\
    &= \mathbb{E}_{\bm{\rho} \sim p_{\bm{\rho}}(\bm{\rho}), \boldeps \sim \mathcal{N}(\bm{0}, \bm{I}), \boldz = \inn(\boldeps, \bm{\rho}), \boldx = \gen(\boldz)}\Big[\log \mathcal{N}(\boldeps|\bm{0}, \bm{I}) - \log |\det(\frac{\partial \inn}{\partial \boldeps})| \\
    &\quad\quad\quad\quad\quad\quad\quad\quad\quad\quad\quad\quad\quad\quad\quad\quad\quad\quad\quad\quad - \log p_{\boldz}(\boldz) + E_{\mathcal{C}_{\bm{\rho}}}(\boldx) + \log Z\Big] \\
    &= \mathbb{E}_{\bm{\rho} \sim p_{\bm{\rho}}(\bm{\rho}), \boldeps \sim \mathcal{N}(\bm{0}, \bm{I}), \boldz = \inn(\boldeps, \bm{\rho}), \boldx = \gen(\boldz)}\Big[- \log |\det(\frac{\partial \inn}{\partial \boldeps})| - \log p_{\boldz}(\boldz) \\
    &\quad\quad\quad\quad\quad\quad\quad\quad\quad\quad\quad\quad\quad\quad\quad\quad\quad\quad\quad\quad + E_{\mathcal{C}_{\bm{\rho}}}(\boldx)\Big] - \mathbb{H}_{\mathcal{N}(\bm{0}, \bm{I})} + \log Z.
\end{split}
\end{equation}

%%%%%%%%%%%%%%%%%%%%%%%%%%%%%%%%%%%%%%%%%%%%%%%%%%%%%%%%%%%%%%%%%%%%%%%%%%%%%%
%\SetCustomAlgoRuledWidth{0.92\linewidth}
\begin{algorithm}[t]
\DontPrintSemicolon
	\While{not converged}{		
		\smallskip
		1. Sample $\boldeps \sim \mathcal{N}(\bm{0}, \bm{I}), \bm{\rho} \sim p_{\bm{\rho}}(\bm{\rho})$\;
		2. Map $\boldeps$ to latent code $\boldz = \inn(\boldeps, \bm{\rho})$\;
		3. Map $\boldz$ to an image $\boldx = \gen(\boldz)$\;
		4. Optimize $\boldtheta$ with gradient $\displaystyle \nabla_{\boldtheta} \Big(- \log |\det(\frac{\partial \inn}{\partial \boldeps})| - \log p_{\boldz}(\boldz) + E_{\mathcal{C}_{\bm{\rho}}}(\boldx)\Big)$\;
	}
	\caption{Extension of Algorithm~\ref{alg:approximate-ebm} to Conditional INN}
	\label{alg:approximate-ebm-conditional}
\end{algorithm}
%%%%%%%%%%%%%%%%%%%%%%%%%%%%%%%%%%%%%%%%%%%%%%%%%%%%%%%%%%%%%%%%%%%%%%%%%%%%%%

\subsection{Derivation for Latent-Space Moment Constraint}
\label{subapp:distributional-derivation}

Given a mapping $\boldgamma: \mathcal{X} \to \mathbb{R}^{K}$, moment constraint \cite{Csiszr2004InformationTA,khalifa2021a} defines the target distribution $p(\boldx|\mathcal{C})$ as
\begin{equation}
\label{eq:moment-constraint-1-copy}
    p(\boldx|\mathcal{C}) = \underbrace{\mathop{\arg\min}_{p(\boldx|\mathcal{C})} \mathbb{D}_{\mathsf{KL}}(p(\boldx|\mathcal{C})\|p_{\boldx}(\boldx))}_{\text{Deviation from the pre-trained distribution}}, \quad\text{s.t. } \underbrace{\mathbb{E}_{\boldx \sim p(\boldx|\mathcal{C})}\big[\boldgamma(\boldx)\big] = \boldmu}_{\text{Moment constraint}},
\end{equation}
where $\boldmu$ is the user-specified constraint. 
Examples are provided in the main text, and we omit them here for brevity. 
% For example, if we want to generate images that distribute uniformly across races, we may use a race classifier as $\boldgamma$ and define $\boldmu = \big(|\mathcal{A}|^{-1}, \ldots, |\mathcal{A}|^{-1}\big)$ where $\mathcal{A}$ is the set of races. 
\cite{Csiszr2004InformationTA} showed that Eq.~(\ref{eq:moment-constraint-1-copy}) can be approximated to an arbitrary precision by 
\begin{equation}
\label{eq:moment-constraint-2-image}
    p(\boldx|\mathcal{C}) \propto p_{\boldx}(\boldx) \exp\big(\hat{\bm{\beta}}^\top \bm{\gamma}(\boldx)\big),
\end{equation}
where $\hat{\bm{\beta}}$ needs to be computed. \cite{khalifa2021a} estimates $\hat{\bm{\beta}}$ by solving the following regression problem:
\begin{equation}
\label{eq:moment-constraint-3-image}
    \hat{\bm{\beta}} = \mathop{\arg\min}_{\bm{\beta}} \mathbb{E}_{\boldx^{(1)}, \ldots, \boldx^{(N)} \overset{\text{i.i.d.}}{\sim}p_{\boldx}(\boldx)}  \bigg\|\frac{\sum_{j=1}^{N}\exp\big(\bm{\beta}^\top \bm{\gamma}(\boldx^{(j)})\big)\boldgamma(\boldx^{(j)})}{\sum_{j'=1}^{N}\exp\big(\bm{\beta}^\top \bm{\gamma}(\boldx^{(j')})\big)} - \boldmu \bigg\|_2^2. 
\end{equation}
In \cite{khalifa2021a}, sampling $\boldx \sim p_{\boldx}(\boldx)$ is straightforward since they focus on autoregressive language models. 
In the context of latent-variable vision generative models, where $\boldx \sim p_{\boldx}(\boldx)$ is implicitly defined as \underline{$\boldz \sim p_{\boldz}(\boldz), \boldx = \gen(\boldz)$}, 
Eq.~(\ref{eq:moment-constraint-3-image}) is equivalent to
\begin{equation}
\label{eq:moment-constraint-3-copy}
    \hat{\bm{\beta}} = \mathop{\arg\min}_{\bm{\beta}} \mathbb{E}_{\boldz^{(1)}, \ldots, \boldz^{(N)} \overset{\text{i.i.d.}}{\sim}p_{\boldz}(\boldz), \boldx^{(j)} = \gen(\boldz^{(j)}) }  \bigg\|\frac{\sum_{j=1}^{N}\exp\big(\bm{\beta}^\top \bm{\gamma}(\boldx^{(j)})\big)\boldgamma(\boldx^{(j)})}{\sum_{j'=1}^{N}\exp\big(\bm{\beta}^\top \bm{\gamma}(\boldx^{(j')})\big)} - \boldmu \bigg\|_2^2. 
\end{equation}
Finally, based on Proposition~\ref{prep:latent-space-ebm}, we can generalize the moment constraint to the latent space as
\begin{align}
\label{eq:moment-constraint-2-copy}
    &p(\boldz|\mathcal{C}) = \frac{p_{\boldz}(\boldz) \exp\Big(\hat{\bm{\beta}}^\top \bm{\gamma}\big(\gen(\boldz)\big)\Big)}{Z}, \quad Z = \int_{\boldz'} p_{\boldz}(\boldz')\exp\Big(\hat{\bm{\beta}}^\top \bm{\gamma}\big(\gen(\boldz')\big)\Big) d\boldz'.
\end{align}
Note that Eq.~(\ref{eq:moment-constraint-3-copy}) and Eq.~(\ref{eq:moment-constraint-2-copy}) are verbatim copies of Eq.~(\ref{eq:moment-constraint-3}) and Eq.~(\ref{eq:moment-constraint-2}), respectively. 

%%%%%%%%%%%%%%%%%%%%%%%%%%%%%%%%%%%%%%%%%%%%%%%%%%%%%%%%%%%%%%%%%%%%%%%%%%%%%%
%\SetCustomAlgoRuledWidth{0.92\linewidth}
\begin{algorithm}[!ht]
\DontPrintSemicolon
	\While{not converged}{		
		\smallskip
		1. Sample $\boldeps \sim \mathcal{N}(\bm{0}, \bm{I}), \bm{\rho} \sim p_{\bm{\rho}}(\bm{\rho})$\;
		2. Map $\boldeps$ to latent codes $\boldz = \inn(\boldeps, \bm{\rho})$ and $\boldz_0 = \inn(\boldeps, \bm{\rho}_0)$\;
		3. Map $\boldz$ to an image $\boldx = \gen(\boldz)$ and $\boldz_0$ to an image $\boldx_0 = \gen(\boldz_0)$\;
		4. Optimize $\boldtheta$ with gradient\;
		\begin{displaymath}
		    \nabla_{\boldtheta} \Big(- \log |\det(\frac{\partial \inn}{\partial \boldeps})| - \log p_{\boldz}(\boldz) + E_{\mathcal{C}_{\bm{\rho}}}(\boldx) + \lambda_{\text{ID}} E_{\text{ID}}(\boldx_0, \boldx)\Big)
		\end{displaymath}
	}
	\caption{Extension of Algorithm~\ref{alg:approximate-ebm-conditional} with ID Energy (Section~\ref{subsec:expression-pose-experiments})}
	\label{alg:approximate-ebm-conditional-id}
\end{algorithm}
%%%%%%%%%%%%%%%%%%%%%%%%%%%%%%%%%%%%%%%%%%%%%%%%%%%%%%%%%%%%%%%%%%%%%%%%%%%%%%

\subsection{Inverse Graphics Control with Identity Energy}
\label{subapp:inverse-graphics-detail}
This subsection provides more detail on how to use an inverse graphics model, DECA \cite{Feng2021LearningAA}, to control the pose of faces generated by StyleGAN2 trained on FFHQ 1024$^2$, as introduced in Section~\ref{subsec:expression-pose-experiments}. 
Recall that we use the conditional INN extension, detailed in Algorithm~\ref{alg:approximate-ebm-conditional}, for this experiment. 
To enable generating different poses of the same identity, we added an additional identity energy using the IR-SE50 model \cite{Deng2019ArcFaceAA}. 
The training algorithm is detailed in Algorithm~\ref{alg:approximate-ebm-conditional-id}. 
Specifically, we generate a canonical latent code $\boldz_0 = \inn(\boldeps, \bm{\rho}_0)$ for each $\boldeps$, where $\bm{\rho}_0$ is the canonical pose.
The latent code $\boldz$ and the canonical latent code $\boldz_0$ are mapped to the image $\boldx$ and the canonical image $\boldx_0$. 
The identity energy in Eq.~(\ref{eq:identity-energy}), copied below, encourages the embeddings of the two images to be similar:
\begin{equation}
\label{eq:identity-energy-copy}
    E_{\text{ID}}(\boldx_0, \boldx) = 1 - \cos\big\langle R(\boldx_0), R(\boldx)\big\rangle, \quad \boldx_0 = \gen(\boldz_0), \boldx = \gen(\boldz).
\end{equation}

\subsection{Experimental Details}
\label{subapp:experiment-details}

Our INN architecture contains 8 blocks. Each block consists of a soft permutation of channels \cite{Ardizzone2019GuidedIG}, an affine coupling layer \cite{DinhSB17}, and an ActNorm layer \cite{Kingma2018GlowGF}. In the affine coupling layer, we model the sub-network as an MLP with one hidden layer, where the hidden dimension is 256 and the non-linear activation is LeakyReLU$_{0.1}$. For the prior distribution $p_{\bm{\rho}}(\bm{\rho})$ of the pose parameter $\bm{\rho}$ in the inverse graphics experiments, we sample the $x$-axis and $y$-axis rotations relative to a canonical pose $\bm{\rho}_0$, whose $x$-axis and $y$-axis rotations are $0.3$ and $0$, respectively. The relative pose's $x$-axis and $y$-axis rotations are uniformly sampled from $[-(1/9)\pi, (1/9)\pi]$. Each experiment was run on an NVIDIA RTX A4000 GPU (with 16G memory). 
Our code implementation is based on the PyTorch framework and can be 
be found at \opensource.\anonymoustext{We will make our code, configurations, and experimental instructions publicly available. }

\section{Additional Results for Image Synthesis based on Text Description}
\label{app:clip-samples}

\begin{figure}
\centering
   
    \subfigure[p\underline{hoto of a cat with closed e}y\underline{es} (AFHQ-Cats)]{
        \includegraphics[width=0.48\linewidth]{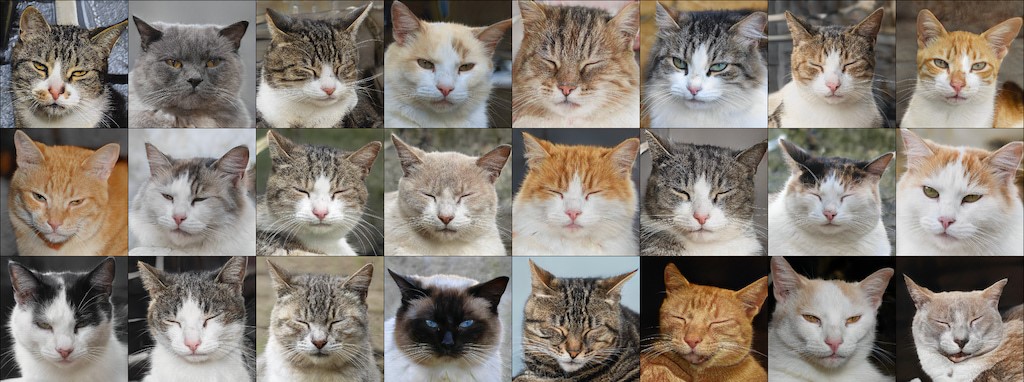}
    }
    \subfigure[p\underline{hoto of a cat with e}y\underline{es wide o}p\underline{en} (AFHQ-Cats)]{
        \includegraphics[width=0.48\linewidth]{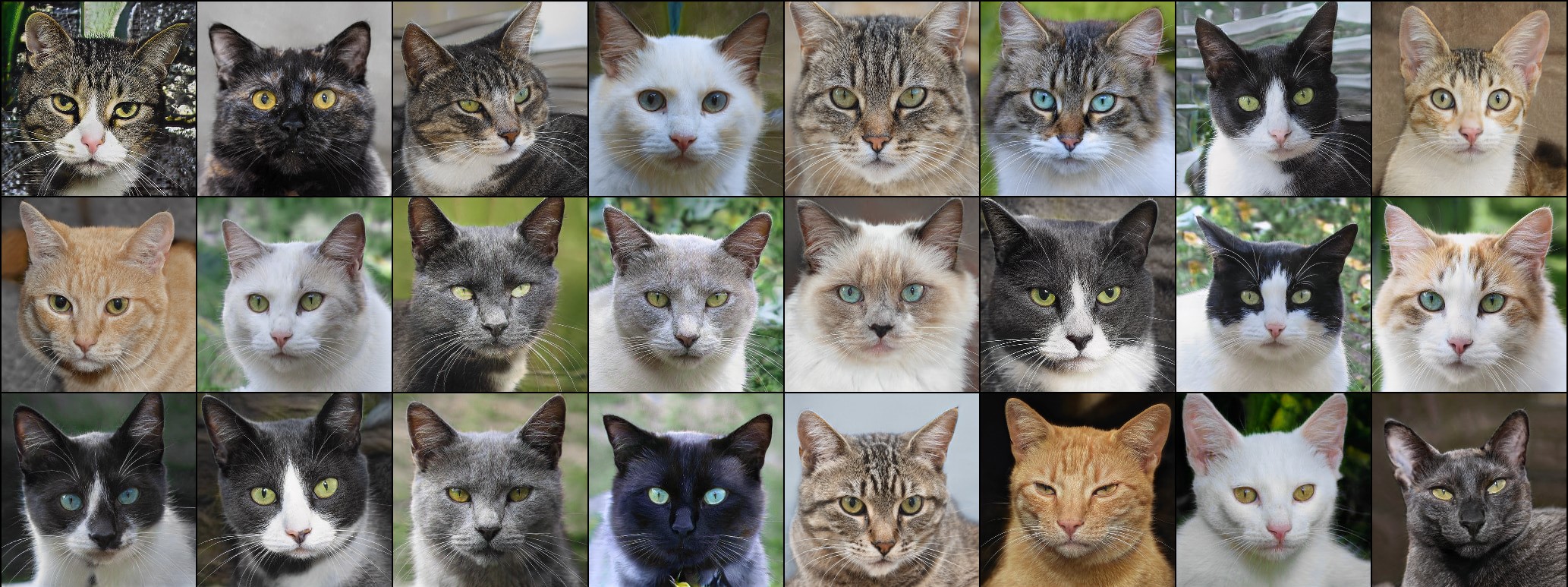}
    }
    \subfigure[p\underline{hoto of a church durin}g\underline{ da}y (LSUN-Churches)]{
        \includegraphics[width=0.48\linewidth]{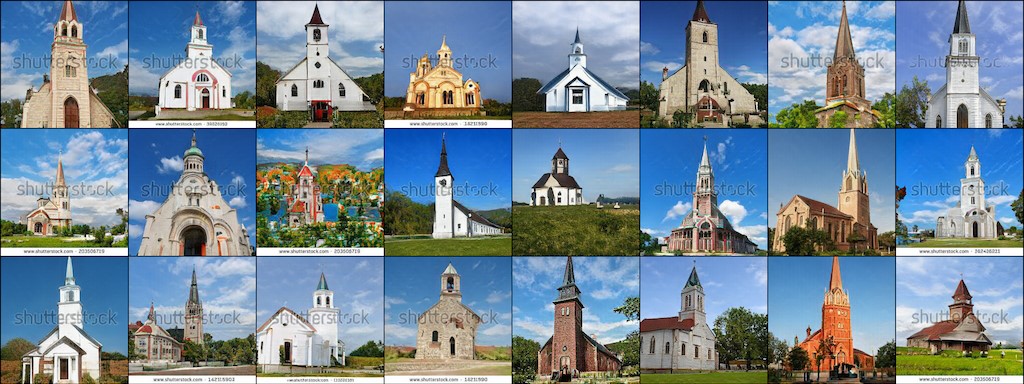}
    }
    \subfigure[p\underline{hoto of a church at ni}g\underline{ht} (LSUN-Churches)]{
        \includegraphics[width=0.48\linewidth]{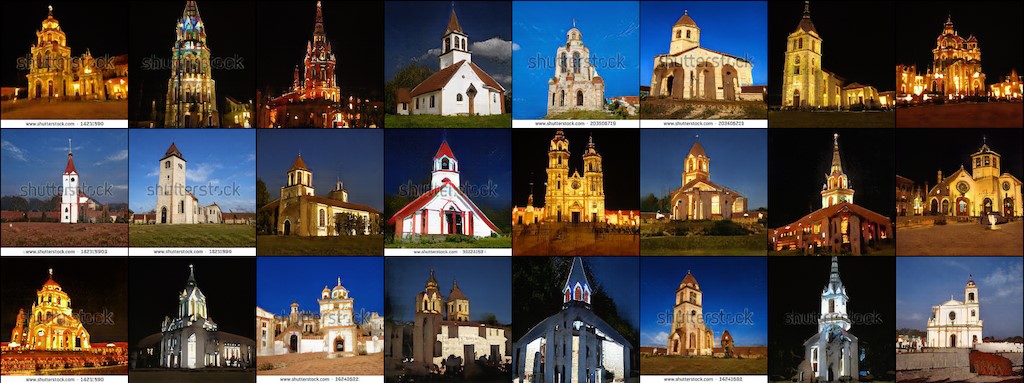}
    }
    \subfigure[p\underline{hoto of a ha}ppy\underline{ }p\underline{erson} (FFHQ)]{
        \includegraphics[width=0.48\linewidth]{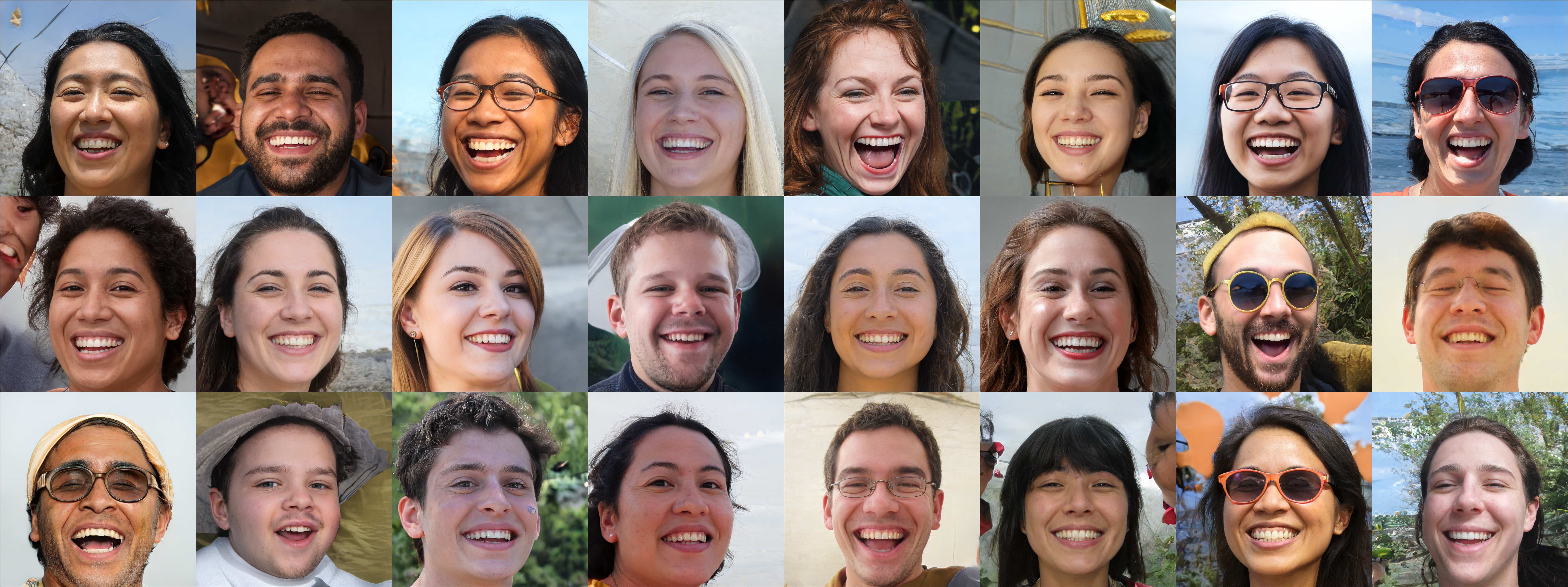}
    }
    \subfigure[p\underline{hoto of a sad }p\underline{erson} (FFHQ)]{
        \includegraphics[width=0.48\linewidth]{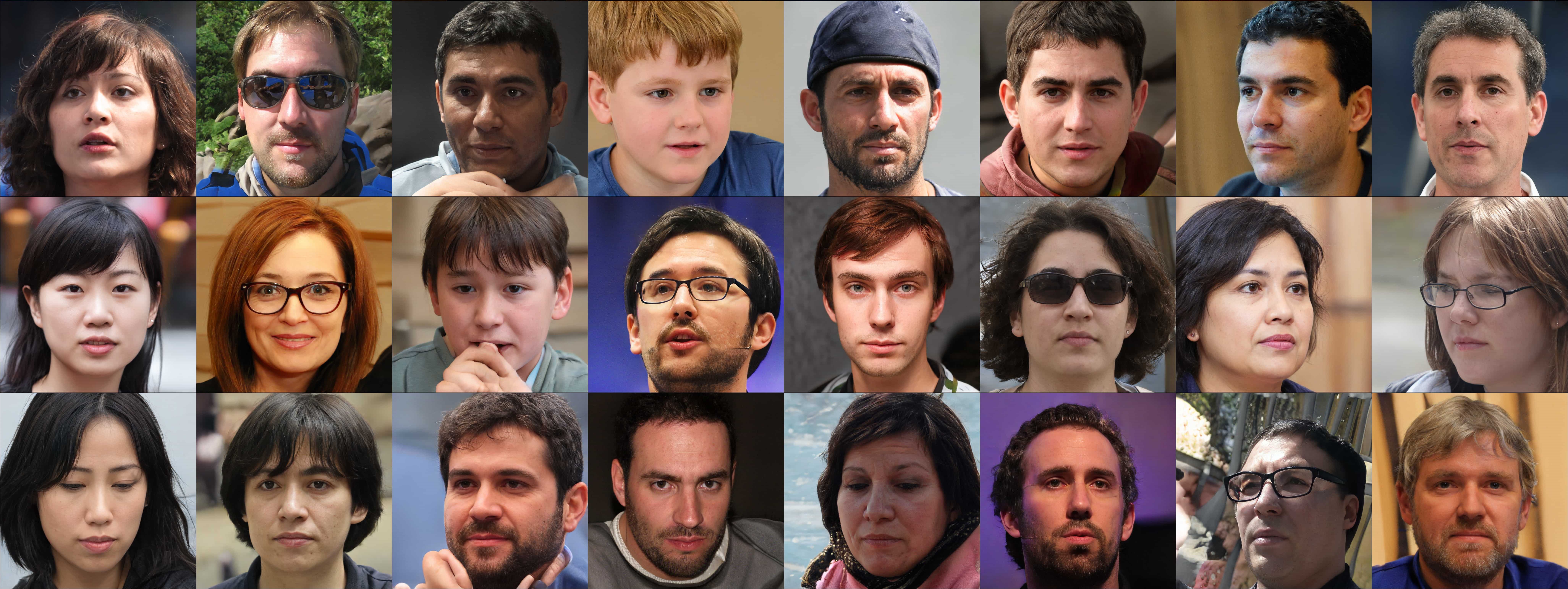}
    }
    \subfigure[p\underline{hoto of a woman}\underline{ with} g\underline{lasses} (FFHQ)]{
        \includegraphics[width=0.85\linewidth]{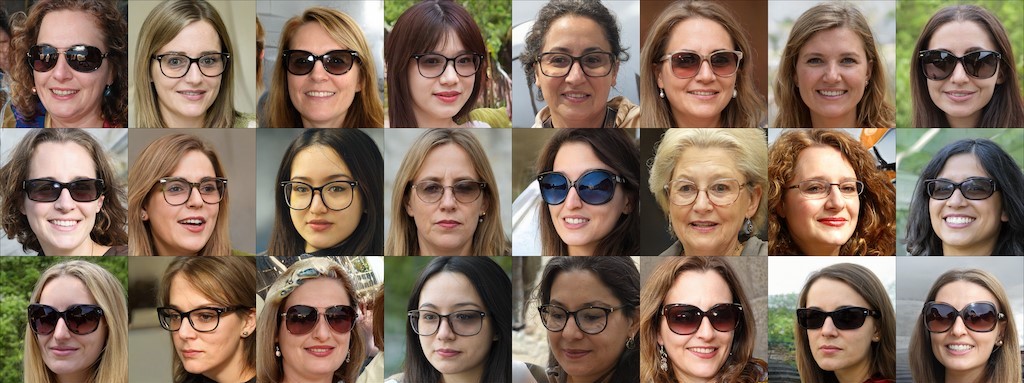}
    }
    \caption{\label{fig:more_clip} Additional results for image synthesis based on a text description guided by the CLIP model. As the pre-trained generative model, we use StyleGAN2 trained on FFHQ, AFHQ-Cats and LSUN-Churches datasets. The captions are the text descriptions given to the CLIP model. }
\end{figure}

Figure~\ref{fig:more_clip} presents additional results of \promptgen for image synthesis from text, using the CLIP model as control. As the pre-trained generative model, we explore StyleGAN2 trained on AFHQ-Cats \cite{choi2020starganv2}, LSUN-Churches \cite{yu15lsun}, and FFHQ \cite{Karras2019ASG}.

%%%%%%%%%%%%%%%%%%%%%%%%%%%%%%%%%%%%%%%%%%%%%%%%%%%%%%%%%%%%%%%%%%%%%%%%%%%%%%
%%%%%%%%%%%%%%%%%%%%%%%%%%%%%%%%%%%%%%%%%%%%%%%%%%%%%%%%%%%%%%%%%%%%%%%%%%%%%%

\section{Additional Results for De-Biasing Generative Models}
\label{app:debias-samples}

%%%%%%%%%%%%%%%%%%%%%%%%%%%%%%%%%%%%%%%%%%%%%%%%%%%%%%%%%%%%%%%%%%%%%%%%%%%%%%
\begin{table}
%\small
    \caption{Quantitative results for de-biasing categorical attributes (Figure~\ref{fig:debias-results}). See details in Appendix~\ref{app:debias-samples}. \promptgen de-biases StyleGAN2 in terms of race, age, and gender. }
    \label{tab:debias-results}
    \centering
    \begin{adjustbox}{width=0.65\linewidth}
    \begin{tabular}{@{}lccccc@{}}
        \toprule
        & \multicolumn{2}{c}{FFHQ} & \multicolumn{3}{c}{MetFaces}  \\
        \cmidrule(lr){2-3} \cmidrule(l){4-6}
        & $\mathbb{D}_\text{KL}^{\text{race}}$\down & $\mathbb{D}_\text{KL}^{\text{age}}$\down & $\mathbb{D}_\text{KL}^{\text{race}}$\down & $\mathbb{D}_\text{KL}^{\text{age}}$\down & $\mathbb{D}_\text{KL}^{\text{gender}}$\down \\
        \midrule
        StyleGAN2                           & 0.860         & 0.597         & 1.624             & 0.546             & 0.019 \\
        \midrule
        \promptgen (ours; $\lambda = 1$)    & 0.286         & 0.357         & 0.687             & 0.397             & \bf 0.000 \\
        \promptgen (ours; $\lambda = 2$)    & \bf 0.099     & \bf 0.172     & \bf 0.189         & \bf 0.247         & 0.005  \\
        %\promptgen (ours; $\lambda = 2$)    & 0.099         & \bf 0.172     & \bf 0.189         & 0.247             & 0.005  \\
        %\promptgen (ours; $\lambda = 3$)    & \bf 0.093     & 0.261         & 0.202             & \bf 0.120         & 0.020  \\
        \bottomrule
    \end{tabular}
    \end{adjustbox}
\end{table}
%%%%%%%%%%%%%%%%%%%%%%%%%%%%%%%%%%%%%%%%%%%%%%%%%%%%%%%%%%%%%%%%%%%%%%%%%%%%%%

\begin{figure}[t]
\centering
    \subfigure[Pre-trained StyleGAN2]{
        \includegraphics[width=0.63\linewidth]{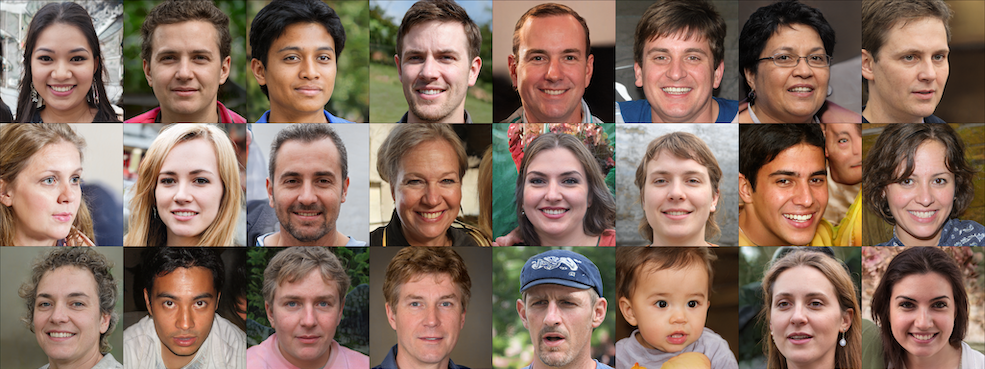}
    }
    \subfigure[\promptgen (race de-biasing; $\lambda = 1$)]{
        \includegraphics[width=0.48\linewidth]{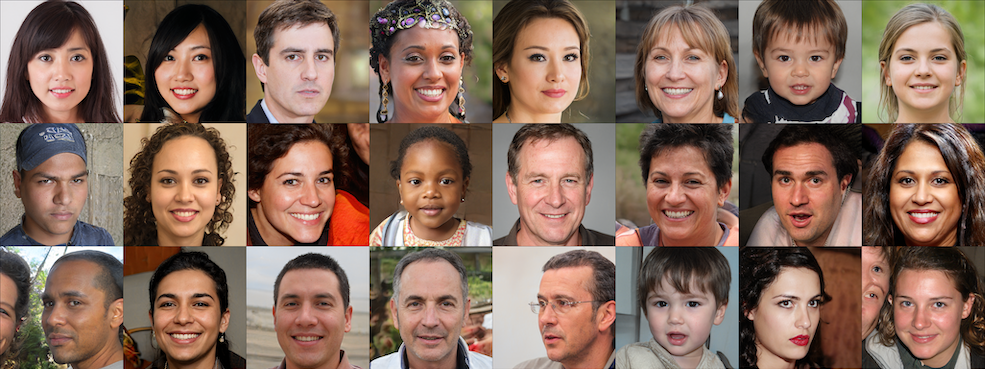}
    }
    \subfigure[\promptgen (race de-biasing; $\lambda = 2$)]{
        \includegraphics[width=0.48\linewidth]{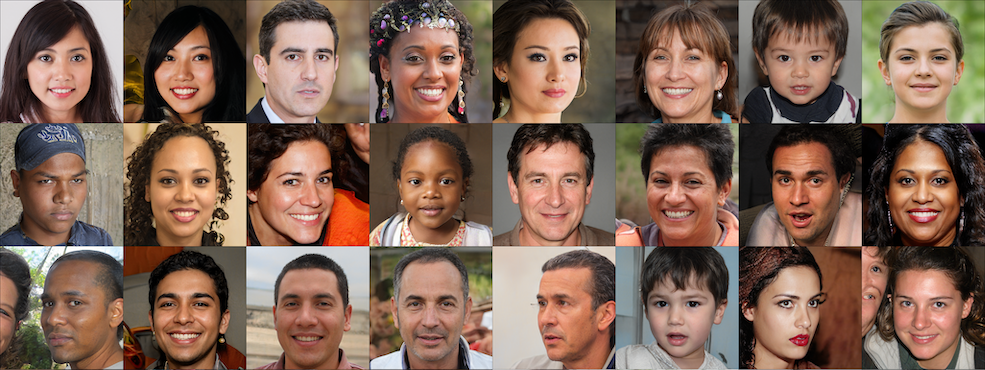}
    }
    \subfigure[\promptgen (age de-biasing; $\lambda = 1$)]{
        \includegraphics[width=0.48\linewidth]{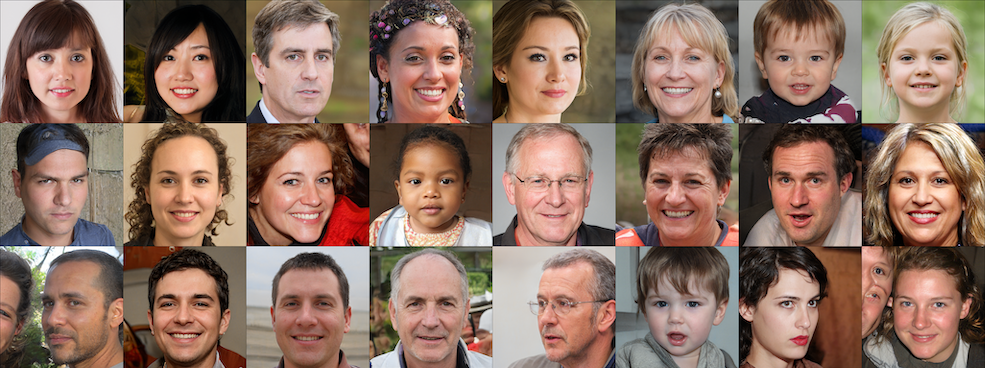}
    }
    \subfigure[\promptgen (age de-biasing; $\lambda = 2$)]{
        \includegraphics[width=0.48\linewidth]{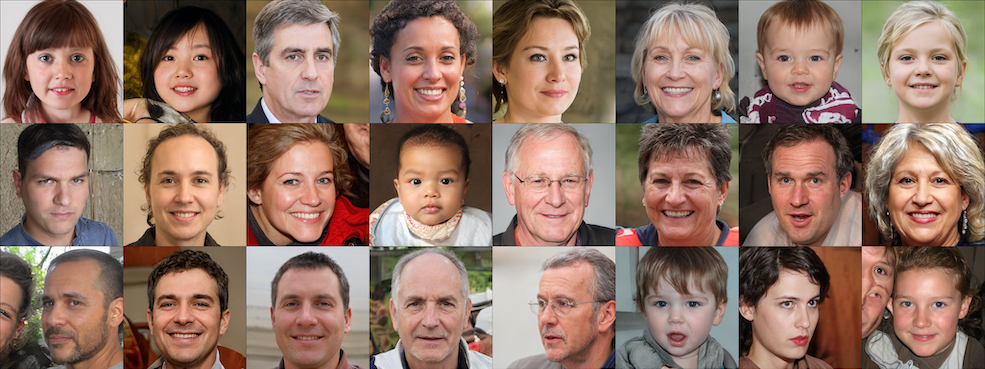}
    }
    \subfigure[Race distribution]{
        \includegraphics[width=0.48\linewidth]{figures/FFHQ-Race.pdf}
    }
    \subfigure[Age distribution]{
        \includegraphics[width=0.48\linewidth]{figures/FFHQ-Age.pdf}
    }
\caption{\label{fig:debias-results-ffhq-full} Using our moment constraint, \promptgen de-biases the racial and age distributions of StyleGAN2 trained on FFHQ with truncation $\psi=0.7$. All synthesized images are $1024^2$ in resolution and resized for visualization. We fixed the random seed for \promptgen, so please zoom in to see detailed differences between images.}
\end{figure}

\begin{figure}[t]
\centering
    \subfigure[Pre-trained StyleGAN2]{
        \includegraphics[width=0.63\linewidth]{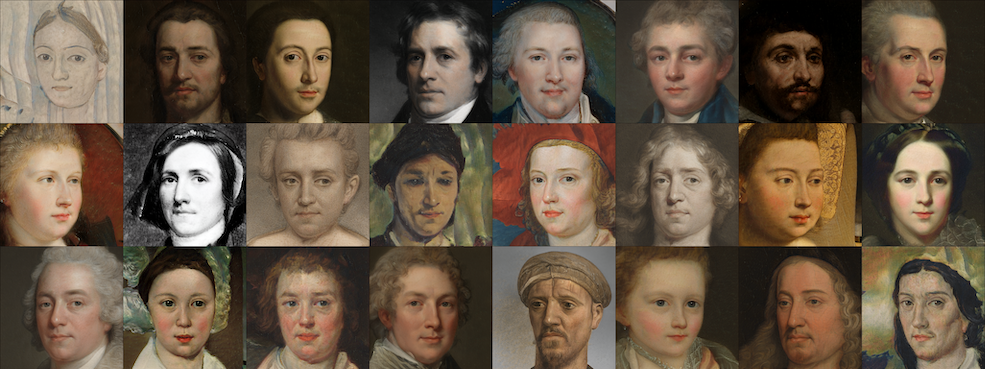}
    }
    \subfigure[\promptgen (race de-biasing; $\lambda = 1$)]{
        \includegraphics[width=0.48\linewidth]{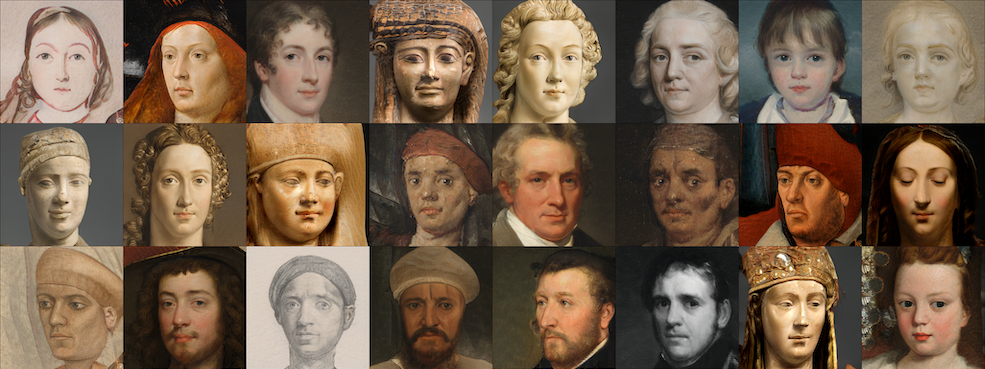}
    }
    \subfigure[\promptgen (race de-biasing; $\lambda = 2$)]{
        \includegraphics[width=0.48\linewidth]{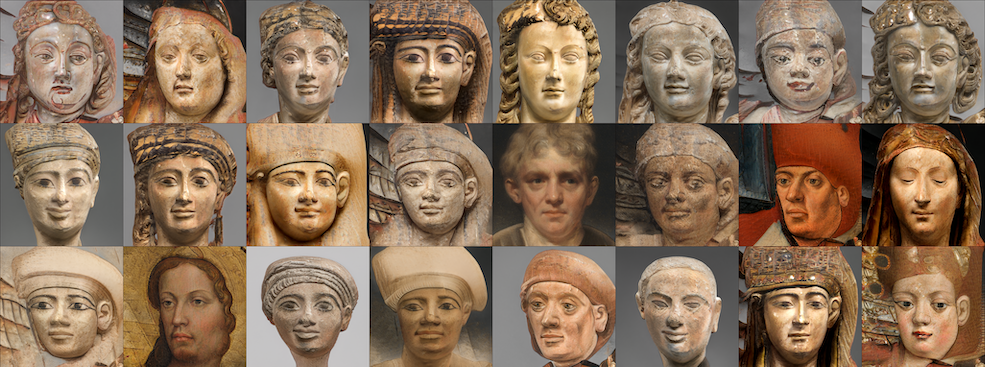}
    }
    %\subfigure[\promptgen (age de-biasing; $\lambda = 1$)]{
    %    \includegraphics[width=0.48\linewidth]{figures/metfaces_age_debiased_lambda1_resized.png}
    %}
    \subfigure[\promptgen (age de-biasing; $\lambda = 2$)]{
        \includegraphics[width=0.48\linewidth]{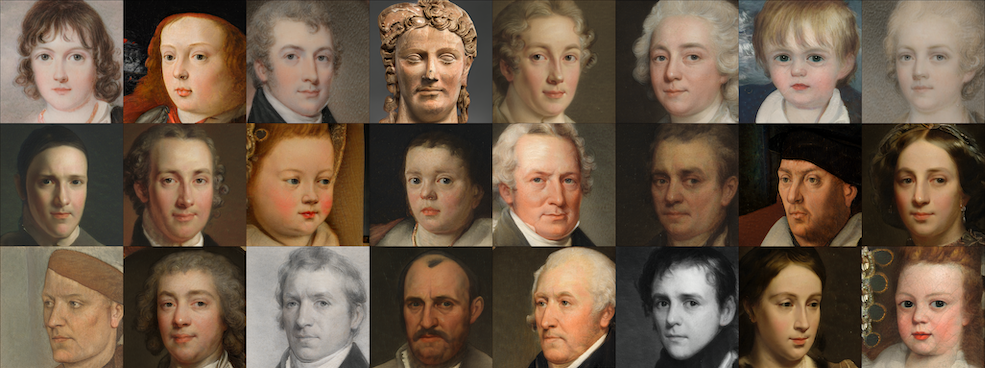}
    }
    %\subfigure[\promptgen (gender de-biasing; $\lambda = 1$)]{
    %    \includegraphics[width=0.48\linewidth]{figures/metfaces_gender_debiased_lambda1_resized.png}
    %}
    \subfigure[\promptgen (gender de-biasing; $\lambda = 2$)]{
        \includegraphics[width=0.48\linewidth]{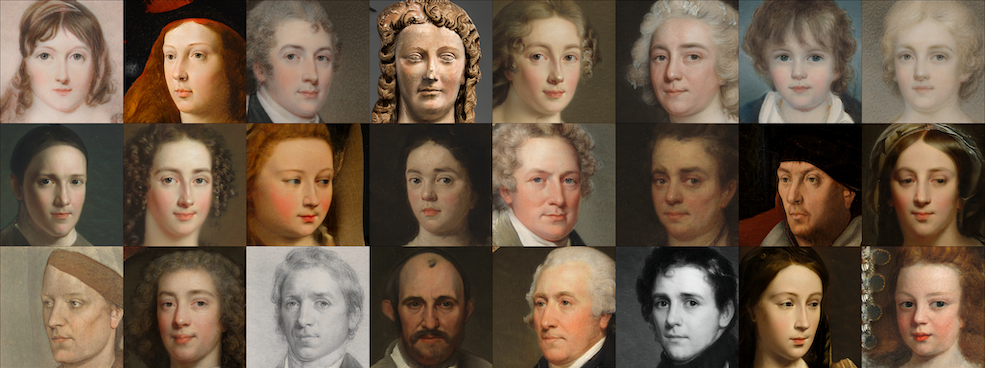}
    }
    \subfigure[Gender distribution]{
        \includegraphics[width=0.316\linewidth]{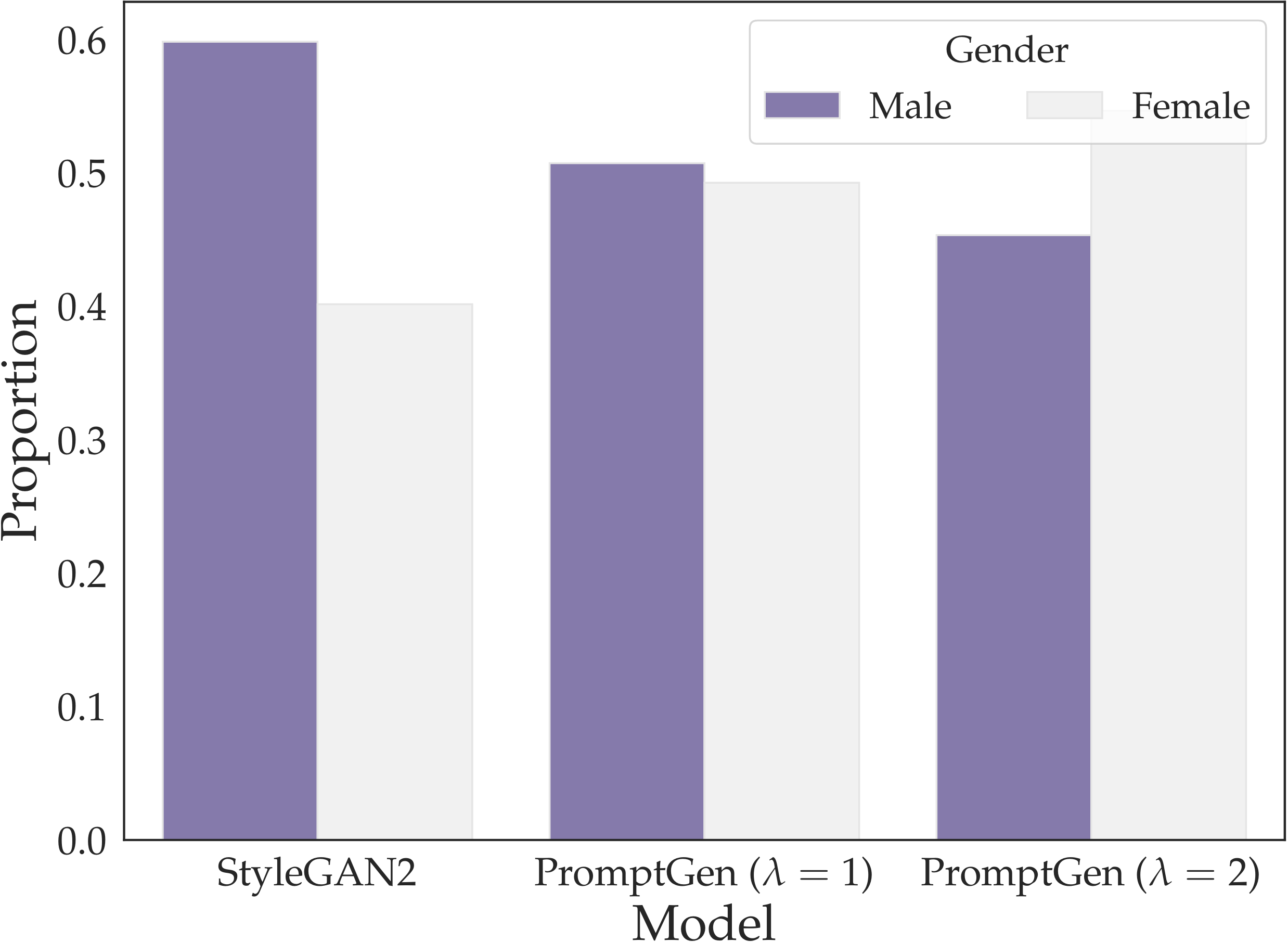}
    }
    \subfigure[Race distribution]{
        \includegraphics[width=0.316\linewidth]{figures/MetFaces-Race.pdf}
    }
    \subfigure[Age distribution]{
        \includegraphics[width=0.316\linewidth]{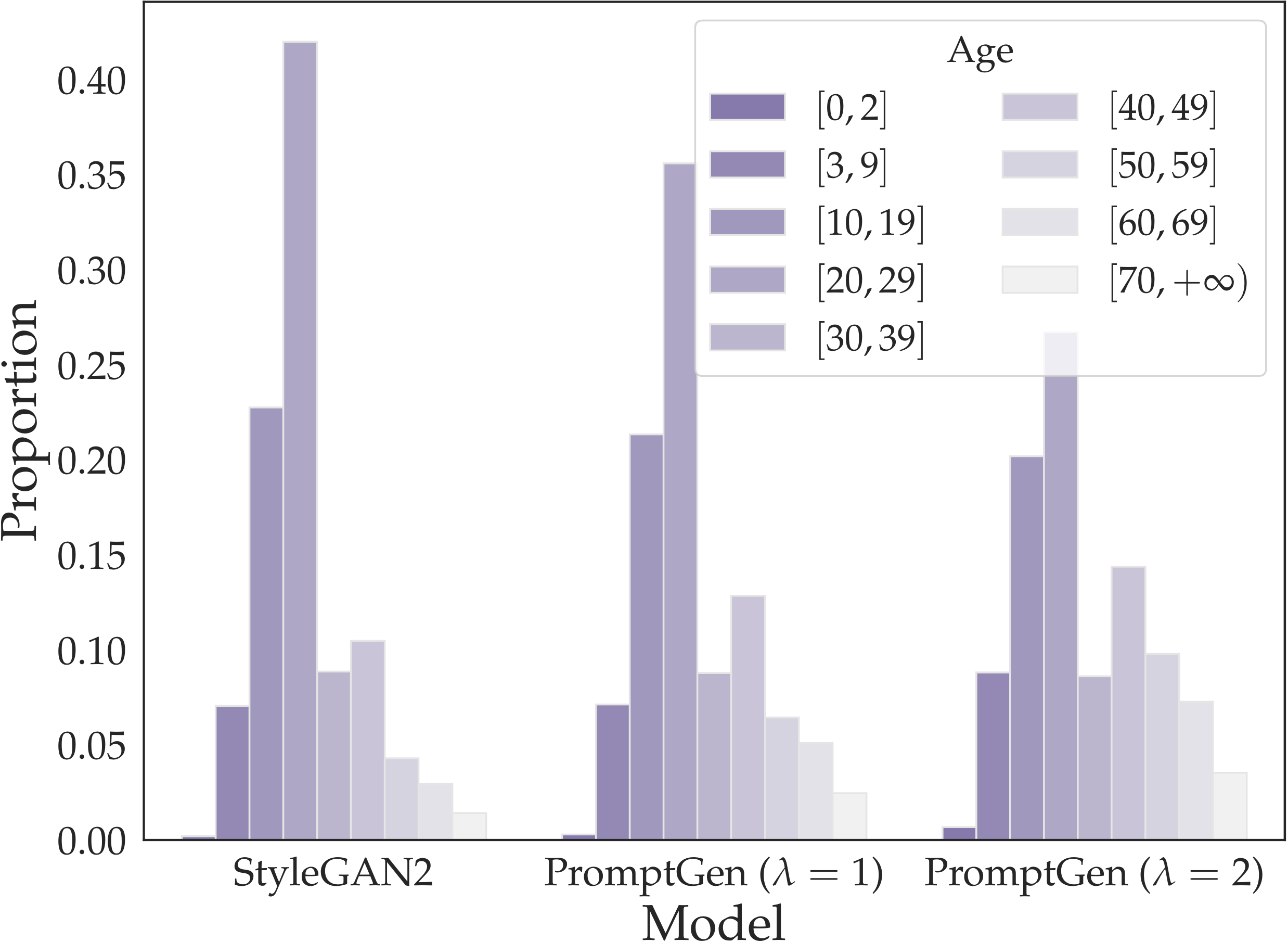}
    }
\caption{\label{fig:debias-results-metfaces-full} Using the moment constraint, \promptgen de-biases the racial, age, and gender distributions of StyleGAN2 trained on MetFaces with truncation $\psi=0.7$. All images are $1024^2$ and resized for visualization. Interestingly, de-biasing the race results in more sculptures, and we postulate that the reason is that almost all paintings and sketches in MetFaces are for white individuals. We fixed the random seed for \promptgen, so we recommend zooming in to see differences between images. }
\end{figure}

Table~\ref{tab:debias-results} provides quantitative results of de-biasing StyleGAN2 across categorical attributes (Figure~\ref{fig:debias-results}). Specifically, we used the pre-trained classifier provided by FairFace \cite{Krkkinen2021FairFaceFA} to classify the attributes of the generated images. We then report the KL divergence between the attribute distribution of the generated images and the uniform distribution. 
Figure~\ref{fig:debias-results-ffhq-full} and Figure~\ref{fig:debias-results-metfaces-full} visualize the de-biasing results. 
We report the KL divergence between the generated distribution and the uniform distribution. 
Interestingly, on MetFaces, de-biasing the race results in more sculptures, and we postulate that the reason is that almost all paintings and sketches in MetFaces are for white individuals.

%%%%%%%%%%%%%%%%%%%%%%%%%%%%%%%%%%%%%%%%%%%%%%%%%%%%%%%%%%%%%%%%%%%%%%%%%%%%%%
%%%%%%%%%%%%%%%%%%%%%%%%%%%%%%%%%%%%%%%%%%%%%%%%%%%%%%%%%%%%%%%%%%%%%%%%%%%%%%

%%%%%%%%%%%%%%%%%%%%%%%%%%%%%%%%%%%%%%%%%%%%%%%%%%%%%%%%%%%%%%%%%%%%%%%%%%%%%%
%%%%%%%%%%%%%%%%%%%%%%%%%%%%%%%%%%%%%%%%%%%%%%%%%%%%%%%%%%%%%%%%%%%%%%%%%%%%%%

\clearpage

\subsection{Decomposing Complex Control via Energy Composition}
\label{subsec:energy-composition}

\begin{figure}[!ht]
\centering
    \subfigure[\label{subfig:complex} CLIP with one complex sentence]{
        \includegraphics[width=0.46\linewidth]{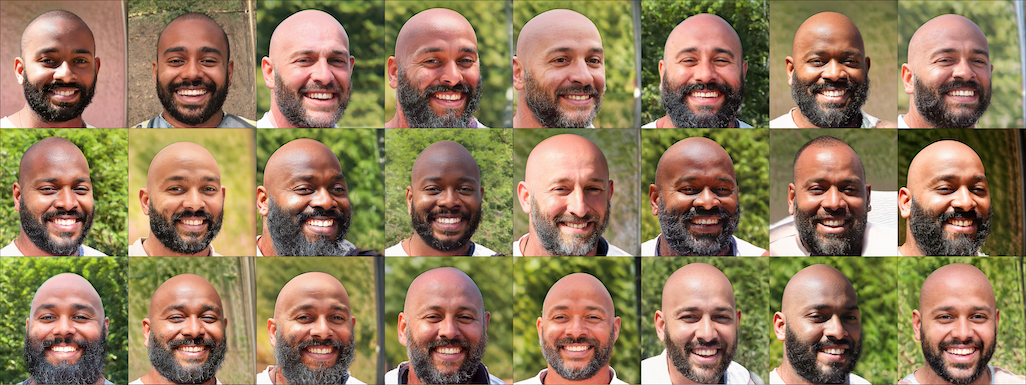}
    }
    \hspace{0.02\linewidth}
    \subfigure[\label{subfig:decompose} CLIP with three simple sentences ]{
        \includegraphics[width=0.46\linewidth]{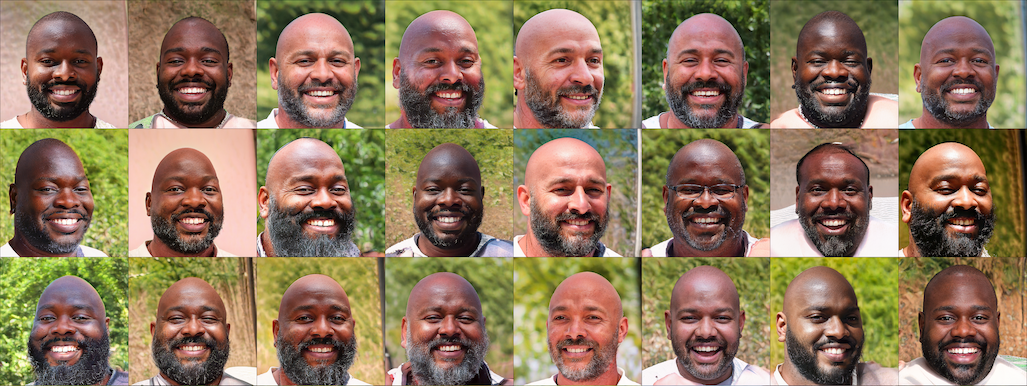}
    }
\caption{\label{fig:decompose-results} Decomposing complex controls (e.g., p\underline{hoto of a bald black man with beard}) into simpler ones improves control performance. See details in Section~\ref{subsec:energy-composition}.
}
\end{figure}

Energy composition $E_{\mathcal{C}}(\boldx) = \sum_{i=1}^{M} \lambda_i E_i(\boldx, \boldy_i)$ in Eq.~\ref{eq:image-energy} allows us to decompose a complex controls into simple ones. For instance, Figure~\ref{subfig:complex} shows that $\boldy_1 =$ p\underline{hoto of a bald black man with beard} ($\lambda_1 = 6000$) generates good samples of men with beard but several of them have a light-skin tone; however, by decomposing it as $\boldy_1 =$ p\underline{hoto of a bald man} ($\lambda_1 = 1500$), $\boldy_2 =$ p\underline{hoto of a black man} ($\lambda_1 = 3000$), and $\boldy_3 =$ p\underline{hoto of a man with beard} ($\lambda_1 = 1500$), we have (on average) darker skin, as shown in Figure~\ref{subfig:decompose}. 
Note that the random seed for Figure~\ref{subfig:complex} and Figure~\ref{subfig:decompose} is the same, and hence the identities are preserved and mostly the skin tone has changed. 

%%%%%%%%%%%%%%%%%%%%%%%%%%%%%%%%%%%%%%%%%%%%%%%%%%%%%%%%%%%%%%%%%%%%%%%%%%%%%%
%%%%%%%%%%%%%%%%%%%%%%%%%%%%%%%%%%%%%%%%%%%%%%%%%%%%%%%%%%%%%%%%%%%%%%%%%%%%%%

\section{Experiments on Generative 3D Face Models}
\label{app:mesh}

%%%%%%%%%%%%%%%%%%%%%%%%%%%%%%%%%%%%%%%%%%%%%%%%%%%%%%%%%%%%%%%%%%%%%%%%%%%%%%
\begin{figure}[t]
\centering
    \subfigure[Controlling mouth stretch]{
        \includegraphics[width=0.48\linewidth]{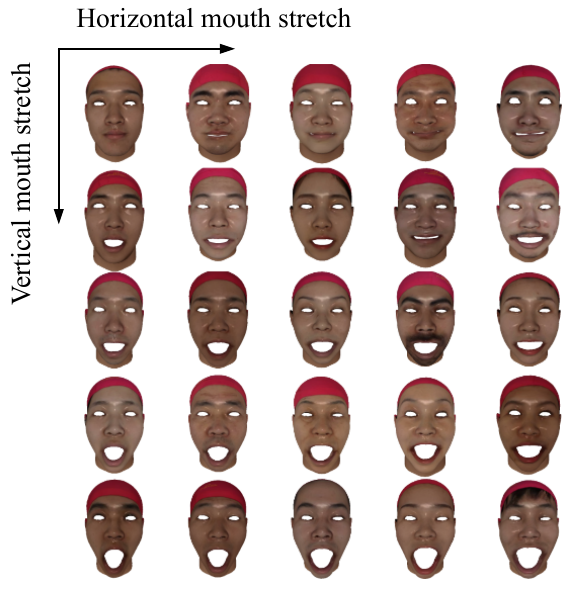}
    }
    %\hspace{0.02\linewidth}
    \subfigure[Controlling eye openness]{
        \includegraphics[width=0.48\linewidth]{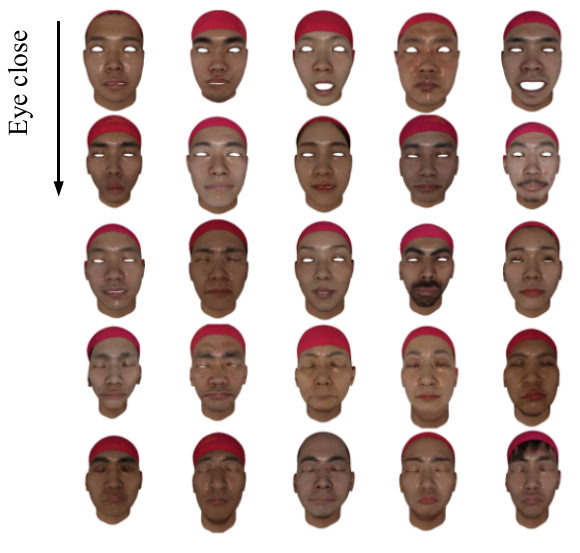}
    }
    \subfigure[Controlling nose-tip stretch]{
        \includegraphics[width=0.48\linewidth]{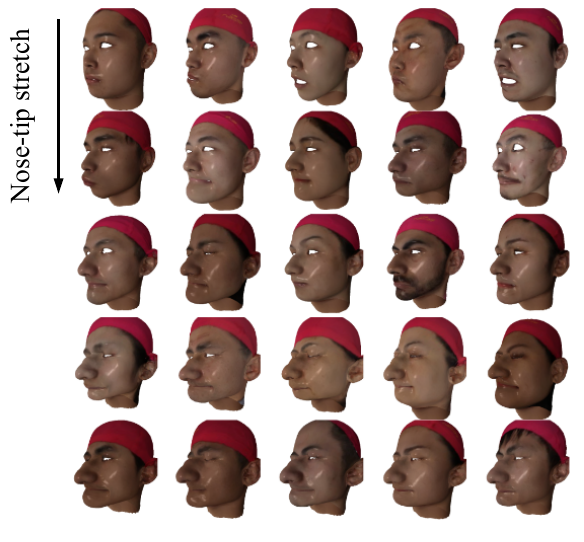}
    }
    %\hspace{0.02\linewidth}
    \subfigure[Controlling lip forward]{
        \includegraphics[width=0.48\linewidth]{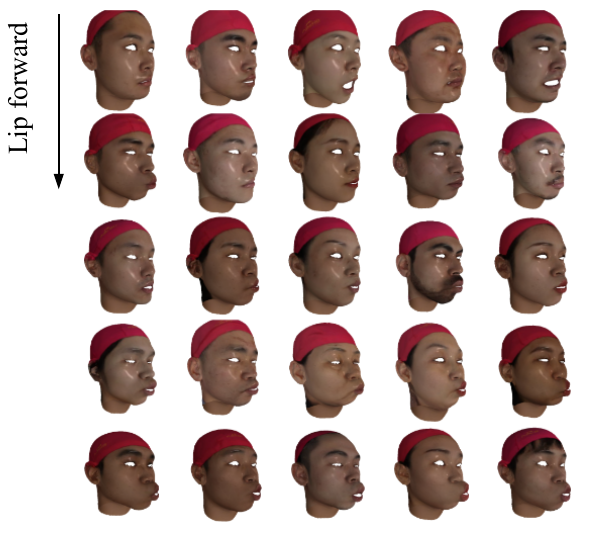}
    }
\caption{\label{fig:mesh} \promptgen with intra-sample signed-distance constraints that generate 3D face models with varying distances in the mouth, eye openness, nose-tip stretch, and lip forward. }
\end{figure}
%%%%%%%%%%%%%%%%%%%%%%%%%%%%%%%%%%%%%%%%%%%%%%%%%%%%%%%%%%%%%%%%%%%%%%%%%%%%%%

3D face modeling has been an active area of research that has recently gained major interest due to applications in virtual humans, deep faces, and digital actors. Existing 3D deep learning generative models build 3D compact representations of shape and appearance capable of modeling non-linearity (e.g., scatter effects, specularities) that is necessary for generating photo-realistic faces. 
%However, existing 3D face generative models learned from data lack the capability to control the generation of 3D faces with constraints such as glasses, beards, or generate faces with a particular geometry. 
Providing a generative model with the capability of generating 3D faces with a particular geometry  (e.g., a specific distance between eyes or nose length) is particularly useful for technical artists when creating new characters. However, this problem remains unaddressed, partially due to the lack of publicly available 3D databases labeled with such constraints. 
We refer to this capability as the intra-sample constraint in 3D generative models, and this section shows how our \promptgen framework is able to achieve this by defining a new energy function based on signed distances. 

We conducted the experiments on the FaceScape dataset~\cite{Yang2020FaceScapeAL}, a large-scale 3D human face dataset consisting of $16,940$ topologically uniformly registered 3D face meshes along with high-quality textures. The dataset contains $847$ identities with $20$ expressions per identity, and each mesh consists of $26,317$ vertices of which $68$ are 3D landmark vertices covering eyes, nose, mouth, jaw, and eyebrows. With some abuse of notation, we denote a 3D mesh as $\boldx \in \mathbb{R}^{3V}$ ($\boldx$ is used to represent an image in other parts of this paper), where $V = 26,317$ is the number of vertices. Given an index $l$ of a vertex (e.g., one of the $68$ landmark vertices), the 3D coordinate of this vertex is defined as $\boldx[l] \in \mathbb{R}^{3}$. We pre-train a generative mesh model $\gen$ that maps each latent code $\boldz \sim \mathcal{N}(\bm{0}, \bm{I})$ to a 3D mesh $\boldx$. 
\promptgen allows us to obtain a geometric control on the 3D meshes.
Given two landmark vertex indices $l_1$ and $l_2$, we put an intra-sample constraint that enforces the signed distance (along a given direction) between the two vertices as $s$. 
To this effect, we define the signed-distance energy as
\begin{equation}
\label{eq:signed-distance}
    E_{\text{signed-distance}}(\boldx) = |d(\bm{v}_1, \bm{v}_2) - s|, \quad \bm{v}_1 = \boldx[l_1], \bm{v}_2 = \boldx[l_2]
\end{equation}
where $d(\cdot, \cdot)$ is the signed-distance (along a given direction), and $|\cdot|$ is the absolute value. 

Figure~\ref{fig:mesh} presents several examples of 3D mesh control, in which random textures are applied to the generated meshes. It shows that \promptgen successfully controls the generative 3D mesh model in terms of the mouth stretch, eye openness, nose-tip stretch, and lip forward.

\section{Error Analysis}
\label{app:error-decomposition}

As we discussed in Section~\ref{sec:conclusion}, the controlled distribution depends on (1) the pre-trained generative model's coverage and (2) the off-the-shelf models used for the control. In this section, we provide some examples of \promptgen failing to generate satisfactory images using the CLIP energy. For the pre-trained generative model, we use StyleGAN2 trained on FFHQ, AFHQ-Wild, or LSUN-Churches. 

\begin{figure}[!ht]
\centering
    \subfigure[\label{subfig:without-beard} p\underline{hoto of a man without beard} (FFHQ)]{
        \includegraphics[width=0.6\linewidth]{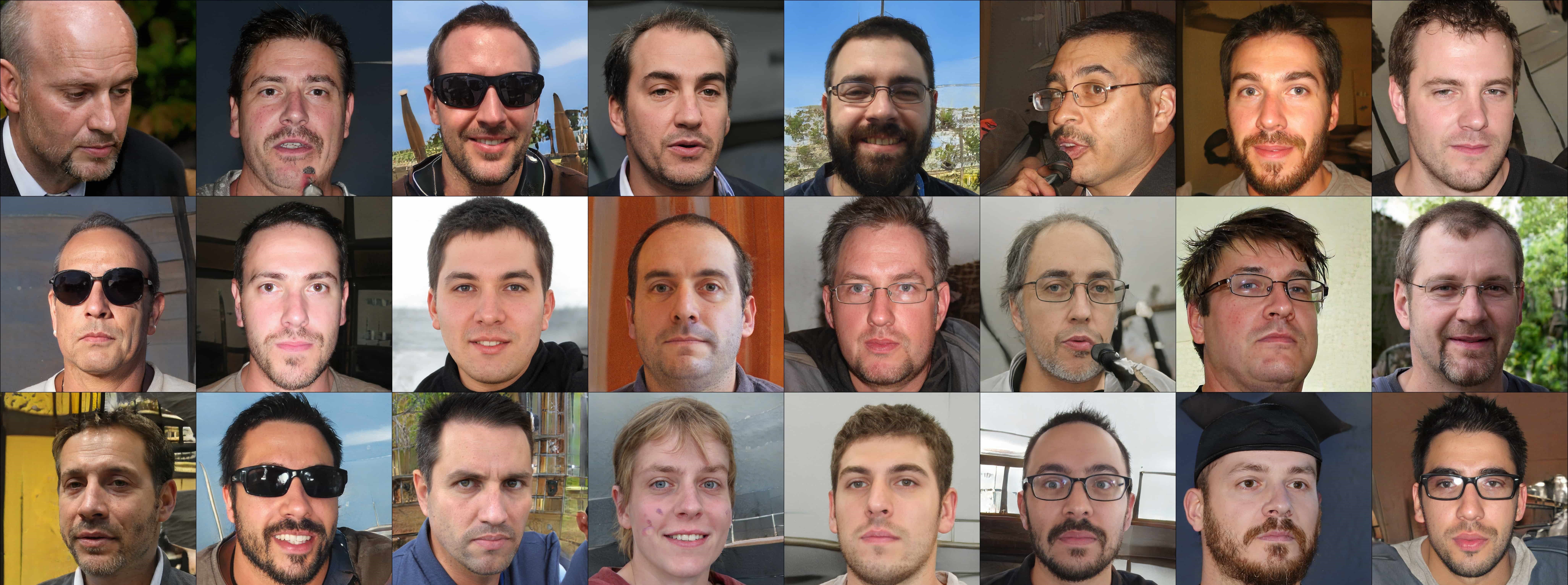}
    }
    \subfigure[\label{subfig:curious-clip-copy} \underline{a }p\underline{hoto of a }p\underline{erson without makeu}p (FFHQ)]{
        \includegraphics[width=0.6\linewidth]{figures/promptgan_person_without_makeup_appendix.png}
    }
    \subfigure[\label{subfig:three-windows} p\underline{hoto of a church with three windows} (LSUN-Churches)]{
        \includegraphics[width=0.6\linewidth]{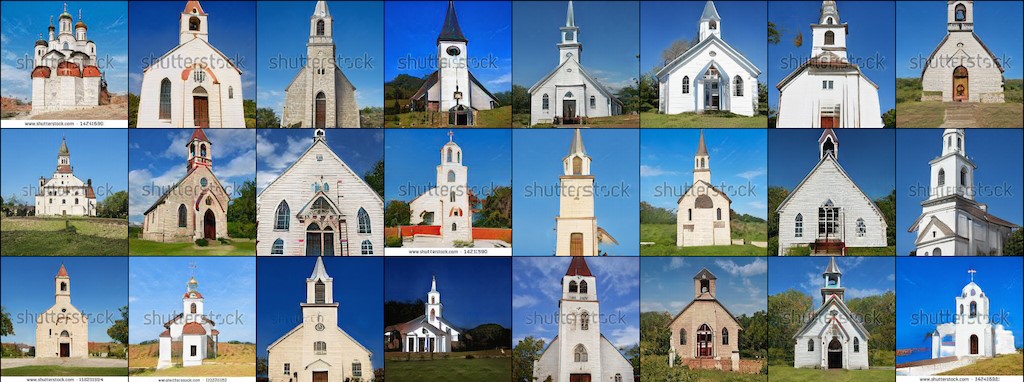}
    }
    \caption{\label{fig:fails-clip} When the text description contains certain linguistic properties (e.g., negation, numerical reasoning), CLIP sometimes fails or shows the ``reporting bias'' that we discuss in Section~\ref{subsec:iterative-control}. For a deeper understanding of these failures and biases, in Figure~\ref{fig:laion-retrieved}, we provide some image samples from the LAION-400M dataset \cite{Schuhmann2021LAION400MOD} with CLIP retrieval, using the same text descriptions as this figure. }
\end{figure}

Our first observation is that the CLIP model sometimes fails in modeling linguistic negation. For instance, the text description p\underline{hoto of a man without beard} results in a distribution of photos of men with beard (Figure~\ref{subfig:without-beard}). Meanwhile, Figure~\ref{subfig:curious-clip-copy} shows that CLIP is capable of modeling the negation of having makeup, but with the ``reporting bias'' discussed in Section~\ref{subsec:iterative-control}. 
Moreover, CLIP seems to have difficulty in numerical reasoning, and gaining control over the count of specific objects in a scene tends to be unsuccessful. We showed this by specifying p\underline{hoto of a church with three windows}, which did not result in the desired specification (Figure~\ref{subfig:three-windows}). 

\begin{figure}[!ht]
\centering
    \subfigure[\label{subfig:without-beard-laion} p\underline{hoto of a man without beard} (CLIP retrieval from LAION-400M)]{
        \includegraphics[width=\linewidth]{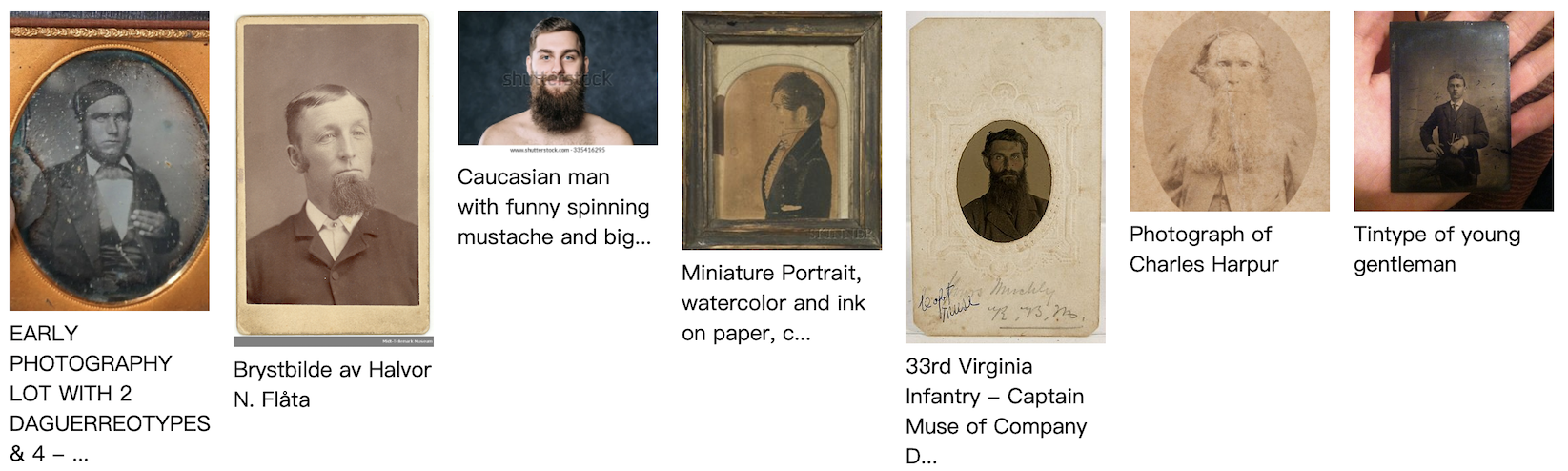}
    }
    \subfigure[\label{subfig:without-makeup-laion} \underline{a }p\underline{hoto of a }p\underline{erson without makeu}p (CLIP retrieval from LAION-400M)]{
        \includegraphics[width=\linewidth]{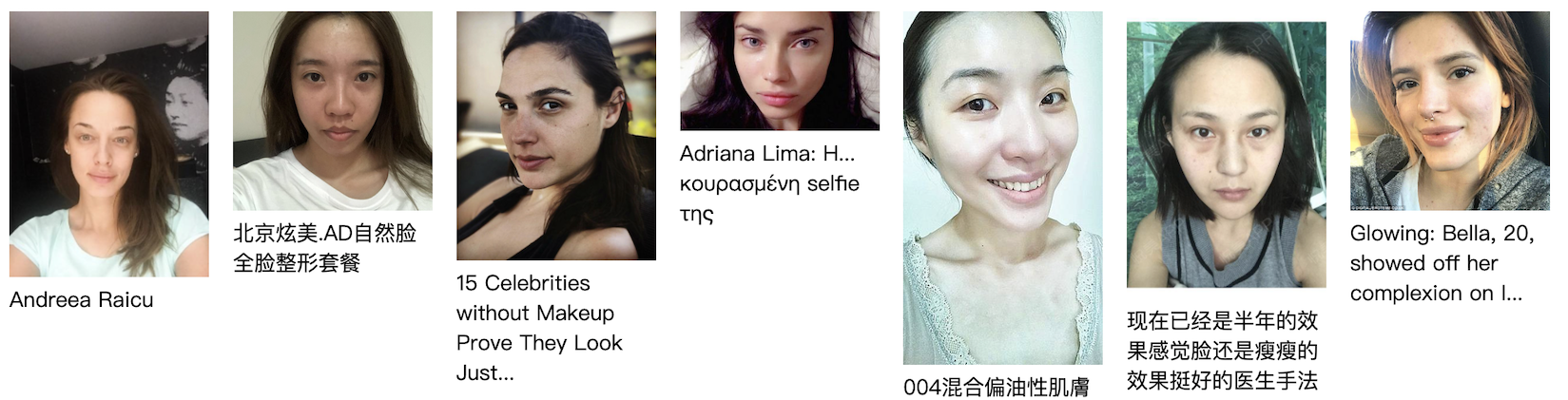}
    }
    \subfigure[\label{subfig:three-windows-laion} p\underline{hoto of a church with three windows} (CLIP retrieval from LAION-400M)]{
        \includegraphics[width=\linewidth]{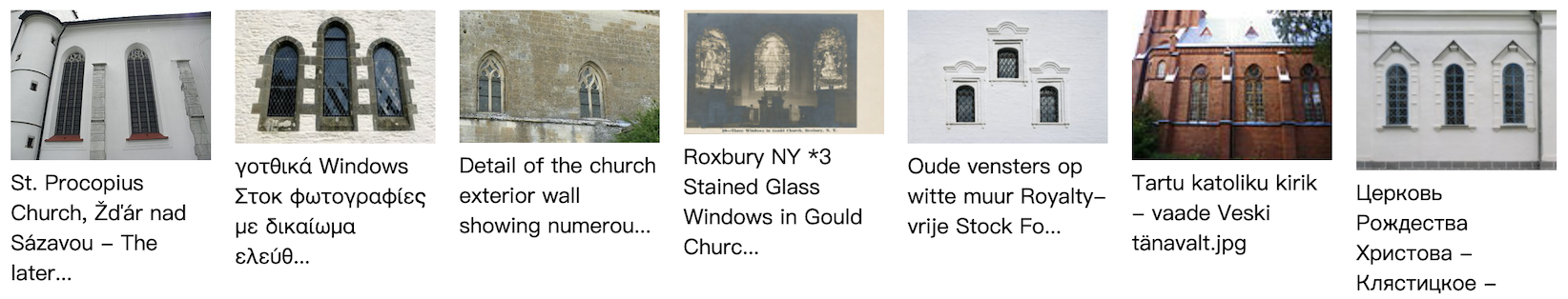}
    }
    \caption{\label{fig:laion-retrieved} For a deeper understanding of the failures and biases illustrated in Figure~\ref{fig:fails-clip}, we provide some image samples from the LAION-400M dataset \cite{Schuhmann2021LAION400MOD} with CLIP retrieval, using the same text descriptions as Figure~\ref{fig:fails-clip}. CLIP retrieval is based on the CLIP embedding similarity between the web images and text descriptions, while the original text below each individual image is not used. In Figure~\ref{subfig:without-beard-laion}, most images have a beard; in Figure~\ref{subfig:without-makeup-laion}, all images are female; in Figure~\ref{subfig:three-windows-laion}, some images do not have exactly three windows. These observations are consistent with those in Figure~\ref{fig:fails-clip}.}
\end{figure}

To gain a deeper understanding of the above failures and biases, in Figure~\ref{fig:laion-retrieved}, we provide some image samples from the LAION-400M dataset \cite{Schuhmann2021LAION400MOD} with CLIP retrieval, using the same text descriptions as Figure~\ref{fig:fails-clip}. 
CLIP retrieval is based on the CLIP embedding similarity between the web images and text descriptions, while the original text below each individual image is not used. 
We observe an impressive consistency between CLIP retrieval and \promptgen: in both Figure~\ref{subfig:without-beard} and Figure~\ref{subfig:without-beard-laion}, most images have beard; in both Figure~\ref{subfig:curious-clip-copy} and Figure~\ref{subfig:without-makeup-laion}, all images are female; in both Figure~\ref{subfig:three-windows} and Figure~\ref{subfig:three-windows-laion}, some images do not have exactly three windows. This consistency suggests that the failures and biases in Figure~\ref{fig:fails-clip} should be mostly attributed to the CLIP model rather than to our \promptgen algorithm. 
We believe that our observation sheds light on the intricacy of contrastive multi-modal (vision-language) pre-training, which is worthy of being further investigated.

\begin{figure}[!ht]
\centering
    \subfigure[\label{subfig:yawn} p\underline{hoto of a }p\underline{erson }y\underline{awnin}g (FFHQ)]{
        \includegraphics[width=0.7\linewidth]{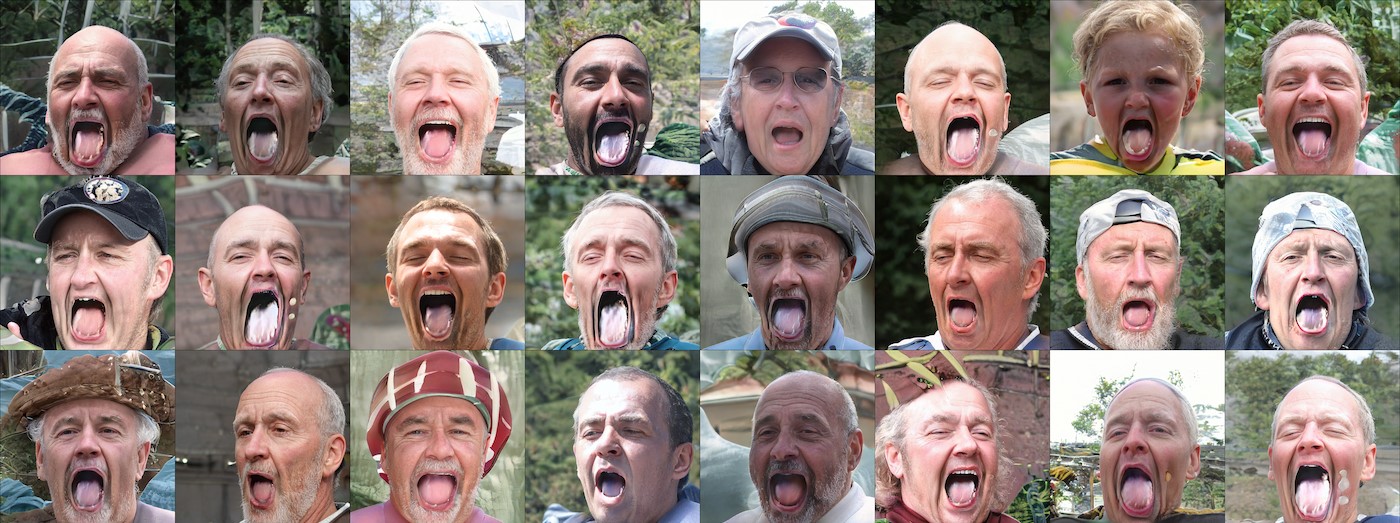}
    }
    \subfigure[\label{subfig:baby-long-hair} p\underline{hoto of a bab}y\underline{ with lon}g\underline{ hair} (FFHQ)]{
        \includegraphics[width=0.7\linewidth]{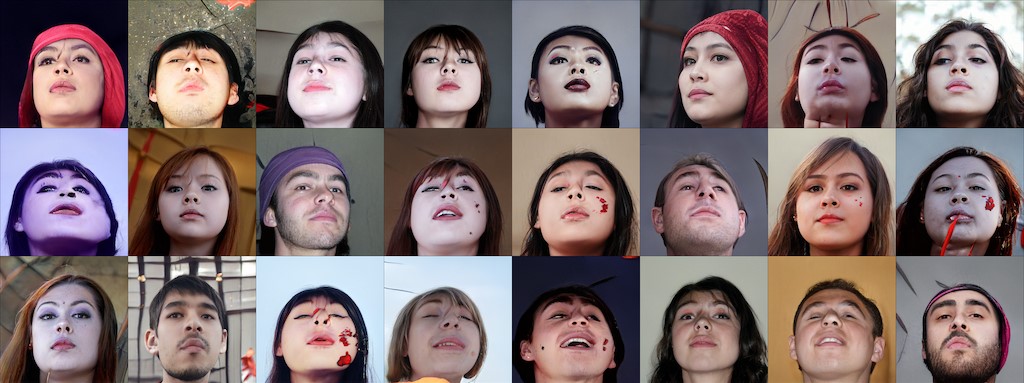}
    }
    \subfigure[\label{subfig:animal-side} p\underline{hoto of an animal}\underline{ from the side} (AFHQ-Wild)]{
        \includegraphics[width=0.7\linewidth]{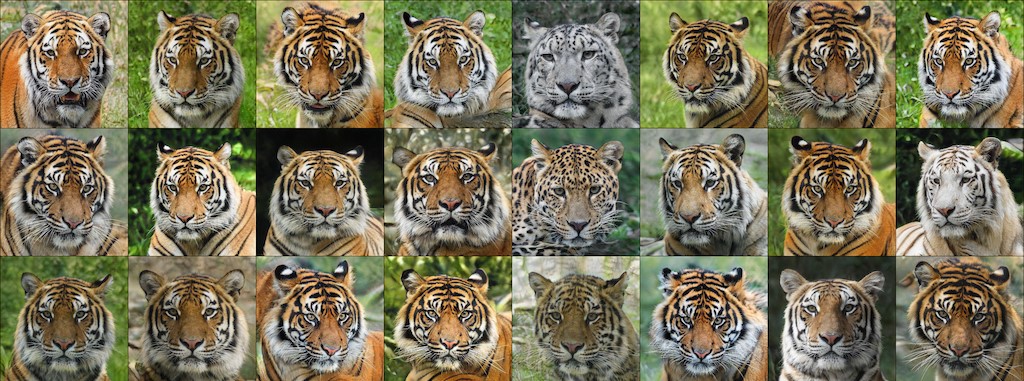}
    }
    \caption{\label{fig:fails-coverage} When the pre-trained generative model fails in covering certain modes required by the text description, unsatisfactory outputs are produced. In this figure, we show several text descriptions that require sampling from low-density regions of the pre-trained generative model's output space. }
\end{figure}

Another observation is that the control tends to fail when the text description requires sampling from low-density regions of the pre-trained generative model's output space. In other words, the control usually fails if the pre-trained generative model does not cover the mode we are trying to gain control over. For example, images faithful to p\underline{hoto of a }p\underline{erson }y\underline{awnin}g and p\underline{hoto of a bab}y\underline{ with lon}g\underline{ hair} are not commonly observed in the FFHQ dataset and, hence, these two text descriptions result in degeneration (Figure~\ref{subfig:yawn}) or weird images (Figure~\ref{subfig:baby-long-hair}). Another example is p\underline{hoto of an animal}\underline{ from the side}, which is not commonly observed in the AFHQ-Wild dataset, and Figure~\ref{subfig:animal-side} shows that the generated images fail to follow this description. Even when the control is successful (e.g., when a complex description is decomposed in Figure~\ref{subfig:decompose}), sampling from low-density regions results in limited diversity (e.g., the backgrounds in Figure~\ref{subfig:decompose} look similar to each other). 

\begin{figure}[!ht]
\centering
    \subfigure[p\underline{hoto of a man without beard} (FFHQ; w/ PPGM)]{
        \includegraphics[width=0.47\linewidth]{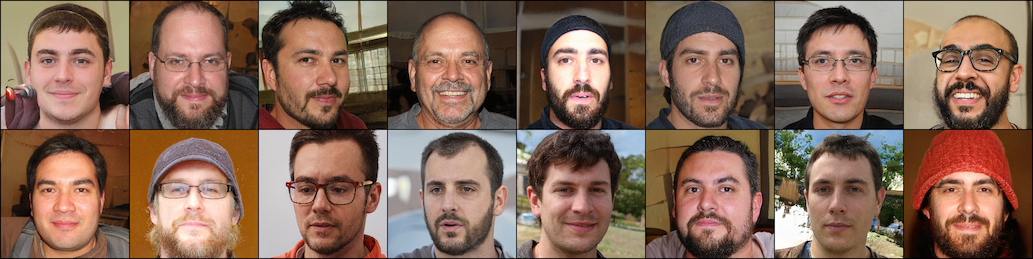}
    }
    \subfigure[p\underline{hoto of a }p\underline{erson }y\underline{awnin}g (FFHQ; w/ PPGM)]{
        \includegraphics[width=0.47\linewidth]{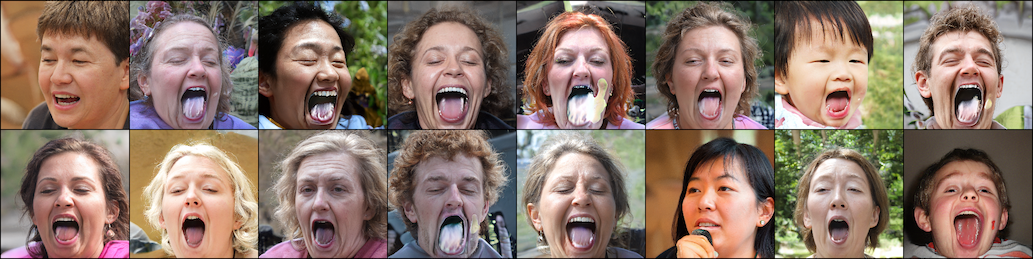}
    }
    \caption{\label{fig:fails-ppgm} Failure modes revealed by \promptgen (Figure~\ref{fig:fails-clip} and Figure~\ref{fig:fails-coverage}) also hold for PPGM. }
\end{figure}

Finally, we also ran some failure controls using PPGM, showing that \promptgen and PPGM reveal similar failure cases of pre-trained generative models and CLIP. Results are shown in Figure~\ref{fig:fails-ppgm}. 

\end{document}